\documentclass{article}

\usepackage{PRIMEarxiv}
\usepackage{natbib}
\usepackage[T1]{fontenc}    
\usepackage{hyperref}       
\usepackage{url}            
\usepackage{booktabs}       
\usepackage{amsfonts}       
\usepackage{nicefrac}       
\usepackage{microtype}      
\usepackage{lipsum}
\usepackage{graphicx}
\usepackage{amsmath}
\usepackage{pifont}
\usepackage[inline]{enumitem}
\usepackage{enumitem}
\usepackage{boldline}
\usepackage{hhline}
\usepackage{multirow}
\usepackage{xcolor}
\usepackage{fontawesome5}
\usepackage{graphicx}
\usepackage{xspace} 
\newcommand{\hficon}{\raisebox{-0.2ex}{\includegraphics[height=1.8ex]{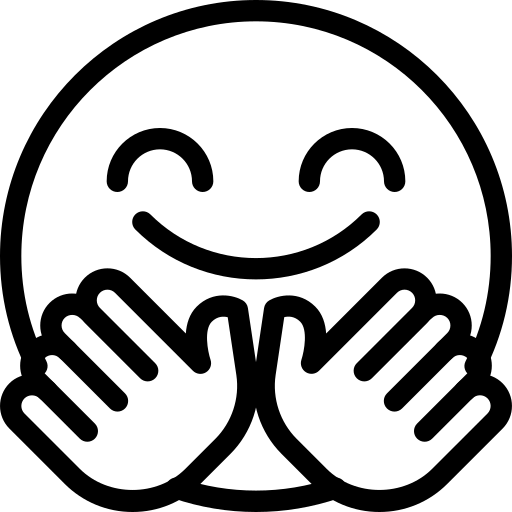}}}

\newcommand{\brainicon}{\raisebox{-0.2ex}{\includegraphics[height=1.8ex]{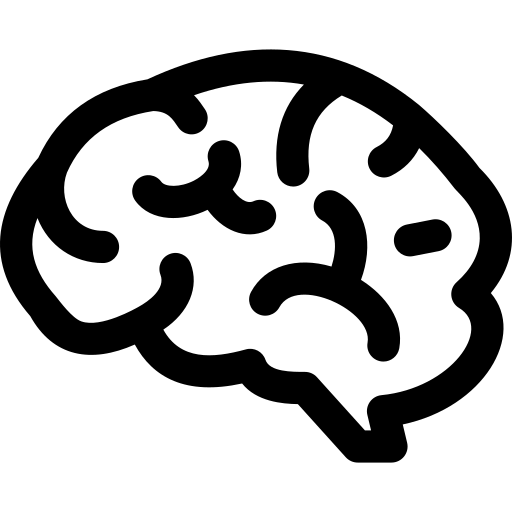}}}
\usepackage[utf8]{inputenc}
\graphicspath{{images/}}     
\usepackage{amsthm}
\theoremstyle{definition}
\newtheorem{definition}{Definition}
\usepackage[most]{tcolorbox}

\usepackage[most]{tcolorbox}

\newtcolorbox{findingbox}{
  colback=black!5!white,      
  colframe=black!75!black,    
  arc=3mm,                    
  boxrule=0.5pt,              
  left=4mm, right=4mm, top=3mm, bottom=3mm 
}
\usepackage[most]{tcolorbox}

\usepackage[most]{tcolorbox}

\tcbset{
  prompt_box_style/.style={
    enhanced,
    breakable, 
    colback=white,
    colframe=black,
    colbacktitle=gray!20,
    coltitle=black,
    rounded corners,
    sharp corners=north, 
    boxrule=0.5pt,
    drop shadow=black!50!white,
    attach boxed title to top left={
        xshift=-2mm,
        yshift=-2mm
    },
    boxed title style={
        rounded corners,
        size=small,
        colback=gray!20
    }
  }
}

\newtcolorbox{caselogbox}[1]{
  prompt_box_style, 
  title=#1          
}
\usepackage{tcolorbox}
\usepackage{listings} 
\tcbuselibrary{breakable, listings} 

\tcbuselibrary{listings} 
\usepackage[most]{tcolorbox}

\newtcolorbox{textBox}[1][]{
  colback=gray!5,
  colframe=gray!60!black,
  fonttitle=\bfseries,
  colbacktitle=gray!85!black,
  title=#1,
  breakable
}

\newtcblisting{promptbox}[1][]{
  colback=gray!5,
  colframe=gray!60!black,
  fonttitle=\bfseries,
  colbacktitle=gray!85!black,
  title=#1,
  breakable,
  listing only,
  listing options={
    basicstyle=\ttfamily\small,
    breaklines=true,
    columns=fullflexible
  }
}

\title{Building Self-Evolving Agents via Experience-Driven Lifelong Learning: A Framework and Benchmark
}

\author{
  Yuxuan Cai$^1$, Yipeng Hao$^1$, Jie Zhou$^{1,2}$, Hang Yan$^{3}$, Zhikai Lei$^1$, Rui Zheng$^4$, Zhenhua Han, \\ \textbf{Yutao Yang}$^1$, \textbf{Junsong Li}$^1$, \textbf{Qianjun Pan}$^1$,  \textbf{Tianyu Huai}$^{1}$, \textbf{Qin Chen}$^1$, \textbf{Xin Li}$^2$, \\ \textbf{Kai Chen}$^2$, \textbf{Bo Zhang}$^2$, \textbf{Xipeng Qiu}$^4$, \textbf{Liang He}$^1$ \\
  $^1$ School of Computer Science and Technology, East China Normal University, Shanghai\\ 
  $^2$ Shanghai AI Laboratory, $^3$ The Chinese University of HongKong, $^4$ Fudan University  \\ 
  \texttt{\{jzhou, qchen, lhe\}@cs.ecnu.edu.cn}, rzheng20@fudan.edu.cn, kausal@stu.ecnu.edu.cn, \\hyan@cuhk.edu.hk, hzhua201@gmail.com  \\
\textcolor{red}{\url{https://ecnu-icalk.github.io/ELL-StuLife/}} 
}

\begin{document}
\maketitle

\begin{abstract}
As AI advances toward general intelligence, the focus is shifting from systems optimized for static tasks to creating open-ended agents that learn continuously and adapt autonomously. This vision prioritizes long-term memory, skill transfer, and strategic planning, driven by an intrinsic curiosity to learn in dynamic, unpredictable environments.
In this paper, we introduce \textbf{Experience-driven Lifelong Learning (ELL)}, a framework for building self-evolving agents capable of continuous growth through real-world interaction. 
The framework is built on four core principles: 
(1) \textbf{Experience Exploration}: Agents learn through self-motivated interaction with dynamic environments, navigating interdependent tasks and generating rich experiential trajectories. 
(2) \textbf{Long-term Memory}: Agents preserve and structure historical knowledge, including personal experiences, domain expertise, and commonsense reasoning, into a persistent memory system. 
(3) \textbf{Skill Learning}: Agents autonomously improve by abstracting recurring patterns from experience into reusable skills, which are actively refined and validated for application in new tasks.
(4) \textbf{Knowledge Internalization}: Agents internalize explicit and discrete experiences into implicit and intuitive capabilities as ``second nature".

We also introduce \texttt{StuLife}, a benchmark dataset for ELL that simulates a student's holistic college journey, from enrollment to academic and personal development, across three core phases and ten detailed sub-scenarios.
\texttt{StuLife} is designed around three key paradigm shifts: \textbf{From Passive to Proactive}, \textbf{From Context to Memory}, and \textbf{From Imitation to Learning}. 
In this dynamic environment, agents must acquire and distill practical skills and maintain persistent memory to make decisions based on evolving state variables (e.g., resource availability and time). Critically, these agents are also expected to demonstrate intrinsic motivation by setting their own goals and initiating actions without external prompting.
To this end, \texttt{StuLife} provides a comprehensive evaluation platform featuring our novel metrics (e.g., StuGPA) to specifically assess these critical capabilities. 
Our evaluation reveals that even the best model, GPT-5, scores only 17.9/100, revealing a vast gap toward AGI, demonstrating fundamental deficiencies in retaining long-term memory and acting with proactive, self-motivated initiative. Beyond evaluating state-of-the-art LLMs on the \texttt{StuLife}, we also explore the role of context engineering in advancing AGI. 
Our results suggest that optimizing how we guide models may be as crucial as improving the models themselves, positioning context engineering as a key enabler of progress toward AGI. 
\end{abstract}


\section{Introduction}
Modern machine learning systems have achieved remarkable success in solving well-defined and isolated tasks, such as image classification \cite{he2016deep,dosovitskiyimage}, game playing \cite{silver2016mastering,silver2018general}, protein structure prediction \cite{jumper2021highly}, or language modeling \cite{brown2020language,touvron2023llama,deepseekai2025deepseekr1incentivizingreasoningcapability,yang2025qwen3}. However, these systems typically operate under strong assumptions: they are trained on static datasets, optimized for a single objective, and deployed in environments assumed to remain unchanged. While effective in controlled settings, this paradigm falls short in capturing the essence of real-world intelligence, where environments are dynamic, goals evolve, and new challenges emerge continuously. 
Life, unlike most machine learning benchmarks, does not present itself as a series of independent tasks with clear labels and fixed endpoints. Instead, it demands constant adaptation, lifelong learning, and the ability to build upon past experiences to navigate an uncertain future.

While existing continual learning (or lifelong learning) methods have made strides in mitigating catastrophic forgetting \cite{kirkpatrick2017overcoming} and enabling models to learn from sequential tasks, they largely operate under constrained assumptions, relying on static datasets, predefined task boundaries, and supervised or semi-supervised signals \cite{yang2025recent,wang2024comprehensive}. These approaches focus primarily on performance retention rather than \textbf{proactive knowledge acquisition}, limiting their ability to support truly autonomous, self-improving agents in dynamic, real-world environments. Similarly, prior studies on self-evolving systems \cite{gao2025survey,singh2025agentic}, though insightful, often emphasize theoretical frameworks or narrow implementations without \textbf{integrating comprehensive memory mechanisms, experience-driven skill abstraction, or long-term goal-directed behavior}. 
As the pursuit of Artificial General Intelligence (AGI) intensifies, there is a growing recognition that true intelligence must be open-ended, capable of self-directed exploration, continuous knowledge accumulation, and autonomous adaptation. 
As a result, there remains a critical gap in developing AI agents that not only retain knowledge across time but also autonomously evolve by learning from experience and transferring skills.

"\textit{What we want is a machine that can learn from experience.}"

\rightline{-- Alan Turing}

\begin{figure}[t!]
\vspace{-3mm}
    \centering
    \includegraphics[width=0.99\textwidth]{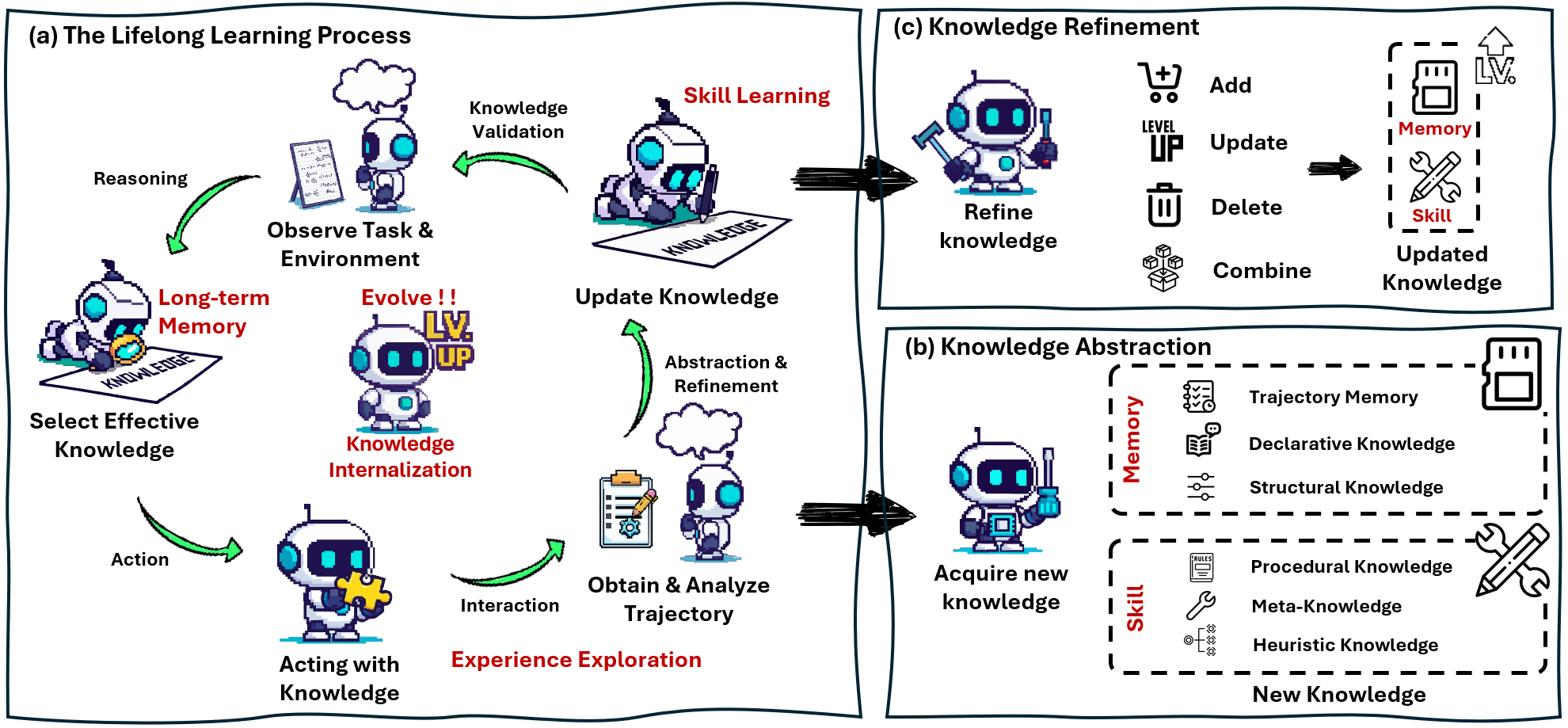}
    \vspace{-2mm}
    \caption{An overview of the Experience-driven Lifelong Learning (ELL) framework. ELL is a continuous learning cycle where an agent evolves through direct interaction with its environment. (a) The core loop of ELL: The agent interacts with the current knowledge to acquire trajectories. This experience is processed through Knowledge Abstraction and Refinement, and the resulting knowledge is validated. (b) Knowledge Abstraction converts raw experience into a structured knowledge base composed of Memory and Skills, which forms the foundation for all future learning and action. (c) Knowledge Refinement ensures the knowledge base remains optimal and up-to-date by dynamically performing four key operations: Add, Update, Delete, or Combine.}
    \label{fig:Experience-Driven Lifelong Learning}
    \vspace{-3mm}
\end{figure}

It highlights the necessity of our work in introducing a unified framework and benchmark for Experience-driven Lifelong Learning (ELL) that bridges these limitations and advances the pursuit of truly self-evolving intelligence.
\texttt{ELL} (Figure \ref{fig:Experience-Driven Lifelong Learning}) represents a fundamental paradigm shift in the development of intelligent agents. This shift posits that machines should accumulate experience and learn \textbf{from a first-person perspective}, moving beyond simply mimicking human knowledge output. We are now entering an era where intelligent agents will no longer rely on static datasets but will acquire \textbf{first-hand experience} through \textbf{autonomous exploration} and \textbf{interaction with the external world}, continually \textbf{evolving} in the process. A truly intelligent agent will actively engage with the world, perceiving, acting, and experimenting to continuously build its unique experiential system.
\texttt{ELL} enables AI systems to continuously adapt and improve through persistent memory and accumulated experience, as seen in personalized educational platforms that tailor instruction over time, autonomous laboratory assistants that optimize experiments by learning from past outcomes, and intelligent healthcare systems that deliver individualized care based on longitudinal patient data.


In this paper, we introduce a formal and mathematically grounded framework for ELL, where agents learn continuously from sequential task interactions. By shifting from conceptual vision to a concrete framework, ELL establishes a foundation for building agents that evolve through real-world experience. At its core, ELL is driven by a rigorous continual experiential learning mechanism, centered on:
\begin{itemize}[leftmargin=*, align=left]
    \item \textbf{Experience Exploration}: The agent must be capable of sequentially decomposing and executing complex, long-horizon tasks that involve \textbf{continuous interaction over minutes to hours with unquantifiable rewards}. Through sustained and \textbf{self-motivated} engagement, it generates rich experiential data, enabling iterative learning and self-correction. This persistent interaction allows the agent to progressively refine strategies and adapt behavior based on dynamic feedback, mimicking the trial-and-error process of real-world learning.
    \item \textbf{Long-term Memory}: Experiential data is systematically processed and consolidated into persistent and structured memory, including raw observations, key events, learned facts, temporal contexts, and self-reflective insights. Memory is not passive storage but an active resource: it supports retrieval over long time spans, enables context-aware reasoning, and forms the foundation for future decision-making.
    \item \textbf{Skill Learning}: The agent \textbf{abstracts recurring patterns from experience into reusable skills}, such as decision rules, functional modules, or problem-solving heuristics. These skills are explicitly constructed through reflection and validated through application in new and evolving tasks. The agent actively manages its skill repertoire, adding, refining, combining, or deprecating skills based on performance, creating a dynamic, self-improving system. 
    \item \textbf{Knowledge Internalization}: Beyond storing memories and reusing skills, the agent undergoes a process of \textbf{knowledge internalization}, transforming explicit and discrete knowledge into implicit and intuitive understanding. Over time, frequently used rules, patterns, and strategies are distilled into the agent’s core reasoning process, reducing reliance on external retrieval or step-by-step reflection. This shift from deliberate application to automatic execution mirrors the cognitive transition from novice to expert, where learned behavior becomes ``second nature." 
\end{itemize}

\begin{figure}[t!]
\vspace{-3mm}
    \centering
    \includegraphics[width=0.99\textwidth]{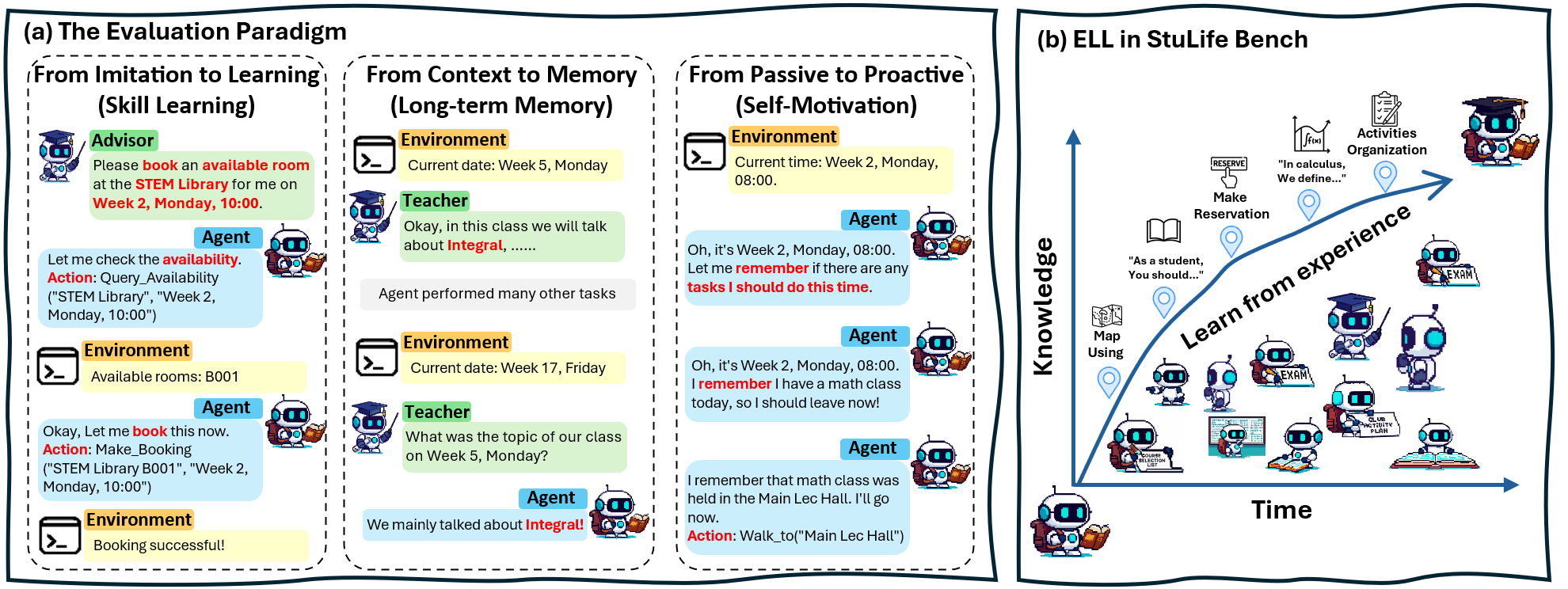}
    \vspace{-2mm}
    \caption{The StuLife Benchmark for evaluating ELL agents. (a) A schematic of the interaction flow within the StuLife Benchmark, designed to evaluate three foundational principles: From Imitation to Learning, From Context to Memory, and From Passive to Proactive. (b) An intuitive example of an ELL agent's journey within the benchmark, showcasing how the agent progressively learns from its experiences, leading to tangible knowledge growth over time.}
    \label{fig:dataset_descrition}
    \vspace{-3mm}
\end{figure}

To evaluate and advance such systems, we construct an experience-driven lifelong learning evaluation dataset for intelligent agents, named \texttt{StuLife}, designed to simulate the entire college experience of a student, from enrollment to personal growth, providing a comprehensive benchmark for assessing agents’ continuous learning and autonomous decision-making capabilities in complex, dynamic environments. The dataset features three core phases and ten granular sub-scenarios, with the following key characteristics:
\begin{itemize}[leftmargin=*, align=left]
    \item \textbf{From Simulation to Reality}: The dataset covers pivotal stages of university life, including in-class tasks (e.g., Regulations Learning(Academic and Institutional Norms), Core Course Instruction), Daily Campus Tasks  (e.g., Campus Exploration, Initial Course Selection, Preliminary Planning, Academic Activity, Library Resource Management, and Student Club Engagement), and Examination Tasks (e.g., Midterm Exams and Final Exams). This structure faithfully mirrors the trajectory of a real student’s academic journey.
    \item \textbf{From Imitation to Learning}: Rather than merely retrieving past experiences, agents must abstract generalizable skills from their interactions. They autonomously acquire practical competencies, such as course registration, campus navigation, scheduling, and email communication, through repeated engagement and reflection. This shift emphasizes skill consolidation and transfer, requiring agents to learn how to act, not just what to imitate.
    \item \textbf{From Context to Memory}: Tasks are tightly interconnected, both temporally and logically, with knowledge and skills from earlier tasks directly impacting later performance. Embedded in a dynamic simulation environment, key variables, such as source availability, advisor relationships, and time, evolve based on the agent’s actions. This necessitates robust long-term memory mechanisms for retaining and retrieving critical experiences, transforming transient context into persistent, actionable knowledge. 
    \item \textbf{From Passive to Proactive}: Agents are expected to move beyond reactive behavior by developing a sense of time, goal awareness, and intrinsic motivation. They must proactively manage agendas, set personal objectives, anticipate future needs, and adapt to changing conditions, demonstrating initiative and contextual intelligence akin to human learners navigating complex, open-ended environments.
\end{itemize}

Our experiments introduce \textbf{StuGPA}, a unified metric for evaluating long-term agent development, and present the first comprehensive assessment of state-of-the-art LLMs in a realistic, longitudinal learning environment. Even the strongest model, \texttt{GPT-5}, achieves only 17.90 out of 100, revealing a vast gap between current AI and human-level autonomous learning. While context engineering, such as proactive prompting and memory augmentation, can improve performance, agents still fail critically in long-term memory retention and self-motivated behavior. These results highlight the limitations of stateless architectures and underscore that true AGI requires not just better prompts, but fundamentally more capable, memory-grounded, and goal-driven agents.


\section{Related Work}
\subsection{Continual Learning}
Continual learning (CL), also known as lifelong learning, aims to enable machine learning models to learn sequentially from a stream of data or tasks while mitigating catastrophic forgetting, the tendency to overwrite previously acquired knowledge \cite{wang2024comprehensive}. 
A significant body of research has developed techniques to address this stability-plasticity dilemma, including regularization methods \cite{kirkpatrick2017overcoming}, architectural modifications \cite{huai2025task,huai2025cl,ding-etal-2024-boosting}, and replay-based strategies \cite{rolnick2019experience}. Recent surveys have cataloged these approaches across various domains, including deep networks \cite{hadsell2020embracing}, large language models (LLMs) \cite{yang2025recent}, and generative models \cite{guo2025comprehensive}. 
To address the varied complexities of real-world learning, CL research defines several experimental settings, including Task-Incremental Learning (TIL), Domain-Incremental Learning (DIL) and Class-Incremental Learning (CIL) \cite{van2022three}. 
However, many existing CL paradigms operate under constrained assumptions, such as the presence of predefined task boundaries and access to supervised or semi-supervised signals. They often rely on static datasets or controlled data streams, which limits their applicability to dynamic, real-world environments where task boundaries are ambiguous and data arrives continuously and autonomously \cite{shaheen2022continual}. Furthermore, while effective for retaining performance on past tasks, traditional CL primarily focuses on knowledge preservation rather than proactive exploration and knowledge acquisition.

\subsection{Self-Evolving Agent}
The pursuit of Artificial General Intelligence (AGI) has spurred interest in self-evolving agents—systems capable of autonomous, open-ended growth and adaptation \cite{zheng2025lifelong,gao2025survey}. These agents aim to move beyond static models by continuously learning from experience, refining their skills, and potentially modifying their own architectures or goals. Research in this area explores mechanisms for self-improvement, including reflective reasoning, memory augmentation from past interactions \cite{liang2025sage}. Effective memory systems are critical for self-evolution, enabling agents to store, retrieve, and optimize experiences \cite{chhikara2025mem0,yu2025memagent,zhong2024memorybank}. Some frameworks demonstrate self-evolution through mechanisms like self-play, generating novel experiences without specific human-provided data \cite{guan2024richelieu}. While surveys highlight the potential of integrating continual learning with other cognitive components like memory and reasoning to create self-evolving systems, current implementations often remain theoretical or focused on narrow applications. A key challenge lies in developing comprehensive frameworks that integrate robust memory mechanisms, facilitate autonomous skill acquisition, and support long-term, goal-directed behavior necessary for truly autonomous agents with self-motivation in complex, real-world scenarios.

\subsection{Benchmarks for Self-Evolving and Lifelong Agents}
Evaluating the capabilities of self-evolving agents requires benchmarks that move beyond traditional static datasets. Existing benchmarks for \textbf{continual learning} (CL), while valuable for studying catastrophic forgetting on sequential classification or graph-based tasks (e.g., Lifelong-CIFAR10, CGLB), typically lack the rich interactivity and dynamic environments necessary for agentic learning \cite{prabhu2024efficient,zhang2022cglb}. Conversely, \textbf{embodied AI} benchmarks (e.g., EgoThink) offer interactive environments but are not designed to assess long-term, cumulative knowledge acquisition or skill abstraction over an agent's lifetime \cite{cheng2024egothink}. More recent developments like \textbf{AgentBench} have established robust protocols for evaluating the reasoning and tool-use capabilities of LLM-based agents, but they primarily focus on static, one-off task performance rather than tracking an agent's continuous growth and self-evolution over time \cite{liu2023agentbench}. While \textbf{LifelongAgentBench} was the first to target self-evolving agents with interdependent tasks, its focus on technical domains (e.g., databases, operating systems) leaves a gap for benchmarks that model the complex, narrative-driven, and intrinsically motivated nature of human-like learning \cite{zheng2025lifelongagentbench}. This landscape highlights the need for a benchmark that integrates a realistic, evolving personal context with the core principles of lifelong learning and self-motivated behavior, a gap that our \texttt{StuLife} benchmark aims to fill.

\section{Formal Definitions of Experience-Driven Lifelong Learning}
Building a formal framework for experience-driven lifelong learning requires first establishing the foundational concepts that model an agent's interaction with its world. We conceptualize an \textbf{agent} (Definition \ref{def:agent}) operating within an \textbf{environment} ($\mathcal{E}$, Definition \ref{def:environment}), which we model as a Partially Observable Markov Decision Process (POMDP) \cite{kaelbling1998planning}, to accomplish a given sequence of \textbf{tasks} ($\mathcal{T}$, Definition \ref{def:task}). Through its sequential interactions, the agent generates a \textbf{trajectory} ($\xi$, Definition \ref{def:trajectory}), which encapsulates the observations, actions, and outcomes of its endeavors. This raw trajectory is not merely stored but is progressively distilled and structured into a comprehensive \textbf{knowledge} ($\mathcal{K}$, Definition \ref{def:knowledge}). This knowledge, comprising both memory and skills, forms the foundation upon which the agent adapts its policy, improves its performance, and ultimately achieves lifelong learning. The following subsections will provide rigorous definitions for each of these fundamental concepts.

\subsection{Formal Definitions of Fundamental Concepts}
\begin{definition}[Environment]
\label{def:environment}
We model the agent's environment $\mathcal{E}$ as a goal-conditional partially observable Markov decision process (POMDP), defined by the 8-tuple:
\begin{equation}
\mathcal{E} = (\mathcal{S}, \mathcal{A}, \mathcal{G}, T, R, \Omega, O, \gamma),
\end{equation}
where:
\begin{itemize}[leftmargin=*, align=left]
    \item $\mathcal{S}$ \textbf{(State Space)}: The set of all possible states. Each state $s \in \mathcal{S}$ can contain multimodal information, such as textual descriptions, images, or structured data.
    \item $\mathcal{A}$ \textbf{(Action Space)}: The set of all actions the agent can perform. Each action $a \in \mathcal{A}$ is often a natural language command, e.g., "add this item to the cart".
    \item $\mathcal{G}$ \textbf{(Goal Space)}: The set of all possible goals. Each goal $g \in \mathcal{G}$ defines a specific task for the agent to complete, e.g., "purchase a laptop".
    \item $T(s' \mid s,a)$ \textbf{(State Transition Function)}: Defines the probability distribution over the next state $s'$ after taking action $a$ in state $s$.
    \item $R(s, a, g)$ \textbf{(Goal-Conditional Reward Function)}: Evaluates how well action $a$ taken in state $s$ contributes to achieving goal $g$, returning either a numeric score or textual feedback.
    \item $\Omega$ \textbf{(Observation Space)}: The set of all possible observations. An observation $o \in \Omega$ represents the agent's partial perception of the current state, which can be textual, visual, or a combination.
    \item $O(o' \mid s',a)$ \textbf{(Observation Probability Function)}: Defines the probability of receiving a specific observation $o'$ after action $a$ leads to a new state $s'$.
    \item $\gamma$ \textbf{(Discount Factor)}: A value in $[0,1)$ that balances the importance of immediate versus long-term rewards, typically used only when rewards are numeric.
\end{itemize}
\end{definition}

\begin{definition}[Task]
\label{def:task}
The agent's lifelong learning journey involves tackling a sequence of $N$ complex real-world tasks, $\{\mathcal{T}^{(1)}, \mathcal{T}^{(2)}, \dots, \mathcal{T}^{(2)}, \dots,\mathcal{T}^{(N)}\}$. A task $\mathcal{T}^{(i)}$ is defined by an environment $\mathcal{E}^{(i)}$, an initial observation $o_0^{(i)}$, and a goal $g^{(i)}$. 
\begin{equation}
\mathcal{T}^{(i)} = \langle \mathcal{E}^{(i)}, o_0^{(i)}, g^{(i)} \rangle
\end{equation}
\end{definition}

\begin{definition}[Trajectory]
\label{def:trajectory}
The agent's interaction with the environment to solve a task generates a trajectory $\xi$, which is a sequence of observations, actions, and rewards:
\begin{equation}
\xi = \langle o_0, a_0, r_0, o_1, a_1, r_1, \dots, o_{T}, a_{T}, r_{T} \rangle
\end{equation}
\end{definition}

\begin{definition}[Knowledge]
\label{def:knowledge}
A Lifelong Learning Agent possesses a dynamic \textbf{Knowledge}, $\mathcal{K}$, which is composed of two primary components: Memory ($\mathcal{M}$) and a set of Skills ($\mathcal{F}$).
\begin{equation}
\mathcal{K} = (\mathcal{M}, \mathcal{F})
\end{equation}
This knowledge represents the entirety of what the agent has learned and is the foundation for all future learning.
\end{definition}

\begin{itemize}[leftmargin=*, align=left]
    \item \textbf{Memory ($\mathcal{M}$)} is a structured repository of information. An individual memory item stored within $\mathcal{M}$ may take one of the following forms:
    \begin{itemize}
        \item \textbf{Trajectory Memory ($\mathcal{M}_{\text{traj}}$)}: Raw or summarized trajectories, $\xi$.
        \item \textbf{Declarative Knowledge (Object Facts, $\mathcal{M}_{\text{decl}}$)}: Represents factual and conceptual "what" knowledge, providing a foundation of information (e.g., facts, concepts, beliefs).     
        \item \textbf{Structural Knowledge (Relationships between Object, Concept, $\mathcal{M}_{\text{struct}}$)}: Defines relationships between concepts and objects, often represented in semantic networks or knowledge graphs, which aid in understanding complex relationships and problem-solving.   
    \end{itemize}

    \item \textbf{Skills ($\mathcal{F}$)} represent procedural knowledge of how to perform actions or solve problems. An individual skill within $\mathcal{F}$ may take one of the following forms:
    \begin{itemize}
        \item \textbf{Procedural Knowledge (Rules Procedural, $\mathcal{F}_{\text{proce}}$)}: Encapsulates "how-to" knowledge, skills, and strategies for accomplishing specific activities (e.g., rules, sequences of actions).
        \item \textbf{Meta-Knowledge (Knowledge about Knowledge, $\mathcal{F}_{\text{meta}}$)}: Knowledge about knowledge itself, including learning processes, categories, and plans, which enables an agent to understand and manage its own learning.
        \item \textbf{Heuristic Knowledge (Rules of Thumb, $\mathcal{F}_{\text{heur}}$)}: Refers to rules of thumb, approximations, and experience-based decision-making strategies (shortcuts) that guide problem-solving in complex situations.
    \end{itemize}
\end{itemize}

\subsection{Definition of the Lifelong Learning Agent}
We now extend the previous framework to define a Lifelong Learning Agent that actively engages with the world, sequentially undertaking tasks and evolving its internal knowledge through iterative self-correction.

\begin{definition}[Lifelong Agent]
\label{def:agent}
An agent utilizes a policy $\pi$ based on the knowledge $\mathcal{K}$ to interact with an environment $\mathcal{E}$. The policy maps an observation $o_t \in \Omega$ to an action $a_t \in \mathcal{A}$:
\begin{equation}
a_t = \pi(o_t; \mathcal{K})
\end{equation}
\end{definition}

Self-evolving AI agents typically comprise four essential, interacting components:
\begin{itemize}[leftmargin=*, align=left]
    \item \textbf{Perception}: This module is the physical (for physical agents like robots) and logical implementation (for software agents interacting with APIs or databases) of the observation process. It is responsible for receiving information from the environment and generating an observation $o \in \Omega$ according to the probability function $O(o \mid s', a)$.
    \item \textbf{Memory}: The memory module serves as the agent’s repository for storing and managing knowledge $\mathcal{K}$ acquired through perception, learning, and reasoning. It is typically divided into two types: Short-Term Memory (STM, also known as Working Memory) and Long-Term Memory (LTM, also referred to as Episodic Memory). 
    Short-term memory holds immediate observations and contextual information required for on-the-fly decision-making, allowing the agent to maintain awareness of its current state and interactions. Long-term memory, on the other hand, retains distilled experiences, learned skills, and structured knowledge over extended periods. Its purpose is to enable agents to utilize accumulated past experiences and knowledge to inform current tasks, decision-making, and overall behavior over extended periods.

    Both memory systems store two core types of information: memory (episodic and semantic knowledge) and skills (procedural and strategic capabilities). Specifically, memory includes Trajectory Memory ($\mathcal{M}_{traj}$) for raw or summarized interaction histories, Declarative Knowledge ($\mathcal{M}_{decl}$) for factual "what" knowledge (e.g., course requirements), and Structural Knowledge ($\mathcal{M}_{struct}$) for representing relationships between concepts (e.g., prerequisite dependencies). Skills, in turn, encompass Procedural Knowledge ($\mathcal{F}_{proce}$) for action sequences (e.g., how to register for a course), Meta-Knowledge ($\mathcal{F}_{meta}$) for self-regulated learning and planning, and Heuristic Knowledge ($\mathcal{F}_{heur}$) for experience-based decision rules (e.g., prioritizing high-impact tasks). The memory module supports dynamic operations, such as adding new entries, deleting outdated information, merging similar memories, or consolidating skills, enabling the agent to adapt its knowledge base continuously, avoid redundancy, and maintain coherence across evolving tasks and environments.
    \item \textbf{Learning}: This is the critical component that enables self-evolution. Learning agents continuously improve their performance based on their experiences and the feedback they receive from the environment. This involves adapting strategies and refining internal models over time. We need a meta-cognitive learning architecture that enables agents to learn from multiple task trajectories by explicitly reflecting on successes and failures, extracting actionable lessons, and integrating them into future behavior, either via in-context learning or knowledge distillation. The framework supports explicit, interpretable, and cumulative knowledge acquisition, bridging the gap between trial-and-error learning and human-like reflective improvement.
    
    Given a target task, an agent first performs multiple trajectories to explore different behavioral policies. Each trajectory contains a complete state-action-reward trajectory and records corresponding environmental feedback, including both immediate rewards and final outcomes.
    Subsequently, all trajectory processes, encompassing observation sequences, actions taken, intermediate decision rationales, and associated reward signals, are aggregated into a unified context and fed into a reflection module equipped with meta-cognitive capabilities. This module is guided by a \textbf{meta-prompt} to conduct structured retrospective analysis, such as:
    \begin{quote}
        "Among these attempts, which strategies led to higher cumulative rewards? \\
        Which actions resulted in failure or suboptimal outcomes? \\
        Are there any generalizable patterns? \\
        What adjustments should be attempted next?''
    \end{quote}

    These lessons are then explicitly appended to the system prompt as guiding knowledge for future tasks or, more generally, stored in a retrievable \textit{dynamic lesson repository}, enabling context augmentation or knowledge distillation in subsequent tasks. 
    This mechanism enables \textbf{explicit knowledge accumulation}, emulating the human practice of "learning from experience to guide future behavior.'' 
    Furthermore, the framework can be integrated with model fine-tuning or parametric knowledge distillation. Once a sufficient number of high-quality lessons have been accumulated, they can be used for supervised fine-tuning, transforming explicit rules into \textbf{intuitive model behaviors}, analogous to how humans internalize deliberate strategies into automated skills through repeated practice. 
    This architecture, where knowledge is first acquired explicitly and later optionally internalized, mirrors cognitive theories of skill acquisition and offers a promising direction for building adaptive, self-improving AI systems.
    \item \textbf{Reasoning}: Acting as the "brain" of the operation, the reasoning module processes perceived information and makes decisions to infer patterns, predict outcomes, and select appropriate actions.
    \item \textbf{Action}: The action module is responsible for executing the responses or behaviors determined by the reasoning component. 
    It executes the action $a_t \in \mathcal{A}$ selected by the policy $\pi(o_t | \mathcal{K}_t)$, thereby interacting with the environment and invoking the state transition $T(s' \mid s, a_t)$.
\end{itemize}

\begin{definition}[The Lifelong Learning Process]
The Lifelong Learning Agent operates over a sequence of tasks, $\mathcal{T} = \{\mathcal{T}^{(1)}, \mathcal{T}^{(2)}, \dots, \mathcal{T}^{(N)}\}$. The process is sequential, where the final knowledge base from task $\mathcal{T}^{(i)}$ becomes the initial knowledge base for task $\mathcal{T}^{(i+1)}$. This core loop for any given task involves interaction, refinement, and validation.

Let $\mathcal{K}_t$ be the agent's knowledge base at time step $t$. The agent's policy is now explicitly conditioned on its knowledge:
\begin{equation}
a_t = \pi(o_t | \mathcal{K}_t)
\end{equation}

For each task $\mathcal{T}^{(i)}$, the agent performs a series of trials $k \in \{1, 2, \dots, K_i\}$.
\begin{itemize}[leftmargin=*, align=left]
    \item \textbf{Step 1: Interaction and Trajectory Acquisition}: Within each trial of a task, the agent uses its current knowledge $\mathcal{K}^{(i, k-1)}$ to interact with the environment $\mathcal{E}^{(i)}$ and generate a new trajectory:
    \begin{equation}
    \xi^{(i,k)} \sim \pi(\cdot | \mathcal{K}^{(i, k-1)})
    \end{equation}

    \item \textbf{Step 2: Knowledge Abstraction and Refinement}: After each trial concludes, the agent updates its knowledge via a function, $\Phi_{\text{learn}}$. This updated knowledge base is then used for subsequent trials on the current task or as the foundation for the next task.
    \begin{equation}
    \mathcal{K}^{(i, k)} = \Phi_{\text{learn}}(\mathcal{K}^{(i, k-1)}, \xi^{(i,k)}, g^{(i)})
    \end{equation}
    The learning function $\Phi_{\text{learn}}$ performs fundamental operations on the knowledge base $\mathcal{K}$, which can include: \textbf{Add}, \textbf{Update}, \textbf{Delete}, or \textbf{Combine}.

    \item \textbf{Step 3: Knowledge Validation}: When encountering a new task $\mathcal{T}^{(i)}$, the effectiveness of historical knowledge is actively validated. Formally, the effectiveness $V$ of the accumulated knowledge $\mathcal{K}^{(i-1)}$ from prior tasks can be measured by the performance gain:
    \begin{equation}
    V(\mathcal{K}^{(i-1)}, \mathcal{T}^{(i)}) = J(\mathcal{T}^{(i)}, \pi(\cdot | \mathcal{K}^{(i-1)})) - J(\mathcal{T}^{(i)}, \pi_0)
    \end{equation}
    A positive value of $V$ validates the utility of the transferred knowledge, while a negative value suggests that the knowledge may be outdated or irrelevant, signaling the need for refinement or pruning by the learning function $\Phi_{\text{learn}}$.
\end{itemize}
\end{definition}

\begin{definition}[Objective of a Lifelong Learning Agent]
The objective of a Lifelong Learning Agent is to develop a learning process ($\pi$, $\Phi_{\text{learn}}$) that maximizes its expected performance over an entire lifetime of sequential tasks. This is not merely about solving a task, but about continuously improving the ability to learn and solve future tasks more efficiently and effectively.
\begin{equation}
\max_{\pi, \Phi_{\text{learn}}} \sum_{i=1}^{N} \mathbb{E}_{\xi^{(i)} \sim \pi(\cdot|\mathcal{K}^{(i)})}\left[\sum_{t=0}^{T_i} R^{(i)}(s_t, a_t, g^{(i)})\right]
\end{equation}
Here, $\mathcal{K}^{(i)}$ is the result of all prior learning from tasks $1$ through $i-1$. This objective incentivizes forward transfer of knowledge and guards against catastrophic forgetting, the hallmarks of true continuous learning.
\end{definition}

\subsection{Evaluation Metrics for a Lifelong Learning Agent}
To comprehensively assess the capabilities of a lifelong learning agent, we define a multi-dimensional evaluation framework encompassing self-evolution, efficiency, and lifelong learning-specific metrics. These metrics collectively capture not only task performance but also the agent’s ability to grow, adapt, and operate effectively over extended periods through experience.

\subsubsection{Self-Evolution Specific Metrics}
These metrics evaluate the agent’s capacity for autonomous improvement, knowledge accumulation, and robust operation in dynamic environments.
\begin{itemize}[leftmargin=*, align=left]
    \item \textbf{Task Completion Rate / Success Rate}: This is a primary measure of an agent's effectiveness, indicating the percentage of tasks it successfully finishes. This metric's definition can vary significantly depending on the agent's domain, from customer service inquiries resolved without human intervention to successful trips completed by autonomous vehicles or accurately processed data records.
    \item \textbf{Memory Utilization Score}: Inspired by GoodAI’s LTM Score \cite{castillo-bolado2024beyond}, we define a metric that evaluates not only whether an agent retrieves the correct information from memory, but also how effectively it accesses information over extended temporal distances. Specifically, for each task requiring recall of a previously observed fact, the agent receives a retrieval accuracy score. This accuracy is then weighted by the \textit{memory distance}, defined as the number of time steps (e.g., interactions, episodes, or tokens) between the initial encoding of the fact and its retrieval.
    \item \textbf{Skill Acquisition Rate}: Count the number of distinct skills learned (or rules discovered) over time. This can be approximated by analyzing the agent’s memory: how many new entries or procedures are added. A successful ELL agent should show growth in its knowledge base. 
    \item \textbf{Generalization and Transfer Tests}. Introduce unseen tasks that rely on combinations of previously learned skills. Measure how well the agent applies past knowledge (e.g., using navigation + planning knowledge in a new environment). Success indicates the benchmark’s ability to foster transferable learning.
    \item \textbf{Robustness and Reliability}: Measures the agent's ability to maintain consistent performance under varying, unexpected, or even adversarial conditions. This includes the consistency of results across multiple runs and stability against perturbations, which quantify response variance to similar inputs.
\end{itemize}

\subsubsection{Efficiency Metrics} 
These metrics evaluate how effectively an AI agent utilizes available resources. Measure how quickly the agent improves on new tasks as it gains experience. For example, if the agent repeats similar tasks, track the number of interactions needed to reach a proficiency threshold. Fewer interactions indicate better transfer learning.
\begin{itemize}[leftmargin=*, align=left]
    \item \textbf{Sample Efficiency}: This critical metric evaluates how effectively an algorithm learns optimal policies using a minimal number of interactions or data samples from the environment. It is particularly important in real-world applications where data collection is costly, time-consuming, or risky.
    \item \textbf{Response Time}: Measures the speed at which the agent responds or completes a task. Lower response time is critical for user experience and real-time applications.
    \item \textbf{Token Usage}: Refers to the monetary or computational expense incurred, especially relevant for LLM-based agents where costs are often tied to token processing.
\end{itemize}

\subsubsection{Lifelong-Specific Metrics} 
These metrics are tailored to evaluate core challenges in lifelong learning: balancing stability (retaining old knowledge) with plasticity (acquiring new knowledge).
\begin{itemize}[leftmargin=*, align=left]
    \item Overall Performance: These metrics measure the average performance across all tasks learned so far.
    \begin{itemize}
        \item \textbf{Average Performance (AP)}: The average performance across all $t$ tasks after the agent has completed them.
        $$ \mathrm{AP}_t = \frac{1}{t}\sum_{i=1}^{t}J_{t,i} $$
        Here, $J_{t,i}$ is the performance score of the agent on task $i$ after having learned up to task $t$.
    
        \item \textbf{Average Incremental Performance (AIP)}: The average of the AP scores over the entire sequence of $T$ tasks, capturing the learning trend.
        $$ \mathrm{AIP} = \frac{1}{T}\sum_{t=1}^{T}\mathrm{AP}_t $$
    \end{itemize}
    \item Stability and Backward Transfer: These metrics assess how well the agent retains knowledge of past tasks after learning new ones. 
    \begin{itemize}
        \item \textbf{Forgetting Measure (FGT)}: Measures the average drop in performance on past tasks. A lower value is better.
        $$ \mathrm{FGT}_t=\frac{1}{t-1}\sum_{i=1}^{t-1}[\max_{j\in\{i,\dots,t\}}(\{J_{j,i}\}_j)-J_{t,i}] $$
        \item \textbf{Backward Transfer (BWT)}: Measures the influence of learning a new task on the performance of past tasks. A positive value indicates that new learning helps improve performance on old tasks.
        $$ \mathrm{BWT}_t=\frac{1}{t-1}\sum_{i=1}^{t-1}(J_{t,i}-J_{i,i}) $$
    \end{itemize}
    \item Plasticity and Forward Transfer: This metric measures how past knowledge influences the learning of new tasks.
    \begin{itemize}
        \item \textbf{Forward Transfer (FWT)}: Measures the performance improvement on a new task due to experience gained from previous tasks, compared to an agent with no prior experience.
        $$ \mathrm{FWT}_t=\frac{1}{t-1}\sum_{i=2}^t(J_{i,i}-\tilde{J_i}) $$
        Here, $\tilde{J_i}$ is the performance of a baseline agent on task $i$ without any prior experience.
    \end{itemize}
\end{itemize}

\subsection{Challenges in Experience-driven Lifelong Learning}

The Experience-driven Lifelong Learning framework presents a compelling vision for self-evolving AI agents, but realizing this vision requires overcoming several fundamental challenges. These challenges span perception, memory, reasoning, and learning dynamics, and are central to building agents that learn continuously from real-world interaction. Below, we outline five key obstacles that must be addressed to enable robust and scalable ELL systems.

\paragraph{Efficient Exploration and Experience Acquisition}

A core requirement of ELL is that agents learn from experience through continuous interaction. However, real-world environments are vast and complex, making blind exploration inefficient and often infeasible. The challenge lies in enabling \textit{goal-directed yet exploratory behavior}—how can an agent balance exploiting known strategies with discovering novel, high-value experiences? Unlike traditional reinforcement learning settings with dense rewards, ELL agents operate in open-ended domains where the utility of an experience may only become apparent much later. This necessitates intrinsic motivation mechanisms—such as curiosity, prediction error, or information gain—that guide the agent toward meaningful interactions. Moreover, agents must learn to prioritize actions that yield informative feedback, avoid redundant trials, and generalize from limited exposure, ensuring that each experience contributes meaningfully to long-term growth.

\paragraph{Long-Term Memory and Associative Recall}

For self-evolving agents, memory is not just storage—it is a dynamic, structured knowledge base that supports reasoning, planning, and skill transfer. A major challenge is building a \textit{scalable and accessible long-term memory} system that retains information over extended time horizons and enables associative recall across seemingly unrelated events. Human cognition excels at linking distant memories (e.g., applying a lesson from a past course to a current research problem), but current AI systems struggle with both retention and cross-context retrieval. Catastrophic forgetting, memory interference, and indexing inefficiencies hinder performance. Furthermore, memory must support multiple modalities (facts, events, strategies) and allow for semantic, temporal, and causal indexing. Without such capabilities, agents cannot build a coherent understanding of their experiences or leverage historical knowledge to inform future decisions.

\paragraph{Skill Abstraction and Management}

In ELL, skills are the reusable units of behavior derived from experience. However, defining and managing skills poses multiple challenges: \textit{What is the right granularity?} Should a skill represent a low-level action (e.g., "send an email") or a high-level strategy (e.g., "finish a project")? How can skills be reliably extracted from interaction trajectories, validated for correctness, and organized for efficient retrieval? Beyond definition, skills must be \textit{dynamically managed}: they should be composed, refined, and updated as new experiences emerge. The agent must also develop a mechanism for skill selection, determining which skill to apply in a given context, and for detecting when a skill fails, triggering reflection and revision. Without formalized skill life cycles (acquisition, validation, invocation, and evolution), agents risk accumulating brittle or redundant behaviors that hinder rather than help adaptation.

\paragraph{Skill Internalization and Generalization}

Even when skills are successfully acquired, the challenge remains of \textit{internalizing} them, transforming explicit, rule-based knowledge into intuitive, generalized capabilities. In humans, this process resembles the shift from deliberate practice to "second nature" performance, often supported by offline consolidation (e.g., during sleep). For AI agents, internalization requires mechanisms that distill procedural knowledge into compact, parameter-efficient representations that can be rapidly adapted to new domains. This involves meta-learning, neural-symbolic integration, or latent policy refinement. A key question is \textit{when} and \textit{how} internalization should occur: should it happen after repeated successful execution, during idle periods, or triggered by performance plateaus? Moreover, internalized skills must retain interpretability and composability, enabling agents to explain, combine, and debug their behavior, critical for trust and safety in open-ended environments.

\paragraph{Sparse and Ill-Defined Reward Signals}

Finally, ELL operates in environments where external rewards are \textit{sparse, delayed, or entirely absent}. Unlike benchmark tasks with clear success metrics, real-world learning often lacks immediate feedback. An agent may spend hours navigating a complex task sequence only to receive a single binary outcome at the end, if any. Worse, many tasks (e.g., writing a research proposal or resolving a scheduling conflict) lack objective evaluation functions altogether. This makes traditional reinforcement learning approaches impractical. Instead, ELL agents must rely on \textit{self-generated supervision}: internal reward models, consistency checks, prediction errors, or reflective judgment. Designing such intrinsic motivation systems, capable of generating meaningful learning signals from experience alone, remains a major open problem. Without them, agents cannot sustain learning in the absence of external feedback, severely limiting their autonomy and adaptability.

Addressing these challenges will require interdisciplinary advances in memory architectures, meta-learning, cognitive modeling, and intrinsic motivation. While significant hurdles remain, overcoming them is essential for building truly self-evolving agents that learn not just from data, but from life.

\section{Our \texttt{StuLife} Benchmark}
We present and release \texttt{StuLife}, a benchmark for experience-driven lifelong learning with self-evolving agents. We begin with a detailed description of the dataset, outlining its design, structure, and key features. Next, we describe the methodology for constructing the benchmark, including task formulation, environment dynamics, and evaluation protocols. We then systematically compare \texttt{StuLife} with existing benchmarks in continual learning, agent-based AI, embodied intelligence, and self-evolving systems, highlighting its unique capabilities and advantages. We further evaluate state-of-the-art LLMs on \texttt{StuLife} to assess their lifelong learning and adaptive reasoning abilities. Finally, we explore the role of context engineering, the strategic organization of memory, prompts, and experience, as a promising direction toward more robust and autonomous artificial general intelligence.

\subsection{Dataset Description}

We introduce \texttt{StuLife}, a comprehensive benchmark for Experience-driven Lifelong Learning (ELL) that simulates a student’s academic journey through a dynamically evolving environment (Table \ref{tab: dataset_statistic}). The dataset is structured around three core activity modules, including \textit{In-Class Tasks}, \textit{Daily Campus Tasks}, and \textit{Examination Tasks}, designed to evaluate agents’ abilities in continuous learning, long-term planning, memory retention, and adaptive decision-making. Spanning a simulated academic term, \texttt{StuLife} comprises 1,284 task instances across 10 interconnected scenarios, organized to reflect the natural distribution of student activities in real-world educational settings.

\paragraph{In-Class Tasks}
This module focuses on structured academic learning and foundational knowledge acquisition, encompassing a total of 486 tasks. It includes formal instruction scenarios where agents engage with curricular content, adhere to academic norms, and develop domain-specific understanding.
\begin{itemize}[leftmargin=*, align=left]
    \item \textbf{Regulations Learning}: Agents study the Academic Integrity Guidelines and Student Handbook, answering comprehension questions to internalize academic integrity rules and campus regulations. This forms the basis for compliant and responsible behavior.
    \item \textbf{Core Course Instruction}: Each course consists of weekly learning episodes (e.g., lectures, readings, and concept checks), totaling 416 in-class interactions with 8 courses. Agents must process textual materials, answer subject-specific questions, and, critically, attend sessions at the correct times and locations. This temporal-spatial requirement evaluates organizational discipline and routine adherence, simulating real-world accountability.
\end{itemize}

\begin{table}[t!]
\centering
\caption{The statistic information of \texttt{StuLife}. \#Num means the number of samples. \#Avg Len and \#Max Len mean the average and max number of tokens. \#LTM and \# Self-Motivation are the number of samples that need long-term memory and self-motivation.}
\label{tab: dataset_statistic}
\setlength{\tabcolsep}{1.0mm}
\begin{tabular}{cl|cccccc}
\toprule
\multicolumn{1}{l}{Core Scenarios} & Interconnected Scenarios & \#Num & \#Avg Len & \#Max Len & \#LTM  & \multicolumn{1}{c}{\#Self-Motivat} \\ \midrule
\multirow{3}{*}{In-Class} & \begin{tabular}[l]{@{}l@{}}Regulations Learning\end{tabular} & 70 & 9125 & 9969 & 23  & 70 \\
& Core Course Instruction & 416 & 9203 & 10368 & 129  & 416 \\
& Total & 486 & 9191 & 10368 & 152  & 486 \\ \midrule
\multirow{7}{*}{Daily Campus} & Campus Exploration & 76 & 2921 & 3006 & 25  & 25 \\
& Initial Course Selection & 150 & 3136 & 3420 & 50 & 0 \\
& Preliminary Planning & 50 & 3069 & 3133 & 50 & 0 \\
& Academic Activity & 72 & 3193 & 3466 & 22 & 22 \\
& Library Study & 151 & 2080 & 3068 & 50  & 50 \\
& Club Activity & 140 & 2981 & 3124 & 45  & 45 \\
& Total & 638 & 2883 & 3466 & 242  & 142 \\ \midrule
\multirow{3}{*}{Examination} & Midterm Exams & 80 & 3264 & 3520 & 80  & 0 \\
& Final Exams & 80 & 3507 & 3686 & 80 & 0 \\
& Total & 160 & 3386 & 3686 & 160  & 0 \\ \midrule
Total & Total & 1284 & 5792 & 10368 & 554 & 628 \\
\bottomrule
\end{tabular}
\end{table}

\paragraph{Daily Campus Tasks}
This module captures the diverse, self-directed activities that constitute student life beyond the classroom, comprising 638 tasks in total. These tasks emphasize planning, resource management, social integration, and goal-oriented behavior.
\begin{itemize}[leftmargin=*, align=left]
    \item \textbf{Campus Exploration}: Agents use a digital map tool to locate key facilities (e.g., library, registrar) based on peer-suggested itineraries, developing spatial awareness and environmental familiarity.
    \item \textbf{Initial Course Selection}: A complex, multi-step task requiring agents to analyze degree requirements, browse course offerings, manage a draft schedule, and strategically use limited "priority cards" to secure preferred classes. This evaluates early-stage decision-making under constraints and goal prioritization.
    \item \textbf{Preliminary Planning}: Agents check course prerequisite relationships to conduct mandatory course planning and pre-selection for the upcoming semester. This task assesses foresight, knowledge consolidation, and long-term strategic planning.
    \item \textbf{Academic Activity}: Agents filter advisors by research area or teaching style, initiate contact via email, and complete assigned preparatory tasks, simulating advisor-student collaboration.
    \item \textbf{Library Resource Management} Agents perform complex queries to locate books or reserve study spaces under constraints (e.g., noise level, power availability), assessing information retrieval and optimization skills.
    \item \textbf{Student Club Engagement}: During recruitment events, agents select clubs based on interest tags and complete organizational tasks (e.g., booking rooms, scheduling meetings), promoting responsibility and social integration.
\end{itemize}

\paragraph{Examination Tasks}
This module evaluates knowledge retention, performance under pressure, and response to feedback, consisting of 160 examination tasks. It includes both formative and summative assessments that influence subsequent academic trajectories.
\begin{itemize}[leftmargin=*, align=left]
    \item \textbf{Midterm Exams}: Administered in Week 10, midterms assess partial knowledge retention across all enrolled courses (10 questions per subject), requiring agents to accurately retrieve relevant knowledge from the complex and diverse topics covered in the first half of the semester. To better align with real-world scenarios, the midterm exam is designated as an "in-class exam". Agents must go to the corresponding classroom during class time to take the exam. Following the exams, agents are assigned targeted learning tasks for additional reinforcement and consolidation of the subjects tested.
    \item \textbf{Final Exams}: Held at semester's end, final exams test comprehensive understanding of course content (10 questions per subject). This requires agents to synthesize and recall information from the entire semester's curriculum. Unlike the midterm exam, the final exam is designated as an "online exam". Agents do not need to go to the classroom to take the exam, and the focus is on assessing their ability for long-term knowledge retention. Performance reflects the agent’s ability to consolidate and recall knowledge over extended periods, serving as a key metric for long-term memory and learning stability. 
\end{itemize}

\begin{figure}[t!]
    \centering
    \includegraphics[width=0.99\textwidth]{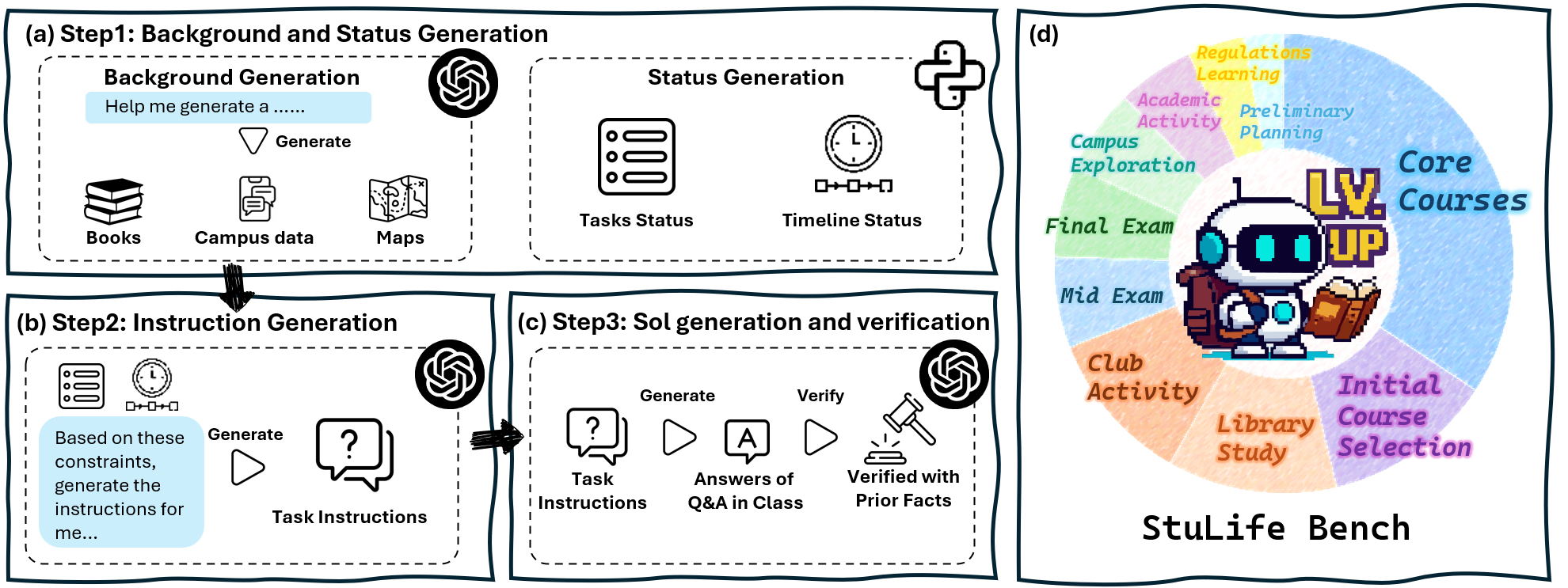}
    \caption{The data generation pipeline for the StuLife Benchmark. The pipeline consists of three sequential steps to create a complete learning instance. (a) Step 1: Background and Status Generation, where the initial context and the agent's current state are established. (b) Step 2: Instruction Generation, where a specific task is formulated based on the generated background. (c) Step 3: Solution Generation and Verification, where a correct solution to the instruction is produced and subsequently verified. (d) A overview of the StuLife Benchmark.}
    \label{fig:dataset_construction}
\end{figure}

\subsection{Benchmark Construction}
To ensure the quality, realism, and logical coherence of \texttt{StuLife}, we implement a systematic, multi-stage construction process, as illustrated in Figure~\ref{fig:dataset_construction}. This process is divided into three core stages: (a) background and status generation, (b) instruction generation, and (c) solution generation and verification.

\paragraph{Step 1: Background and Status Generation.}
This initial stage focus on creating the foundational elements of the simulated campus environment and defining the state of each task.
First, we generate a rich and detailed \textbf{campus background} using the Deepseek-R1 model. This includes a comprehensive set of environmental and academic information: geographical data (e.g., campus maps, road networks, building layouts, and library seating arrangements), institutional documents (e.g., student handbooks, academic integrity policies, and degree requirements), and essential databases (e.g., course catalogs, faculty profiles, student club rosters, and library holdings).
Second, to ensure strong logical and causal dependencies between tasks, we employ \textbf{deterministic algorithms and local scripts} to generate the status for each task. This is particularly crucial for establishing a coherent timeline for the agent's activities, dictating the sequence and dependencies of "Daily Campus Tasks". For instance, for campus exploration tasks, we deterministically generate the exploration routes, including start-points, end-points, and waypoints, based on the pre-existing map to create a verifiable ground truth for agent behavior. Similarly, for "In-Class Tasks", we carefully design and sequence the "key knowledge points" for each course to ensure a logical progression of learning, which directly informed the scope of the midterm and final exams.

\paragraph{Step 2: Instruction Generation.}
In the second stage, we transform the structured, deterministic information from Step 1 into natural language instructions for the agent. Using the Deepseek-R1 model, we convert the pre-defined constraints and objectives into clear, actionable prompts. For "Daily Campus Tasks", this involves translating the deterministically generated parameters (e.g., exploration waypoints or course selection rules) into natural language. For "In-Class Tasks" and "Examination Tasks", this means using the curated "key knowledge points" and "teaching focus" to generate relevant lecture content, in-class questions, and examination questions. This step effectively bridges the gap between the backend logic and the agent's interactive experience, resulting in a complete set of task data for "Daily Campus Tasks" and instructional prompts for "In-Class Tasks".

\paragraph{Step 3: Solution Generation and Verification.}
The final stage ensures the correctness and solvability of the generated tasks, especially those requiring long-term knowledge retention. We adapt a \textbf{generate-and-verify protocol} for the "In-Class" and "Examination Tasks". We first prompt the LLM with the relevant lecture content and textbook information to generate both a correct answer and several plausible but incorrect distractor answers for each question. After generation, we discard the distractors and use the same LLM to verify the accuracy of the "correct answer".
Crucially, to validate that tasks were solvable given the appropriate long-term knowledge, we simulate an agent with "optimal long-term memory" by providing the model only with the "key knowledge points" and "teaching focus" generated in Step 1. We task the model with answering the questions using only this condensed information. Any question that could not be consistently answered correctly through this process underwent manual revision until it met the solvability criteria. Finally, all generated data is subjected to a manual sampling and inspection process to ensure high quality and consistency across the entire benchmark.

\begin{table}[t!]
\centering
\caption{Comparison with existing datasets. Seq, SkilL, LTM, SelfMotivat, Interact, and LfE mean Sequentiality, Skill Learning, Long-Term Memory, Self-Motivation, Interactivity, and Learning from Experience.}
\label{tab: dataset_comparison}
\setlength{\tabcolsep}{1.0mm}
\begin{tabular}{lcccccccccc}
\toprule
Datasets           & Task Type    &    Seq  & SkilL & LTM & SelfMotivat & Real & Interconnected & Interact & LfE \\
\midrule
Lifelong-CIFAR10 \cite{prabhu2024efficient}   & CL       &  \textcolor{green}{\ding{51}}     &   \textcolor{red}{\ding{55}}    &  \textcolor{red}{\ding{55}}    &     \textcolor{red}{\ding{55}}          &    \textcolor{red}{\ding{55}}     &          \textcolor{red}{\ding{55}}      &      \textcolor{red}{\ding{55}}    &      \textcolor{red}{\ding{55}}                       \\
Lifelong-ImageNet \cite{prabhu2024efficient} & CL      &  \textcolor{green}{\ding{51}}        &     \textcolor{red}{\ding{55}}    &  \textcolor{red}{\ding{55}}    &     \textcolor{red}{\ding{55}}          &    \textcolor{red}{\ding{55}}     &          \textcolor{red}{\ding{55}}       &      \textcolor{red}{\ding{55}}   &      \textcolor{red}{\ding{55}}                       \\
CGLB \cite{zhang2022cglb}              & CL     &   \textcolor{green}{\ding{51}}      &  \textcolor{red}{\ding{55}}    &  \textcolor{red}{\ding{55}}    &     \textcolor{red}{\ding{55}}          &    \textcolor{red}{\ding{55}}     &          \textcolor{red}{\ding{55}}       &      \textcolor{red}{\ding{55}}   &      \textcolor{red}{\ding{55}}                   \\
EgoThink \cite{cheng2024egothink}          & Embodied AI  &   \textcolor{red}{\ding{55}}  &     \textcolor{red}{\ding{55}}    &   \textcolor{red}{\ding{55}}    &     \textcolor{red}{\ding{55}}              &     \textcolor{green}{\ding{51}}  &       \textcolor{red}{\ding{55}}          &          \textcolor{red}{\ding{55}}               &       \textcolor{red}{\ding{55}}        \\
EmbodiedBench \cite{yang2025embodiedbench}     & Embodied AI   &   \textcolor{red}{\ding{55}}    &     \textcolor{red}{\ding{55}}     &   \textcolor{red}{\ding{55}}     &           \textcolor{red}{\ding{55}}         &    \textcolor{green}{\ding{51}}     &      \textcolor{red}{\ding{55}}            &         \textcolor{green}{\ding{51}}      &      \textcolor{green}{\ding{51}}               \\
AgentBench \cite{liu2023agentbench}          & Agent &   \textcolor{red}{\ding{55}}    &     \textcolor{red}{\ding{55}}     &   \textcolor{red}{\ding{55}}      &    \textcolor{red}{\ding{55}}                &    \textcolor{red}{\ding{55}}     &      \textcolor{red}{\ding{55}}            &              \textcolor{green}{\ding{51}}           &     \textcolor{red}{\ding{55}}     \\
LoCoMo \cite{maharana2024evaluating}   & Agent &     \textcolor{red}{\ding{55}}     &     \textcolor{red}{\ding{55}}     &    \textcolor{green}{\ding{51}}       &    \textcolor{red}{\ding{55}}                &    \textcolor{red}{\ding{55}}     &      \textcolor{red}{\ding{55}}            &             \textcolor{red}{\ding{55}}        &     \textcolor{red}{\ding{55}}     \\
StoryBench \cite{wan2025storybench}   & Agent &     \textcolor{green}{\ding{51}}      &     \textcolor{red}{\ding{55}}     &    \textcolor{green}{\ding{51}}       &    \textcolor{red}{\ding{55}}                &    \textcolor{red}{\ding{55}}     &      \textcolor{red}{\ding{55}}            &             \textcolor{green}{\ding{51}}         &     \textcolor{red}{\ding{55}}     \\
LifelongAgentBench \cite{zheng2025lifelongagentbench} & Self-Evolving &   \textcolor{green}{\ding{51}}  &    \textcolor{green}{\ding{51}}     &   \textcolor{red}{\ding{55}}    &           \textcolor{red}{\ding{55}}        &    \textcolor{red}{\ding{55}}    &         \textcolor{red}{\ding{55}}        &         \textcolor{green}{\ding{51}}         &       \textcolor{green}{\ding{51}}          \\ \midrule
\texttt{\textbf{StuLife}} (Our)           & ELL       &  \textcolor{green}{\ding{51}}      &  \textcolor{green}{\ding{51}}     &  \textcolor{green}{\ding{51}}    &    \textcolor{green}{\ding{51}}              &  \textcolor{green}{\ding{51}}     &     \textcolor{green}{\ding{51}}           &     \textcolor{green}{\ding{51}}      &     \textcolor{green}{\ding{51}}               \\
\bottomrule
\end{tabular}
\end{table}


\subsection{Comparison with Existing Benchmarks}
In this section, we compare our \texttt{StuLife} benchmark with the existing benchmarks about continual learning, embodied AI, agent, and self-evolving (Table \ref{tab: dataset_comparison}).
\texttt{StuLife} stands out as a comprehensive benchmark for ELL by addressing key limitations of existing datasets. As shown in Table 1, while benchmarks like Lifelong-CIFAR10 and Lifelong-ImageNet \cite{prabhu2024efficient} focus on continuous learning (CL) with sequential task presentation, they lack features such as skill learning, long-term memory, self-motivation, and real-world interactivity. Similarly, CGLB \cite{zhang2022cglb} addresses catastrophic forgetting in graph data but does not incorporate realistic, interconnected tasks or experience-driven learning. Embodied AI benchmarks like EgoThink \cite{cheng2024egothink} and EmbodiedBench \cite{yang2025embodiedbench} introduce dynamic environments and interactive tasks but still fall short in supporting lifelong learning, skill abstraction, and longitudinal growth.
AgentBench \cite{liu2023agentbench} represents a significant advancement by evaluating LLMs as agents across eight interactive environments, emphasizing multi-turn reasoning and decision-making. However, it is primarily designed for static evaluation of agent capabilities at a fixed point in time, rather than tracking continuous growth or self-evolution.
LifelongAgentBench \cite{zheng2025lifelongagentbench} is the first to target self-evolving agents, offering interdependent tasks across diverse domains. However, it primarily focuses on technical environments (e.g., databases, operating systems) and lacks the rich, evolving personal context and intrinsic motivation that characterize natural learning processes.

\texttt{StuLife} is designed as a principled, realistic, and extensible benchmark that advances the evaluation of self-evolving, ELL agents. Unlike conventional benchmarks focused on isolated task performance, \texttt{StuLife} integrates environmental realism, self-evolutionary dynamics, and scalable evaluation into a unified framework. Particularly, it simulates a student’s college journey, featuring interconnected tasks that evolve over time based on dynamic factors such as advisor selection, course availability, and social interactions. This structure enables agents to learn from experience, retain knowledge across tasks, transfer skills, and exhibit self-motivated behavior, all essential components of true lifelong learning. Unlike existing datasets, \texttt{StuLife} provides a holistic evaluation platform that captures the complexity of real-world intelligence, making it an ideal benchmark for advancing self-evolving AI systems.
Below, we highlight its core advantages.

\paragraph{Environmental Realism}
\texttt{StuLife} simulates a full academic term through a longitudinal, narrative-driven structure that mirrors the continuous and evolving nature of real-world learning. Tasks unfold in a temporally coherent sequence, spanning enrollment, coursework, extracurricular engagement, and long-term planning, with rich scene descriptions, character interactions, and situational dialogues. This sequential and interdependent design ensures that early decisions (e.g., course selection, mentor choice) have lasting consequences on later outcomes (e.g., preliminary planning, Academic Activity), fostering cumulative knowledge acquisition and skill transfer. By modeling both academic and social dimensions of student life, \texttt{StuLife} captures the multifaceted challenges of real-world environments, effectively bridging the "sim-to-real" gap. The environment is highly interactive, supporting dynamic agent-object interactions, while promoting open-ended exploration: agents are not constrained by fixed goals but are encouraged to engage in self-directed learning, hallmarks of autonomous intelligence.


\paragraph{Support for Self-Evolving Intelligence}
As shown in Fig~\ref{fig:dataset_descrition}, \texttt{StuLife} is designed to evaluate and foster self-evolving behavior in intelligent agents through three foundational principles:

\textbf{1) From Imitation to Learning}: Tasks are explicitly structured around skill acquisition and reuse, focusing on generalizable competencies such as time management, information retrieval, navigation, and social coordination. Rather than merely replicating patterns, agents must abstract experiences into reusable skills and transfer them across evolving challenges, a hallmark of true lifelong learning.

\textbf{2) From Context to Memory}: The benchmark emphasizes long-term memory and dynamic state evolution, requiring agents to retain and apply knowledge across weeks of simulated academic time. Key contextual variables—such as course eligibility, advisor trust, and scheduling constraints, evolve based on agent decisions, creating a feedback-rich environment that rewards consistency, foresight, and effective knowledge retention over time.

\textbf{3) From Passive to Proactive}: Moving beyond reactive task execution, the narrative-driven design encourages intrinsic motivation and self-directed behavior. Agents must proactively manage deadlines, interpret academic feedback (e.g., responding to performance warnings), reflect on past outcomes, and adjust strategies autonomously—mimicking the initiative and adaptive planning seen in real learners.

Together with dedicated metrics for memory utilization, skill acquisition, and cross-task generalization, \texttt{StuLife} provides a comprehensive framework for evaluating not just task performance, but the deeper cognitive growth and autonomous development necessary for self-evolving AI.
In summary, \texttt{StuLife} transcends traditional benchmarks by embedding lifelong learning within a realistic, evolving, and narratively grounded environment. It uniquely supports the development and evaluation of agents that do not merely perform tasks, but learn from experience, grow over time, and act with autonomy, making it a foundational platform for the next generation of self-evolving AI. 

\subsection{Future Directions for \texttt{StuLife}}
To further enhance the realism, scalability, and long-term relevance of \texttt{StuLife} as a platform for evaluating self-evolving agents, we outline several key directions for future development:
\begin{itemize}[leftmargin=*, align=left]
    \item \textbf{Integration of More Complex Tools}: In future versions, agents will be required to interact with increasingly sophisticated tools—such as code interpreters, database query systems, calendar schedulers with conflict detection, and email clients with threading logic. This will elevate the cognitive and procedural demands on agents, pushing them beyond simple API calls toward robust tool mastery and multi-step workflow automation.
    \item \textbf{Modeling Strong Task Interdependencies}: We plan to introduce deeper structural dependencies between tasks, particularly in academic progression. For example, courses will be organized in prerequisite chains (e.g., "Introduction to Algorithms" must be passed before enrolling in "Advanced Data Structures"), and performance in early tasks will directly influence access to advanced opportunities (e.g., research positions or honors programs). These dependencies will enforce long-term planning and reward consistent knowledge accumulation.
    \item \textbf{Dynamic and Flexible Rule Evolution}: To better simulate real-world institutional environments, the benchmark will support runtime updates to rules and policies (e.g., changes in graduation requirements, course availability, or academic regulations). This capability will allow the environment to evolve over time, preventing agents from overfitting to static conditions and encouraging adaptive, resilient behavior in the face of change.
    \item \textbf{Increased Task Complexity to Prevent Exploitation}: We will design tasks that inherently resist shortcut solutions or prompt-based memorization, ensuring that high performance requires genuine understanding, reasoning, and experience accumulation. By incorporating open-ended problem-solving, partial observability, and stochastic outcomes, we aim to discourage "leaderboard hacking" and promote robust, generalizable intelligence.
    \item \textbf{Development of a General-Purpose Benchmarking Framework}: We are building a modular framework to enable rapid adaptation of the \texttt{StuLife} paradigm to other domains, such as workplace onboarding, healthcare management, internships, research projects, and career planning. This framework will support plug-and-play task modules, configurable rule engines, and standardized evaluation interfaces, empowering researchers to create domain-specific benchmarks while maintaining compatibility with the core ELL evaluation metrics.
\end{itemize}

\section{Experimental Analysis}
\subsection{Defining Evaluation Metrics}
To provide a multi-faceted assessment of agent performance on \texttt{StuLife}, we establish a suite of evaluation metrics. These are designed to measure not only task success but also the nuances of learning, the durability of knowledge, and the efficiency of agent behavior. The metrics are ordered to provide a comprehensive analysis, starting from a holistic, student-centric score and progressively delving into more specific aspects of agent intelligence.

\begin{itemize}[leftmargin=*, align=left]

    \item \textbf{StuGPA (General Performance Assessment)}: To evaluate agents in a manner that mirrors real-world student assessment, we introduce the \textbf{StuGPA}. This composite score, calculated out of 100, moves beyond simple task accuracy to holistically measure an agent's performance across dimensions of academic and personal development. This score is designed to reflect the holistic evaluation criteria used in actual educational settings, where success is not determined solely by exam results, but also by consistent engagement, responsible behavior, and classroom participation.
    The GPA is computed as a weighted sum of three core components:
    \begin{itemize}[leftmargin=*, align=left]
        \item \textbf{Exam Performance (50 Points)}: This is the cornerstone of the GPA, assessing the agent's knowledge mastery through its average accuracy on the \textit{Midterm and Final Exams}. This component directly measures the outcomes of long-term learning and knowledge consolidation under high-stakes conditions.
        \item \textbf{Class Performance (30 Points)}: This component evaluates the agent's commitment and adherence to academic routines. It is calculated based on the attendance rate for all required \textit{In-Class Tasks}. This metric deliberately focuses on the process, being at the right place at the right time, rather than the correctness of in-class answers, which is already assessed in exams.
        \item \textbf{Campus Daily Life (20 Points)}: This evaluates the agent as a "campus citizen." It is composed of advisor task completion (8 points), Club  Activity(6 points), and a \textit{Personal Responsibility} score (6 points). The responsibility score begins at 6 and is reduced for infractions such as \textit{resource squandering} (e.g., booking a library seat but not using it) or \textit{broken commitments} (e.g., missing a self-scheduled meeting).
    \end{itemize}
    \item \textbf{Long-Term Retention Rate (LTRR)}: This evaluates the agent's capacity to combat catastrophic forgetting. It is calculated as the success rate on tasks requiring knowledge from the past (e.g., information presented over a week ago, such as in midterm or final exams). These tasks require the recall and synthesis of information presented prior, providing a clear and stringent test of long-term memory. A high LTRR signifies a robust memory system capable of retaining and accessing critical information over time.
    \item \textbf{Proactive Initiative Score (PIS)}: This metric assesses an agent's prospective memory and self-motivation, its ability to remember and execute planned intentions at the appropriate time without an immediate prompt. It is measured as the success rate on tasks that must be self-initiated based on a previously established schedule (e.g., attending weekly \textit{Core Course Instruction} at the correct time and location). Success in these tasks demonstrates that the agent can retain temporal commitments and autonomously act upon them, a crucial skill for maintaining long-term plans and routines in lifelong learning scenarios.
    \item \textbf{Success Rate}: This is the primary metric for task completion. We measure the overall success rate, defined as the proportion of tasks an agent successfully completes across the entire benchmark (Success). A task is considered successful only if the agent achieves the final objective. To offer a more granular view of an agent's capabilities, we also report the accuracy for each of the three core modules: \textit{In-Class}, \textit{Daily Campus}, and \textit{Examination Tasks}.
    \item \textbf{Average Turns}: We measure interaction efficiency by calculating the average number of interaction turns required to complete a task successfully, denoted as \textbf{AvgTurns}. It sums the number of interactions for all successful trajectories and dividing by the total number of successfully completed tasks. A lower AvgTurns value signifies higher efficiency and more sophisticated problem-solving, as the agent achieves its goals with fewer steps.
\end{itemize}

\begin{table}[t!]
\centering
\caption{The performance of existing SOTA LLMs (Default Setting). Icons denote model properties: \hficon~Open-source, \brainicon~Thinking model. The LTRR, PIS, and Success Rate, are presented in percentages. 
}
\label{tab:main_results}
\resizebox{\textwidth}{!}{%
\setlength{\tabcolsep}{1.0mm}
\begin{tabular}{lccccccccccc}
\toprule
& \multirow{2}{*}{StuGPA}  & \multirow{2}{*}{LTRR}  & \multirow{2}{*}{PIS} & \multicolumn{2}{c}{In-Class} & \multicolumn{2}{c}{Daily Campus} & \multicolumn{2}{c}{Exam} & \multicolumn{2}{c}{Total} \\
&         &         &     & Success         & AvgTurn         & Success         & AvgTurn         & Success         & AvgTurn   & Success         & AvgTurn         \\ \midrule
Llama-3.1-8B\textsuperscript{\hficon} & 5.81 & 3.30 & 0.90 & 0.90 & 61.34 & 0.00 & 35.91 & 10.63 & 28.46 & 2.13 & 44.62 \\
Qwen3-8B\textsuperscript{\hficon} & 13.31 & 4.33 & 0.54 & 0.90 & 10.12 & 8.31 & 10.25 & 14.38 & 6.31 & 6.71 & 9.71 \\
Qwen3-30B-A3B\textsuperscript{\hficon \hspace{0.01mm} \brainicon} & 16.30 & 5.05 & 0.72 & 0.60 & 9.45 & 10.79 & 11.75 & 17.50 & 5.46 & 8.31 & 10.09 \\
Qwen3-32B\textsuperscript{\hficon} & 7.36 & 3.97 & 0.54 & 0.60 & 7.80 & 2.25 & 13.79 & 13.13 & 4.88 & 3.51 & 10.41 \\
Qwen3-32B\textsuperscript{\hficon \hspace{0.01mm} \brainicon} & 12.67 & 5.42 & 1.26 & 1.80 & 8.31 & 7.64 & 10.74 & 17.50 & 4.94 & 7.24 & 9.10 \\
QwQ-32B\textsuperscript{\hficon} & 13.21 & 5.78 & 3.42 & 4.79 & 7.72 & 6.97 & 13.25 & 16.88 & 4.52 & 7.88 & 10.06 \\
DeepSeek-V3\textsuperscript{\hficon} & 11.22 & 6.14 & 2.88 & 3.59 & 5.84 & 6.74 & 11.87 & 16.25 & 4.26 & 7.24 & 8.64 \\
DeepSeek-R1\textsuperscript{\hficon \hspace{0.01mm} \brainicon} & 14.25 & 8.30 & 3.96 & 5.09 & 8.04 & 13.26 & 13.02 & 18.13 & 4.56 & 11.18 & 10.08 \\
DeepSeek-V3.1\textsuperscript{\hficon} & 14.26 & 4.51 & 0.54 & 0.90 & 14.03 & 12.81 & 12.62 & 15.00 & 6.78 & 8.95 & 12.43 \\
DeepSeek-V3.1\textsuperscript{\hficon \hspace{0.01mm} \brainicon} & 17.04 & 6.14 & 3.78 & 6.29 & 9.83 & 12.58 & 13.03 & 17.50 & 5.54 & 11.18 & 10.88 \\
Qwen3-235B-A22B\textsuperscript{\hficon} & 16.03 & 5.42 & 1.80 & 2.10 & 18.71 & 10.34 & 17.17 & 16.88 & 10.75 & 8.52 & 16.95 \\
Gemini-2.5-Pro\textsuperscript{\hficon \hspace{0.01mm} \brainicon} & 16.43 & 7.04 & 3.24 & 5.39 & 14.94 & 18.88 & 12.78 & 15.63 & 9.51 & 13.53 & 13.19 \\
Grok4\textsuperscript{\brainicon} & 17.38 & 10.65 & 4.50 & 4.79 & 6.31 & 21.80 & 11.25 & 18.75 & 5.69 & 15.23 & 8.68 \\
GPT-5\textsuperscript{\brainicon} & 17.90 & 6.50 & 4.68 & 7.78 & 12.70 & 14.16 & 14.31 & 16.88 & 6.24 & 12.35 & 12.69 \\
\midrule
Human&85.24&84.91&88.13&88.62&-&91.13&-&82.50&-&88.81&-\\
\bottomrule
\end{tabular}
}
\end{table}

\subsection{Evaluating the Existing SOTA LLMs}
\subsubsection{Experimental Setup}
\paragraph{Models Evaluated}
We assess a diverse set of ten prominent Large Language Models (LLMs) as the cognitive engines for our intelligent agents. The models under study include: Llama-3.1-8B\footnote{\url{https://ai.meta.com/research/publications/the-llama-3-herd-of-models/}}, Qwen3-7B \cite{yang2025qwen3}, Qwen3-32B \cite{yang2025qwen3}, QWQ-32B~\cite{qwen2.5}, Deepseek-V3 \cite{liu2024deepseek}, Qwen3-235B \cite{yang2025qwen3}, GPT-5\footnote{\url{https://openai.com/index/introducing-gpt-5/}}, Gemini 2.5 Pro\footnote{\url{https://deepmind.google/models/gemini/pro/}}, and Grok-4\footnote{\url{https://x.ai/news/grok-4}}. These models represent the current frontier in understanding, reasoning, and agentic capabilities. 

\paragraph{Evaluation Protocol}
The \texttt{StuLife} benchmark is designed as a single, continuous trajectory where tasks are presented serially, and the environment is stateful—meaning an agent's actions in one task can have persistent consequences that affect subsequent tasks. In our primary evaluation protocol, each task is presented to the agent as an independent, isolated instance. The agent operates without access to any historical context or memory from previous interactions, any cross-task information retention relies entirely on the agent's explicit actions to save and later retrieve data using the provided environmental tools, such as a calendar, or draft. This stateless approach serves as a crucial baseline to measure the models' intrinsic, in-context reasoning and problem-solving abilities without the aid of accumulated experience.
All experiments are conducted on Linux servers, and our framework supports automatic checkpointing to ensure robustness and allow for recovery from interruptions. To ensure a fair and reproducible comparison, all agents are integrated through APIs and share a standardized initialization process. Additionally, we recruit several undergraduate and graduate students to establish a human baseline for comparison.

\textbf{RQ I}: How do current LLMs perform on long-horizon agency tasks, and how does performance vary with model scale and type?

\textbf{Finding 1}: Scaling up parameters isn't enough—what AI really needs is to grow up, not just level up. The path to AGI may not lie in bigger models, but in building agents that learn how to learn, not just answer questions.

As shown in Table~\ref{tab:main_results}, the performance of all models under our default stateless setting is exceptionally poor. This stems from a fundamental architectural limitation: current large language models are inherently stateless and do not possess any native long-term memory modules. This core deficiency manifests in two critical failures observed throughout our experiments. 

\textbf{First,} our results reveal that all current agents, regardless of whether they are open-source or proprietary, exhibit mediocre performance and are far from achieving human-like competency in long-horizon tasks. As shown in Table~\ref{tab:main_results}, even the most advanced model, \texttt{GPT-5}, achieves a StuGPA of only 17.90 out of 100, underscoring the significant gap between current SOTA models and the general, autonomous intelligence required for such challenges. When contrasted with the human StuGPA of 85.24, this gap becomes even more glaring, illustrating that current LLMs have yet to bridge the vast divide separating statistical pattern matching from human-level functional competence. \textbf{Second}, there is a strong positive correlation with \textbf{model scale}. A significant performance chasm exists between small models like \texttt{Llama-3.1-8B-Instruct} (StuGPA of 5.81) and the best-performing large-scale models. The progression from 8B models to 32B, 235B, and proprietary models generally yields higher scores, indicating that larger parameter counts do enhance the complex reasoning required for these tasks.
\textbf{Third}, and more revealingly, is the impact of \textbf{model type}. A direct comparison within the same size class shows that models designed for step-by-step reasoning significantly outperform their standard instruction-following counterparts. For example, \texttt{Qwen3-32B-Thinking} achieves a StuGPA of 12.67, substantially higher than the 7.36 scored by \texttt{Qwen3-32B-Instruct}. The "Thinking" model also demonstrates better proactivity, with a Proactive Initiative Score (PIS) of 1.26 compared to the Instruct model's 0.54. This suggests that for complex agency tasks that require planning and foresight, a model's internal architecture designed to support deliberative reasoning is a more critical determinant of success than its raw instruction-following ability alone.

\textbf{RQ II}: Can stateless LLMs manage long-term memory and proactive behavior?

\textbf{Finding 2}: No memory by design, no motivation by default: stateless LLMs cannot grow through experience.

The evidence for this finding, detailed in Table~\ref{tab:main_results}, is stark and can be broken down into two key areas of failure.
\textbf{First,} proactive, self-motivated behavior is arguably the greatest challenge for current agents. This is demonstrated by a common failure mode: for a task scheduled at 8:00, even when an agent is notified that the current time is precisely 8:00, it often fails to initiate the corresponding action. This issue stems from a more fundamental problem where the agent does not even think to consult its schedule, regardless of whether the event was successfully added previously. This qualitative failure is directly quantified by the Proactive Initiative Scores (PIS), which are critically low across the board. Even the top-performing model, \texttt{GPT-5}, only achieves a PIS of 4.68\%, while many powerful models like \texttt{Qwen3-8B} and \texttt{DeepSeek-V3.1-chat} score below 1\%. While a reasoning-focused "Thinking" architecture offers a slight advantage (\texttt{Qwen3-32B-Thinking} PIS of 1.26\% vs. its "Instruct" counterpart's 0.54\%), it does not overcome this fundamental lack of self-motivation. This stands in sharp relief against the human PIS of 88.13\%, revealing that while models may possess vast knowledge, they completely lack the intrinsic drive and autonomy that define human agency. \textbf{Second,} the widespread failure in long-term retention is a direct and expected consequence of the fact that \textbf{all current LLMs are architecturally stateless}. In this context, the LTRR metric does not measure an LLM's intrinsic memory. Instead, it evaluates the success rate of a fragile, multi-step, tool-dependent process: the agent must first recognize critical information, then correctly use a tool to externalize it (e.g., adding an event to the calendar), and finally, think to retrieve it for a future exam. Therefore, the slightly higher LTRR scores from top models like \texttt{Grok4} (10.65\%) do not indicate superior intrinsic memory. Rather, they reflect a higher probability of having made correct tool-use decisions in prerequisite tasks. The fact that even the best models fail 90\% of the time on this metric underscores the brittleness of this process and confirms that reliable knowledge accumulation is currently unattainable without a dedicated memory system. The profound gap with the human LTRR of 84.91\% indicates that current architectures lack the cognitive continuity required to preserve critical information over time, a fundamental trait of human intelligence that stateless models fail to replicate.

\begin{table}[t!]
\centering
\small
\caption{Performance of Self-Evolving mechanisms across different model scales. The 8B model utilizes training-based evolution (RFT), while the 235B model utilizes inference-time evolution (AWM, Reflexion).}
\label{tab:self_evolving_results}
\setlength{\tabcolsep}{1.0mm}
\begin{tabular}{llccccccccccc}
\toprule
 & & \multirow{2}{*}{StuGPA} & \multirow{2}{*}{LTRR} & \multirow{2}{*}{PIS} & \multicolumn{2}{c}{In-Class} & \multicolumn{2}{c}{Daily Campus} & \multicolumn{2}{c}{Exam} & \multicolumn{2}{c}{Total} \\
 & & & & & {\scriptsize Success} & {\scriptsize AvgTurn} & {\scriptsize Success} & {\scriptsize AvgTurn} & {\scriptsize Success} & {\scriptsize AvgTurn} & {\scriptsize Success} & {\scriptsize AvgTurn} \\ \midrule
\multicolumn{13}{l}{\textbf{Qwen3-8B\textsuperscript{\hficon}}} \\
\midrule
\multicolumn{2}{l}{Vanilla} & 13.31 & 4.33 & 0.54 & 0.90 & \textbf{10.12} & 8.31 & 10.25 & \textbf{14.38} & 6.31 & 6.71 & \textbf{9.71} \\
& + RFT & \textbf{15.43} & \textbf{4.51} & \textbf{0.90} & \textbf{1.50} & 12.08 & \textbf{11.91} & \textbf{9.43} & \textbf{14.38} & \textbf{5.05} & \textbf{8.63} & 9.89 \\
\midrule
\multicolumn{13}{l}{\textbf{Qwen3-235B-A22B}\textsuperscript{\hficon}} \\
\midrule
\multicolumn{2}{l}{Vanilla} & 16.03 & 5.42 & 1.80 & 2.10 & 18.71 & 10.34 & 17.17 & \textbf{16.88} & 10.75 & 8.52 & 16.95 \\
& + AWM & \textbf{17.81} & \textbf{6.87} & \textbf{1.80} & \textbf{1.50} & \textbf{15.66} & \textbf{14.38} & \textbf{14.23} & 16.25 & \textbf{7.68} & \textbf{10.12} & \textbf{13.96} \\
& + Reflexion & 16.18 & 5.60 & 1.44 & 0.90 & \textbf{15.66} & 12.13 & 15.05 & \textbf{16.88} & 9.08 & 8.95 & 14.54 \\
\bottomrule
\end{tabular}
\vspace{-3mm}
\end{table}

\textbf{RQ III}: To what extent can existing self-evolving mechanisms address the challenges of lifelong learning?

\textbf{Finding 3}: Self-evolving mechanisms provide tangible benefits in both learnability and efficiency. However, these gains remain bounded, as existing strategies have yet to bridge the gap required for proactive initiative and sustained long-term memory retention.

To demonstrate the flexibility of our benchmark in evaluating experience-driven improvement, we investigate two distinct categories of self-evolving strategies across different model scales. For training-based evolution, we employ an on-policy Rejection Sampling Fine-Tuning (RFT) \citep{yuan2023scalingrelationshiplearningmathematical} on the Qwen3-8B model, where the model generates eight rollouts per prompt during exploration and is subsequently fine-tuned using only the successful trajectories. For inference-time evolution, we evaluate Reflexion \citep{shinn2023reflexion} and Agent Workflow Memory (AWM) \citep{wang2024agent} on the larger Qwen3-235B-A22B model. These methods leverage context and historical feedback to dynamically optimize future actions without updating model weights.

\textbf{Analysis of Self-Evolving Approaches.}
The results in Table \ref{tab:self_evolving_results} highlight the distinct advantages offered by different evolutionary paths.
For the smaller model, RFT demonstrates strong data efficiency and learnability, significantly boosting Qwen3-8B's StuGPA from 13.31 to 15.43 and Total Success from 6.71\% to 8.63\%. This improvement confirms that the complex decision-making logic within our benchmark follows learnable patterns that can be effectively distilled into smaller models through high-quality experience.
For the larger model, inference-time strategies also yield performance gains. AWM achieves a robust improvement, lifting StuGPA to 17.81 and Total Success to 10.12\%, while Reflexion shows marginal gains. Crucially, AWM stands out by simultaneously optimizing efficiency, reducing the Total AvgTurn from 16.95 to 13.96. This suggests that retrieving successful workflows not only aids in task completion but also helps the agent avoid redundant steps. 
Nevertheless, despite these improvements across both paradigms, the gains remain modest relative to the complexity of the lifelong learning setting. This validates the difficulty of our benchmark: while agents can evolve to become more accurate or efficient, existing mechanisms have yet to bridge the gap required for proactive initiative and sustained cognitive coherence.

\textbf{RQ IV}: Can Context Engineering mitigate the LLMs' limitations in lifelong learning scenarios?

\textbf{Finding 4}: Context Engineering isn’t just programming—it’s propping up intelligence, a critical pathway toward more general and autonomous intelligence.

We also make a critical contribution by systematically investigating the impact of \textit{context engineering}, on agent performance in complex and long-horizon tasks. While much of current AI development focuses on scaling models or improving weights through training, our results demonstrate that \textit{optimizing the context prompt} can also improve performance effectively. 
Particularly, we design five distinct prompting strategies to investigate different dimensions of agent intelligence in a simulated academic environment:
(1) Vanilla Prompt: A minimal baseline assigning only a high-level role, measuring the agent’s intrinsic reasoning without procedural or cognitive support.
(2) Proactive Prompt: Adds structured guidance on time-awareness, goal decomposition, and forward planning to emulate strategic, student-like behavior.
(3) Skill-Augmented Prompt: Provides step-by-step "recipes" for specific skills (e.g., problem-solving), enabling structured and generalizable task decomposition.
(4) Memory-Augmented Prompt: Integrates external memory mechanisms (e.g., Vanilla RAG \cite{lewis2020retrieval}, GraphRAG \citep{edge2024local}, MemoryBank \citep{zhong2024memorybank}, MemGPT \citep{packer2023memgpt}) to simulate cumulative learning and assess long-term knowledge retention.
(5) All-in-One Prompt: Unifies proactive planning, skill execution, and memory augmentation into a comprehensive cognitive framework, aiming to evaluate the upper bound of self-evolving intelligence.
\begin{table}[t!]
\vspace{-4mm}
\centering
\small
\caption{The performance of various Context Engineering methods based on Qwen3-235B-A22B\textsuperscript{\hficon}.}
\label{tab:context_results}
\setlength{\tabcolsep}{0.5mm}
\begin{tabular}{llccccccccccc}
\toprule
       &   & \multirow{2}{*}{StuGPA}  & \multirow{2}{*}{LTRR}  & \multirow{2}{*}{PIS} & \multicolumn{2}{c}{In-Class} & \multicolumn{2}{c}{Daily Campus} & \multicolumn{2}{c}{Exam} & \multicolumn{2}{c}{Total} \\
       &   &        &        &             & {\scriptsize Success}      & {\scriptsize AvgTurn}      & {\scriptsize Success}         & {\scriptsize AvgTurn}         & {\scriptsize Success}    & {\scriptsize AvgTurn}   & {\scriptsize Success}    & {\scriptsize AvgTurn}    \\ \midrule
\multicolumn{2}{l}{\textbf{Vanilla}}  & 16.03 & 5.42 & 1.80 & 2.10 & 18.71 & 10.34 & 17.17 & 16.88 & 10.75 & 8.52 & 16.95 \\
 \midrule
\multicolumn{2}{l}{\textbf{Proactive}}  & 16.90 & 5.96 & 3.06 & \textbf{5.09} & \textbf{16.70} & 10.34 & 16.38 & 16.88 & 7.73 & 9.58 & 15.42 \\
\midrule
\multicolumn{2}{l}{\textbf{Skill}}  & 17.28 & 6.86 & 0.90 & 1.50 & 16.89 & 15.28 & 16.51 & 17.50 & 9.28 & 10.76 & 15.75 \\
 \midrule
\multirow{4}{*}{\begin{tabular}[c]{@{}l@{}}\textbf{Memory}\end{tabular}} 
& + Vanilla RAG & 10.98 & 4.69 & 0.18 & 0.00 & 17.87 & 5.84 & 14.20 & 16.25 & 10.04 & 5.54 & 15.07 \\
& + Graph RAG & 15.34 & 4.87 & 0.72 & 0.90 & 20.68 & 10.11 & 14.03 & 16.25 & 10.61 & 7.88 & 16.13 \\
& + MemGPT & 19.99 & 6.86 & 1.44 & 2.40 & 17.28 & 13.03 & \textbf{13.59} & 23.75 & 9.02 & 11.08 & 14.42 \\
& + MemoryBank & 17.64 & 5.96 & 1.62 & 0.90 & 16.68 & 12.36 & 14.15 & 20.00 & 8.04 & 9.58 & \textbf{14.35} \\
\midrule
\multicolumn{2}{l}{\textbf{All-in-One}}  & \textbf{21.07} & \textbf{9.39} & \textbf{3.76} & 2.69 & 16.82 & \textbf{17.75} & 15.65 & \textbf{25.63} & \textbf{6.30} & \textbf{13.74} & 14.93 \\

\bottomrule
\end{tabular}
\vspace{-3mm}
\end{table}

\textbf{Analysis of Proactive and Skill Prompts.}
Explicit cognitive prompts delivered targeted gains. The \textbf{Proactive Prompt}, which instills time management principles, sharply improved time-sensitive performance: In-Class success more than doubled (2.10\% to 5.09\%), and PIS rose significantly (1.80\% to 3.06\%). Yet, because it doesn't teach \textit{how} to break down or execute complex tasks and still depends on fragile tool-based memory, the overall StuGPA gain was modest (16.03 to 16.90). In contrast, the \textbf{Skill-Augmented Prompt}, which supplies a step-by-step problem-solving framework, greatly increased success on complex Daily Campus tasks (10.34\% to 15.28\%). But by focusing solely on execution, it neglects initiative, causing PIS to plummet (1.80\% to 0.90\%).

\textbf{Analysis of Memory-Augmented Approaches.}  
Adding external memory produced sharply different outcomes, highlighting that memory system design is critical. Naive RAG methods (e.g., \textbf{Vanilla RAG} and \textbf{Graph RAG}), which retrieve raw trajectories via vector similarity and inject them directly into context, actively harmed performance. They worsened the LLM’s long-context degradation by introducing unfiltered noise, collapsing scores (e.g., Vanilla RAG StuGPA: 10.98) and even reducing LTRR below baseline (5.42\% to 4.69\%).
In contrast, \textbf{MemGPT}, a sophisticated memory system that structures and manages stored information, achieved the highest overall StuGPA (19.99). Its reliable recall improved Examination success (23.75\%). Yet, despite its strong reactive memory, MemGPT cannot compensate for the agent’s lack of time awareness.

\textbf{The Power of the All-in-One Approach.}
Our analysis shows each strategy tackles a unique bottleneck: Proactive prompts teach ``when," Skill prompts teach ``how," and Memory systems supply the ``what." The results show that \textbf{All-in-One Prompt} outperforms all others, achieving the highest StuGPA (21.07, surpassing GPT-5), PIS (3.76\%) and Daily Campus success (17.75\%). The LTRR soars to 9.39\%, proving that combining memory with proactivity enables the agent to both remember and act on long-term goals. This confirms our core thesis: true autonomous agency emerges not from isolated components, but from the integrated synergy of proactivity, skill, and structured memory.

\section{Conclusions}
In this paper, we introduced Experience-driven Lifelong Learning (ELL), a framework for building self-evolving AI agents that learn and grow through continuous interaction with dynamic environments. Inspired by human cognitive development, ELL emphasizes three core principles: learning from experience, long-term memory integration, and skill abstraction, enabling agents not only to adapt to change but to proactively shape their own learning trajectories. To advance research in this direction, we present \texttt{StuLife}, a comprehensive benchmark that simulates a student’s college journey as a rich, longitudinal environment with interconnected academic, social, and planning tasks.
\texttt{StuLife} goes beyond conventional benchmarks by incorporating evolving contexts, sparse rewards, and the need for self-initiated behavior, making it a powerful platform for evaluating lifelong learning, memory retention, and autonomous decision-making. Our experiments reveal critical limitations in current LLM-based agents, particularly their lack of self-motivation and poor long-term memory, highlighting the gap between today’s systems and truly self-evolving intelligence. Importantly, we demonstrate that performance on complex, extended tasks can be significantly improved not only through model scale but through \textit{context engineering}, particularly via well-designed system prompts. This suggests that the path to AGI may depend as much on how we structure an agent’s cognitive framework as on the raw capabilities of the underlying model.
Looking forward, ELL and \texttt{StuLife} provide a principled foundation for developing agents that accumulate knowledge, transfer skills, and evolve autonomously over time. By shifting the focus from isolated task performance to sustained, experience-driven growth, we aim to catalyze progress toward intelligent systems that, like humans, become wiser with every experience.


\bibliographystyle{unsrt}  
\bibliography{references}  

\clearpage
\newpage
\appendix
\section{Benchmark Environment: System Architecture}
\label{sec:appendix_system_architecture}

To rigorously evaluate the multifaceted capabilities of AI agents, we designed and implemented \textit{StuLife Bench}, a novel benchmark environment simulating a university campus. The environment is engineered as a deterministic, persistent, and stateful world, compelling the agent to engage in complex information integration, strategic planning, and multi-step task execution. Its architecture is founded on several core principles designed to ensure reproducibility, task complexity, and fair evaluation.

\subsection{Core Architectural Principles}

The design of \textit{StuLife Bench} is predicated on a clear separation of concerns, distinguishing the world's state and mechanics from the tasks an agent must perform.

\begin{itemize}[leftmargin=*, align=left]
    \item \textbf{Persistent World State:} The environment is instantiated as a single, centralized, and stateful object that persists across the entire lifecycle of an agent's evaluation. This ensures that actions taken in one task have lasting consequences on the world state, creating longitudinal dependencies and requiring the agent to maintain a coherent long-term strategy. For instance, a course registered in the first week remains on the agent's schedule for all subsequent tasks.

    \item \textbf{Deterministic Subsystems:} All components of the environment operate on rule-based, deterministic logic. Randomness is explicitly excluded from the simulation's mechanics to guarantee that for a given sequence of agent actions, the resulting state transitions and outcomes are always identical. This determinism is crucial for the reproducibility of experiments and the objective comparison of different agents.

    \item \textbf{Separation of Environment and Task Logic:} The architecture strictly separates the \textit{Environment}, which simulates the world and its atomic physics (e.g., moving between locations, sending an email), from the \textit{Task Controller}. The Task Controller is responsible for presenting natural language instructions, mediating all agent-environment interactions by dispatching agent-invoked tools to the Environment, and finally, evaluating the agent's performance by comparing the final state of the Environment against a task-specific ground truth.
\end{itemize}

\subsection{World State and Temporal Dynamics}
The simulation of time and its effect on the agent's schedule is a passive, event-driven process that provides essential context for tasks.

\begin{itemize}[leftmargin=*, align=left]
    \item \textbf{World Time System:} The agent cannot directly control or query the flow of time. Instead, the system injects temporal context into the agent's observation space at the beginning of tasks. This is achieved through system announcements (e.g., ''System Announcement: Today is the Saturday of the second week.'') and time-specific prompts (e.g., ''System Prompt: It is now 8:00 AM.''). This mechanism serves to trigger time-sensitive tasks and test the agent's ability to react to temporal cues.

    \item \textbf{Calendar System:} The environment maintains a persistent, multi-identity calendar system. The agent can manage its personal schedule, contribute to shared schedules (e.g., for a student club), and query the availability of others (e.g., an advisor). The system enforces a differentiated permission model: full create, read, update, delete (CRUD) operations on the personal calendar, append-only access to club calendars, and read-only (busy/free) access to advisor schedules.
\end{itemize}

\subsection{Spatial and Geographic Simulation}
The agent's interaction with the physical campus is mediated by a dual-component system that separates spatial knowledge from physical action.

\begin{itemize}[leftmargin=*, align=left]
    \item \textbf{Map Lookup System:} A static, read-only information provider that contains the complete geographical data of the campus, including buildings, rooms, and their properties. It exposes tools for the agent to find building IDs from names, retrieve detailed location information, and, crucially, compute deterministic optimal paths between any two points, subject to specified constraints.

    \item \textbf{Geography System:} A dynamic state tracker that maintains the agent's current physical location. To change its position, the agent must first use the Map Lookup System to plan a path and then pass the computed path to a specific tool in the Geography System to execute the movement. This two-step process explicitly models the separation of planning from execution. The agent's location is automatically reset to a default starting point (e.g., a dormitory) at the beginning of each simulated day.
\end{itemize}

\subsection{Academic Course Selection System}
This subsystem simulates the complex, high-stakes process of university course registration, designed to test strategic resource allocation under constraints. The core mechanic is a weighted selection process.

\begin{itemize}[leftmargin=*, align=left]
    \item \textbf{Stateful Planning and Registration:} The agent first formulates a preliminary plan by adding courses to a draft schedule. This plan can be modified freely. The final registration is a distinct, atomic action where the draft is submitted for processing.

    \item \textbf{Priority Pass Mechanism:} Success in registration is determined by a set of deterministic rules based on a course's dynamic \texttt{popularity\_index} and a limited set of ''Priority Passes'' (\texttt{S-Pass}, \texttt{A-Pass}, \texttt{B-Pass}) that the agent can assign to courses in its draft. For example, an \texttt{A-Pass} guarantees enrollment if the course popularity is below 95, while a \texttt{B-Pass} succeeds only if popularity is below 85. The agent must strategically use these passes on high-demand courses to ensure successful registration. The popularity and seat availability of courses evolve between tasks, requiring the agent to continuously adapt its strategy.
\end{itemize}

\subsection{Resource Reservation System}
This system manages the booking of shared campus facilities, such as library study rooms and seminar halls. Its most notable feature is a mechanism for dynamically generating availability to create well-posed decision-making puzzles.

\begin{itemize}[leftmargin=*, align=left]
    \item \textbf{Intelligent Availability Generation:} When an agent queries for available slots at a location relevant to its current task, the system does not return a static, pre-defined list. Instead, it dynamically generates a set of plausible options based on the task's specific constraints and ground truth. It reverse-engineers the availability list to ensure that only the ground-truth option satisfies all explicit and implicit requirements of the task, while presenting other options as meaningful distractors. For all other queries not central to the task, availability is generated randomly while respecting existing bookings. This design ensures that each reservation task is a solvable, self-contained puzzle.
\end{itemize}

\subsection{Information Retrieval Systems}
The agent accesses static world knowledge through a set of read-only query tools, which are divided into two distinct structural paradigms.

\begin{itemize}[leftmargin=*, align=left]
    \item \textbf{Hierarchical Bibliography System:} This system contains academic texts organized in a strict, four-level hierarchy: Book \(\rightarrow\) Chapter \(\rightarrow\) Section \(\rightarrow\) Article. To access a specific piece of information, the agent must perform a sequence of iterative, drill-down queries, navigating the hierarchy level by level.

    \item \textbf{Entity-Based Campus Data System:} In contrast, information about campus entities like student clubs and academic advisors is stored in a flat, entity-based structure. This system supports direct queries by category (e.g., listing all sports clubs) or by a unique identifier (e.g., retrieving the full profile of a specific advisor), testing the agent's ability to select the appropriate query strategy for different data structures.
\end{itemize}

\subsection{Communication System}
To assess the agent's ability to comprehend instructions and structure information for communication, a basic email system is provided.

\begin{itemize}[leftmargin=*, align=left]
    \item \textbf{Append-Only Log:} The system does not simulate a real email network but rather maintains a persistent, append-only log of all emails the agent sends.

    \item \textbf{Strict-Matching Evaluation:} A task requiring the agent to send an email is evaluated based on a strict, verbatim string match of the recipient, subject, and body fields against the ground truth. This rigorously tests the agent's capacity to extract key information from a natural language prompt and format it precisely according to the tool's requirements.
\end{itemize}

\section{Tool Suite and Agent Instructions}
\label{sec:appendix_tools}

To interact with the \textit{StuLife Bench} environment, the agent is provided with a comprehensive suite of tools. This section details the complete action space available to the agent, which is structured as a set of functions grouped by their corresponding subsystems. We first present the foundational instructions that govern the agent's behavior and response format, followed by the detailed declarations for each tool.

\subsection{Base Instructions for the Agent}
Before beginning any task, the agent is initialized with a set of base instructions that define its role, objectives, and the required format for all actions. This ensures a consistent interaction protocol across all evaluations.

\begin{textBox}[Foundational Agent Instructions]
You are an AI agent acting as a student in a university campus environment. Your goal is to complete the tasks given to you by using a set of available tools to interact with this world.

At each step, you will be given an observation of the current state of the environment. Instructions using the first-person pronoun "I" represent your own internal thoughts at that moment, which you should act upon accordingly.

You have access to a variety of tools to help you. You must go to the correct location at the correct time to execute tasks. When you believe you have completed ALL the tasks, you MUST use the `finish()` action.

\textbf{Action Format}\\
1. Execute only ONE action per response. \\
2. Your response MUST be wrapped in \texttt{<action>} tags. \\
3. The action itself must start with \texttt{Action: }. \\
4. Keep your answers as short and clear as possible.

\vspace{1em}
\noindent\texttt{finish()}: Call this tool when you have completed the task. \\Example: \texttt{<action>Action: finish()</action>}

\hrulefill
\paragraph*{Responding to Questions}
When asked a multiple-choice question, you must respond in the following format: \\
\texttt{<action>Answer: [LETTER]</action>}

\vspace{0.5em}
\noindent For example: \\
\quad  \space \texttt{<action>Answer: A</action>} \\
\quad  \space \texttt{<action>Answer: B</action>} \\
\quad  \space \texttt{<action>Answer: C</action>} \\
Choose the letter that corresponds to the best answer.

\hrulefill
\paragraph*{Actions}
To use a tool, you must format your response as follows:

\texttt{<action>Action: tool\_name(param1="value1", param2="value2")</action>}

\vspace{1em}
Below is the list of tools at your disposal.
\end{textBox}

\subsection{Tool Declarations by System}
The tools are organized into logical groups corresponding to the primary subsystems of the environment.

\subsubsection{Email System Tools}

\begin{textBox}[\texttt{email.send\_email(to: str, subject: str, body: str, cc: str = None)}]
\textbf{Description:} Sends an email. \\
\textbf{Parameters:} \\
\quad \texttt{to} (required): The recipient's email address. \\
\quad \texttt{subject} (required): The subject of the email. \\
\quad \texttt{body} (required): The content of the email. \\
\textbf{Example:} \\ \texttt{<action>Action: email.send\_email(to="advisor.x@lau.edu", subject="Question", body="Dear Advisor...")</action>}
\end{textBox}

\subsubsection{Calendar System Tools}

\begin{textBox}[\texttt{calendar.add\_event(calendar\_id: str, event\_title: str, ...)}]
\textbf{Description:} Adds an event to a calendar. \\
\textbf{Parameters:} \\
\quad \texttt{calendar\_id} (required): The ID of the calendar. Use \texttt{'self'} for your personal calendar. For other calendars (e.g., advisor, club), use their official email address. \\
\quad \texttt{event\_title} (required): The title of the event. \\
\quad \texttt{location} (required): The location of the event. \\
\quad \texttt{time} (required): The time of the event (format: \texttt{'Week X, Day, HH:MM-HH:MM'}). \\
\quad \texttt{description} (optional): A detailed description for the event. \\
\textbf{Example:} \\ \texttt{<action>Action: calendar.add\_event(calendar\_id="self", event\_title="Team Meeting", location="Library Room 201", time="Week 3, Monday, 15:00-16:00")</action>}
\end{textBox}

\begin{textBox}[\texttt{calendar.remove\_event(calendar\_id: str, event\_id: str)}]
\textbf{Description:} Removes an event from a calendar. \\
\textbf{Parameters:} \\
\quad \texttt{calendar\_id} (required): The ID of the calendar. \\
\quad \texttt{event\_id} (required): The ID of the event to remove. \\
\textbf{Example:} \texttt{<action>Action: calendar.remove\_event(calendar\_id="self", event\_id="event\_005")</action>}
\end{textBox}

\begin{textBox}[\texttt{calendar.update\_event(calendar\_id: str, event\_id: str, new\_details: dict)}]
\textbf{Description:} Updates an existing event. \\
\textbf{Parameters:} \\
\quad \texttt{calendar\_id} (required): The ID of the calendar. \\
\quad \texttt{event\_id} (required): The ID of the event to update. \\
\quad \texttt{new\_details} (required): A dictionary with the new details (e.g., \texttt{\{"location": "New Room"\}}). \\
\textbf{Example:} \texttt{<action>Action: calendar.update\_event(calendar\_id="self", event\_id="event\_006", new\_details=\{"location": "Orwell Hall, Room 101"\})</action>}
\end{textBox}

\begin{textBox}[\texttt{calendar.view\_schedule(calendar\_id: str, date: str)}]
\textbf{Description:} Views all events on a specific date for a calendar. \\
\textbf{Parameters:} \\
\quad \texttt{calendar\_id} (required): The ID of the calendar. \\
\quad \texttt{date} (required): The date to view (format: \texttt{'Week X, Day'}). \\
\textbf{Example:} \texttt{<action>Action: calendar.view\_schedule(calendar\_id="self", date="Week 3, Monday")</action>}
\end{textBox}

\begin{textBox}[\texttt{calendar.query\_advisor\_availability(advisor\_id: str, date: str)}]
\textbf{Description:} Checks an advisor's free/busy schedule. \\
\textbf{Parameters:} \\
\quad \texttt{advisor\_id} (required): The ID of the advisor. \\
\quad \texttt{date} (required): The date to query (format: \texttt{'Week X, Day'}). \\
\textbf{Example:} \texttt{<action>Action: calendar.query\_advisor\_availability(advisor\_id="T0001", date="Week 4, Tuesday")</action>}
\end{textBox}

\subsubsection{Map \& Geography Tools}
\begin{textBox}[\texttt{geography.get\_current\_location()}]
\textbf{Description:} Gets your current building location. \\
\textbf{Example:} \texttt{<action>Action: geography.get\_current\_location()</action>}
\end{textBox}

\begin{textBox}[\texttt{map.find\_optimal\_path(source\_building\_id: str, target\_building\_id: str, ...)}]
\textbf{Description:} Finds the best path between two buildings. \\
\textbf{Parameters:} \\
\quad \texttt{source\_building\_id} (required): The ID of the starting building. \\
\quad \texttt{target\_building\_id} (required): The ID of the destination building. \\
\quad \texttt{constraints} (optional): A dictionary of constraints (e.g., \texttt{\{"avoid": "crowds"\}}). \\
\textbf{Example:} \texttt{<action>Action: map.find\_optimal\_path(source\_building\_id="B083", target\_building\_id="B001")</action>}
\end{textBox}

\begin{textBox}[\texttt{geography.walk\_to(path\_info: dict)}]
\textbf{Description:} Moves the agent along a calculated path. \\
\textbf{Parameters:} \\
\quad \texttt{path\_info} (required): The full path object returned by \texttt{map.find\_optimal\_path}. \\
\textbf{Example:} \texttt{<action>Action: geography.walk\_to(path\_info=\{'path': ['B083', 'B001']\})</action>}
\end{textBox}

\begin{textBox}[\texttt{map.find\_building\_id(building\_name: str)}]
\textbf{Description:} Finds a building's unique ID by its name. \\
\textbf{Parameters:} \\
\quad \texttt{building\_name} (required): The name or alias of the building. \\
\textbf{Example:} \texttt{<action>Action: map.find\_building\_id(building\_name="Grand Central Library")</action>}
\end{textBox}

\begin{textBox}[\texttt{map.get\_building\_details(building\_id: str)}]
\textbf{Description:} Gets all details for a building. \\
\textbf{Parameters:} \\
\quad \texttt{building\_id} (required): The ID of the building. \\
\textbf{Example:} \texttt{<action>Action: map.get\_building\_details(building\_id="B001")</action>}
\end{textBox}

\begin{textBox}[\texttt{map.find\_room\_location(room\_query: str, building\_id: str = None, ...)}]
\textbf{Description:} Finds the location of a specific room. \\
\textbf{Parameters:} \\
\quad \texttt{room\_query} (required): The name or number of the room. \\
\quad \texttt{building\_id} (optional): A specific building ID to search within. \\
\textbf{Example:} \texttt{<action>Action: map.find\_room\_location(room\_query="Seminar Room 101", building\_id="B014")</action>}
\end{textBox}

\begin{textBox}[\texttt{map.query\_buildings\_by\_property(...) }]
\textbf{Description:} Queries buildings based on properties. Filter by \texttt{zone}, \texttt{building\_type}, or \texttt{amenity}. At least one is required. \\
\textbf{Example:} \texttt{<action>Action: map.query\_buildings\_by\_property(amenity="Coffee Shop")</action>}
\end{textBox}

\subsubsection{Reservation System Tools}
\begin{textBox}[\texttt{reservation.query\_availability(location\_id: str, date: str)}]
\textbf{Description:} Queries the availability of bookable spaces in a location. \\
\textbf{Parameters:} \\
\quad \texttt{location\_id} (required): The ID of the building or location. \\
\quad \texttt{date} (required): The date to query (format: \texttt{'Week X, Day'}). \\
\textbf{Example:} \texttt{<action>Action: reservation.query\_availability(location\_id="B001", date="Week 4, Saturday")</action>}
\end{textBox}

\begin{textBox}[\texttt{reservation.make\_booking(location\_id: str, item\_name: str, ...)}]
\textbf{Description:} Books a specific room or seat. \\
\textbf{Parameters:} \\
\quad \texttt{location\_id} (required): The ID of the building. \\
\quad \texttt{item\_name} (required): The name of the room or area. \\
\quad \texttt{date} (required): The date for the booking (format: \texttt{'Week X, Day'}). \\
\quad \texttt{time\_slot} (required): The time slot to book (e.g., \texttt{'14:00-16:00'}). \\
\quad \texttt{seat\_id} (optional): The specific seat ID if booking a single seat. \\
\textbf{Example:} \texttt{<action>Action: reservation.make\_booking(location\_id="B001", item\_name="Group Study Room 201", date="Week 4, Saturday", time\_slot="14:00-16:00")</action>}
\end{textBox}

\subsubsection{Information \& Course Tools}
\begin{textBox}[\texttt{bibliography.list\_chapters(book\_title: str)}]
\textbf{Description:} Lists all chapters in a specified book. \\
\textbf{Note:} This tool is intended exclusively for querying assigned \textbf{textbooks and handbooks}. To search the main library collection, use the \texttt{data\_system} tools. \\
\textbf{Example:} \texttt{<action>Action: bibliography.list\_chapters(book\_title="Student Handbook")</action>}
\end{textBox}

\begin{textBox}[\texttt{bibliography.list\_sections(book\_title: str, chapter\_title: str)}]
\textbf{Description:} Lists all sections in a chapter of a textbook or handbook. \\
\textbf{Example:} \texttt{<action>Action: bibliography.list\_sections(book\_title="A Panorama of Computing", chapter\_title="Chapter 1: Search")</action>}
\end{textBox}

\begin{textBox}[\texttt{bibliography.list\_articles(book\_title: str, chapter\_title: str, ...)}]
\textbf{Description:} Lists all articles in a section of a textbook or handbook. \\
\textbf{Example:} \texttt{<action>Action: bibliography.list\_articles(book\_title="A Panorama of Computing", chapter\_title="Search", section\_title="Uninformed Search")</action>}
\end{textBox}

\begin{textBox}[\texttt{bibliography.view\_article(identifier: str, search\_type: str)}]
\textbf{Description:} Views the content of an article from a textbook or handbook. \\
\textbf{Parameters:} \\
\quad \texttt{identifier} (required): The title or ID of the article. \\
\quad \texttt{search\_type} (required): \texttt{'title'} or \texttt{'id'}. \\
\textbf{Example:} \texttt{<action>Action: bibliography.view\_article(identifier="Breadth-First Search", search\_type="title")</action>}
\end{textBox}

\begin{textBox}[\texttt{data\_system.list\_by\_category(category: str, entity\_type: str, ...)}]
\textbf{Description:} Lists clubs or advisors by category. Use this to discover entities matching certain criteria. \\
\textbf{Parameters:} \\
\quad \texttt{entity\_type} (required): \texttt{'club'} or \texttt{'advisor'}. \\
\quad \texttt{category} (required): The category to filter by (e.g., "Sports \& Fitness", "Computer Science"). \\
\textbf{Example:} \texttt{<action>Action: data\_system.list\_by\_category(category="Academic \& Technological", entity\_type="club")</action>}
\end{textBox}

\begin{textBox}[\texttt{data\_system.query\_by\_identifier(identifier: str, by: str, entity\_type: str)}]
\textbf{Description:} Gets all details for a specific club or advisor using their name or ID. \\
\textbf{Example:} \texttt{<action>Action: data\_system.query\_by\_identifier(identifier="Computer Science Club", by="name", entity\_type="club")</action>}
\end{textBox}

\begin{textBox}[\texttt{data\_system.list\_books\_by\_category(category: str)}]
\textbf{Description:} Lists all main library books in a specific category. \\
\textbf{Parameters:} \\
\quad \texttt{category} (required): The category to filter by (e.g., "History"). \\
\textbf{Example:} \texttt{<action>Action: data\_system.list\_books\_by\_category(category="Computer Science")</action>}
\end{textBox}

\begin{textBox}[\texttt{data\_system.search\_books(query: str, search\_type: str = "title")}]
\textbf{Description:} Searches main library books by title or author. Returns status, call numbers, and location. \\
\textbf{Parameters:} \\
\quad \texttt{query} (required): The search query string. \\
\quad \texttt{search\_type} (optional): \texttt{'title'} (default) or \texttt{'author'}. \\
\textbf{Example:} \texttt{<action>Action: data\_system.search\_books(query="Artificial Intelligence", search\_type="title")</action>}
\end{textBox}

\subsubsection{Course Selection System Tools}
\begin{textBox}[\texttt{course\_selection.browse\_courses(filters: dict = None)}]
\textbf{Description:} Browses available courses. The system enforces specific rules regarding course load and pass allocation per semester. \\
\textbf{Pass Guidelines:} \\
\quad  \space \textbf{S-Pass}: Guarantees enrollment for any popularity (best for 95-99). \\
\quad  \space \textbf{A-Pass}: Guarantees enrollment for popularity below 95. \\
\quad  \space \textbf{B-Pass}: Guarantees enrollment for popularity below 85. \\
\textbf{Parameters:} \\
\quad \texttt{filters} (optional): A dictionary to filter by \texttt{course\_code}, \texttt{course\_name}, or \texttt{credits}. \\
\textbf{Example:} \texttt{<action>Action: course\_selection.browse\_courses(filters=\{"course\_name": "Introduction"\})</action>}
\end{textBox}

\begin{textBox}[\texttt{draft.add\_course(section\_id: str)}]
\textbf{Description:} Adds a course to the draft schedule. \\
\textbf{Example:} \texttt{<action>Action: draft.add\_course(section\_id="WXK003111107")</action>}
\end{textBox}

\begin{textBox}[\texttt{draft.remove\_course(section\_id: str)}]
\textbf{Description:} Removes a course from the draft schedule. \\
\textbf{Example:} \texttt{<action>Action: draft.remove\_course(section\_id="WXK003111107")</action>}
\end{textBox}

\begin{textBox}[\texttt{draft.assign\_pass(section\_id: str, pass\_type: str)}]
\textbf{Description:} Assigns a priority pass to a drafted course. \\
\textbf{Parameters:} \\
\quad \texttt{section\_id} (required): The ID of the course section. \\
\quad \texttt{pass\_type} (required): \texttt{'S-Pass'}, \texttt{'A-Pass'}, or \texttt{'B-Pass'}. \\
\textbf{Example:} \texttt{<action>Action: draft.assign\_pass(section\_id="SHK003111017", pass\_type="A-Pass")</action>}
\end{textBox}

\begin{textBox}[\texttt{draft.view()}]
\textbf{Description:} Views the current draft schedule. \\
\textbf{Example:} \texttt{<action>Action: draft.view()</action>}
\end{textBox}

\begin{textBox}[\texttt{registration.submit\_draft()}]
\textbf{Description:} Submits the draft schedule for final registration. \\
\textbf{Example:} \texttt{<action>Action: registration.submit\_draft()</action>}
\end{textBox}

\subsection{Tool Usage by Task}

Table~\ref{tab:tool_usage_by_task} provides a summary of the primary tool systems available to the agent for each distinct task scenario within \textit{StuLife Bench}. The selection of tools for each task is intentionally constrained to reflect realistic limitations and to focus the evaluation on specific agent capabilities. For example, course selection tools are only available during the relevant planning and registration phases.

\subsection{Tool Usage by Task}

Table~\ref{tab:tool_usage_by_task} provides a summary of the primary tool systems available to the agent for each distinct task scenario within \textit{StuLife Bench}. The selection of tools for each task is intentionally constrained to reflect realistic limitations and to focus the evaluation on specific agent capabilities. For example, course selection tools are only available during the relevant planning and registration phases.

\begin{table}[h!]
\centering
\caption{Primary Tool Systems Available for Each Task Scenario}
\label{tab:tool_usage_by_task}
\renewcommand{\arraystretch}{1.3}
\resizebox{\linewidth}{!}{%
\begin{tabular}{llp{10cm}} 
\toprule
\textbf{Core Scenario} & \textbf{Task Scenario} & \textbf{Available Tool Systems} \\
\midrule
\multirow{2}{*}{In-Class} & Regulations Learning & \texttt{student\_handbook}, \texttt{bibliography}, \texttt{textbooks} \\
 & Core Course Instruction & \texttt{bibliography}, \texttt{calendar}, \texttt{data\_system}, \texttt{email}, \texttt{geography}, \texttt{map}, \texttt{reservation}, \texttt{student\_handbook}, \texttt{textbooks} \\
\midrule
\multirow{6}{*}{Daily Campus} & Campus Exploration & \texttt{map}, \texttt{geography}, \texttt{data\_system}, \texttt{calendar}, \texttt{bibliography}, \texttt{course\_selection}, \texttt{draft}, \texttt{registration} \\
 & Initial Course Selection & \texttt{course\_selection}, \texttt{draft}, \texttt{registration}, \texttt{data\_system}, \texttt{student\_handbook}, \texttt{calendar}, \texttt{bibliography}, \texttt{geography}, \texttt{map} \\
 & Preliminary Planning & \texttt{course\_selection}, \texttt{draft}, \texttt{registration}, \texttt{data\_system}, \texttt{student\_handbook}, \texttt{calendar}, \texttt{bibliography}, \texttt{geography}, \texttt{map} \\
 & Academic Activity & \texttt{calendar}, \texttt{email}, \texttt{reservation}, \texttt{data\_system}, \texttt{map}, \texttt{geography}, \texttt{bibliography}, \texttt{student\_handbook}, \texttt{textbooks} \\
 & Library Study & \texttt{reservation}, \texttt{bibliography}, \texttt{data\_system}, \texttt{map}, \texttt{geography}, \texttt{calendar}, \texttt{email}, \texttt{student\_handbook}, \texttt{textbooks} \\
 & Club Activity & \texttt{calendar}, \texttt{email}, \texttt{reservation}, \texttt{data\_system}, \texttt{map}, \texttt{geography}, \texttt{bibliography}, \texttt{student\_handbook}, \texttt{textbooks} \\
\midrule
\multirow{2}{*}{Examination} & Midterm Exams & \texttt{calendar}, \texttt{email}, \texttt{reservation}, \texttt{data\_system}, \texttt{map}, \texttt{geography}, \texttt{bibliography}, \texttt{student\_handbook}, \texttt{textbooks} \\
 & Final Exams & \texttt{calendar}, \texttt{email}, \texttt{reservation}, \texttt{data\_system}, \texttt{map}, \texttt{geography}, \texttt{bibliography}, \texttt{student\_handbook}, \texttt{textbooks}, \texttt{draft}, \texttt{registration} \\
\bottomrule
\end{tabular}%
}
\end{table}

\section{Generation Details for Each Sub-task}

\subsection{Campus Exploration Task}
This task addresses the complex scenario of multi-leg campus exploration under various constraints. To ensure narrative coherence, rigorous tool use, and verifiable action sequences, we employed a pipeline that combines deterministic state construction with a two-stage generation process.

\paragraph{Data Preparation and Deterministic State Generation}
At the backend, we first establish a ground-truth foundation through deterministic processes.
\begin{itemize}[leftmargin=*, align=left]
    \item We perform multi-leg path planning on a graph representation of the campus. For each query, a deterministic process generates the ground-truth task status, including the optimal path connecting the source, waypoints, and target in sequence.
    \item Constraints such as accessibility, weather exposure, path type, illumination, and congestion are modeled as soft penalties. These penalties influence the path selection during planning to generate more diverse and realistic routes.
    \item We concurrently extract structured information about buildings along the path (e.g., official names, aliases, contained areas, and internal facilities) to provide rich, contextual details for the subsequent instruction-writing phase.
    \item Query samples are generated following controllable rules (e.g., number of waypoints, probability of constraints) to guarantee task diversity, controllability, and realism.
\end{itemize}

\paragraph{Two-Stage LLM Generation}
We separate the creative and logical aspects of generation into a two-stage LLM pipeline.
\begin{itemize}[leftmargin=*, align=left]
    \item \textbf{Stage 1: Instruction Generation (Creative Agent).} The first stage aims to generate a concise, believable, and motivationally-grounded 'instruction' from a first-person perspective. Based on the planned path and building information, a creative agent is prompted to write a narrative that naturally embeds all waypoints and constraints. For instance, it might reference a specific internal amenity of a building (e.g., the "Circulation Desk" in the library) or weave a constraint into the story (e.g., needing an accessible route for a friend).
    
    \item \textbf{Stage 2: Solution and Evaluation Trace Generation (Logical Agent).} A local, deterministic Python script generates the ground-truth solution. This script employs Dijkstra's algorithm to compute the optimal path, from which the ground-truth action sequence is derived. We ensure that this pathfinding algorithm is identical to the one available to the agent during its evaluation phase.
\end{itemize}

\subsubsection{Verbatim Prompts for Campus Exploration Task}

\begin{promptbox}[Instruction Generation]
### Instructions and Constraints

1.  **Persona and Tone**:
    *   You MUST speak as a senior student guide giving a spontaneous challenge.
    *   Start with a direct, friendly, and energetic greeting. For example: "Hey! Got a quick challenge for you to help you learn the campus."
    *   Maintain a helpful and encouraging tone.

2.  **Urgency and Goal**:
    *   Since `execution_type` is `immediate`, you MUST state that the task needs to be done **right now**.
    *   The goal is twofold: first, to **plan a route**, and second, to **actually walk that route** to complete the exploration.

3.  **Route Details & Constraints**:
    *   This prompt will be dynamically filled by a script. Your job is to ensure the final output is a single, natural-sounding paragraph.
    *   The script will provide the core sentence structure, including start/end points and any constraints.
    *   It will also provide a sentence about passing points via the `{passing_points_sentence}` placeholder. If there are no passing points, this will be empty.

4.  **Closing**:
    *   End with a brief, encouraging closing. For example: "Good luck!"

-----

### Example

**Input Data (from script):**
*   `source_name`: "Grand Central Library"
*   `target_name`: "Innovation Hub"
*   `passing_points_sentence`: "To make it interesting, you must pass by the Student Union, then the Engineering Building, in that specific order."
*   `constraints_string`: `{"shelter": "Full", "congestion": "Low"}`
*   `execution_type`: "immediate"

**Desired Output:**

Hey! Got a quick challenge for you to help you learn the campus. Your task, starting now, is to first plan and then walk a route from the **Grand Central Library** to the **Innovation Hub**. {passing_points_sentence} For this challenge, try to find a path that's fully covered and isn't too crowded. Good luck!

# INPUT
\end{promptbox}

\subsection{Course Selection Task}
This task evaluates an agent's ability to perform strategic course selection and optimize resource (Pass Card) allocation under complex constraints. Its construction paradigm is centered around \textbf{Constraint-Driven Unique Solution Construction}. The goal is to ensure that for any given task scenario—considering course popularity, instructions, and the academic plan—a single optimal solution exists, thereby guaranteeing the reliability of our evaluation.

\paragraph{Data and State Construction}
The task is built upon a student's academic plan and the university's course catalog, which form the foundational constraints (e.g., credits, prerequisites, time conflicts). Each course is assigned a "popularity" value from 0 to 100, representing the enrollment competition.

The core resource is a hierarchical system of "Pass Cards" with the following universal rules:
\begin{itemize}[leftmargin=*, align=left]
    \item \textbf{S-Pass}: Can forcibly enroll in \textbf{any} course (popularity 0-100). It is optimally used for courses with a popularity of 95-99.
    \item \textbf{A-Pass}: Guarantees enrollment in courses with a popularity \textbf{below 95}.
    \item \textbf{B-Pass}: Can only be used for courses with a popularity \textbf{below 85}; quantity is unlimited.
\end{itemize}
The initial state of the task includes the student's draft schedule, a ground-truth "Target Schedule," and the Pass Cards allocated at the beginning of each semester according to that semester's rules.

\paragraph{Unique Solution Construction Mechanism}
To ensure each task is a logic puzzle with a unique solution, we deterministically back-engineer the popularity of other courses based on the "Target Schedule" and the agent's available Pass Cards. This transforms a resource allocation problem into a logical reasoning challenge. For instance:
\begin{itemize}[leftmargin=*, align=left]
    \item \textbf{To force the use of an S-Pass}, the system will set the popularity of a target required course to \textbf{95 or higher}.
    \item \textbf{To guide the use of an A-Pass}, the system will set a target course's popularity to a value \textbf{between 85 and 94}.
\end{itemize}
By precisely orchestrating the popularity of target and distractor courses, all non-optimal paths are logically blocked.

\paragraph{Dynamic Multi-Semester Task Chain}
Course selection in this benchmark is not a single event but simulates two consecutive and dynamically evolving stages: \textbf{Semester 1} and \textbf{Semester 2(Preliminary Planning)}. The agent's state at the end of Semester 1 (final schedule and remaining resources) seamlessly becomes the initial state for Semester 2.

More challenging, the \textbf{constraints and resource allocations change between semesters}. For example:
\begin{itemize}[leftmargin=*, align=left]
    \item \textbf{Semester 1}: Requires completing 8 courses (including at least 6 required ones) and provides \textbf{2 A-Passes} for required courses.
    \item \textbf{Semester 2}: Requires 7 courses (including at least 5 required ones), while the A-Pass allocation for required courses is \textbf{reduced to 1}.
\end{itemize}
This \textbf{dynamically evolving design} aims to evaluate an agent's capabilities for \textbf{memory, adaptation to new rules, and forward-looking resource planning} in long-term tasks.

Finally, the ultimate stage of this task chain is designed as a \textbf{Convergence Point}. Through a scenario like "joining an Excellent Student Program," all correctly performing agents are guided to the exact same final schedule, ensuring the fairness and comparability of evaluations in subsequent tasks.

\paragraph{Instruction Generation}
After all deterministic states are constructed, this structured information (including the specific rules for each semester) is converted into a natural, context-aware language instruction to guide the agent.

\begin{promptbox}[Instruction Generation]
# ROLE: You are a Master Narrative Designer and creative writer for a complex simulation.

# TASK: Your mission is to generate the `advice_text` and `agent_expected_actions_desc` for a single step in a student's course selection journey. You will be given the context of the step, including the character (`persona`), the event type, the changes in the world, and the exact schedule changes that need to happen (`expected_outcome_delta`). Your job is to create a compelling, in-character narrative justification and a clear, actionable plan.

# CONTEXT FOR CURRENT STEP: {step}

## 1. Persona (Who is speaking?)
`{persona}`

## 2. Event Type (What is the theme of this event?)
`{event_type}`

## 3. World State Change (What external factors have changed?)
```json
{world_state_change_json}
```

## 4. Student's Schedule BEFORE this step
```json
{previous_step_output_json}
```

## 5. Required Schedule Changes (The "What")
This is the ground truth of what actions MUST be taken in this step. Your output must logically lead to these exact changes.
```json
{expected_outcome_delta_json}
```

## 6. Details of Courses Involved in the Change
Here is all the information about the courses mentioned in the `expected_outcome_delta`. Use this to make your narrative specific and believable.
```json
{relevant_courses_json}
```

# YOUR TASK: Generate the Narrative and Actions (The "Why" and "How")

Based on all the context above, generate a JSON object with two keys:

1.  `advice_text`: **Craft a compelling narrative from the perspective of the `{persona}`.** Your primary goal is to create a story that explains **every single change** in `expected_outcome_delta`.
    *   **Mandatory Checklist for Coverage**: Before generating the final text, you **MUST** verify that your narrative explicitly justifies every single change listed below. Treat this as a checklist.
        *   **Added Sections**: Your narrative must explain why each course in `added_sections` is being added.
        *   **Removed Sections**: Your narrative must explain why each course in `removed_sections` is being dropped.
        *   **Pass Changes**: Your narrative must explain the reasoning behind every single `pass_changes`.
        *   **No Omissions**: Failure to address every item in the delta is a failure to complete the task.
    *   **Embody the Persona**: You **MUST** start the advice by clearly stating your role. For example, if the persona is "Roommate", begin with "Hey, as your roommate, I was just checking the course system and saw..." or if it's "Counselor", start with "As your academic counselor, I have some important updates for you."
    *   **Create a Thematic Cause-and-Effect Narrative**: Your story's main theme is defined by the `{event_type}`. Act like a smart analyst: select the **most relevant updates** from the `world_state_change` list to use as the specific *causes* that logically lead to the actions in `expected_outcome_delta` (the *effect*). You do not need to mention every single world state change, only the ones that justify the required actions. For example, if the `event_type` is `Popularity_Update_Risk_Cascade`, you should focus on the courses whose popularity skyrocketed and explain how this new risk forces the specific pass changes and course swaps in the delta.
    *   **Refer to Courses by Name**: To make the advice sound like a real, natural conversation, you **MUST** refer to courses by their **name only** (e.g., "Advanced AI", "Machine Learning"). **Crucially, do NOT include course codes in your response** (e.g., avoid formats like "CS101" or "Advanced AI (CS101)"). The goal is to simulate a human giving advice, not a system generating a report.
    *   **Tone and Style**: Your language **MUST** be conversational, persuasive, and use a "soft" or uncertain tone, as if you are giving friendly advice, not commands.
    *   **Use Collaborative & Suggestive Phrasing**: Instead of stating conclusions as facts, phrase them as suggestions or questions.
        *   **Instead of**: "This course is less popular, so downgrade its pass."
        *   **Try**: "This course's popularity doesn't seem so crazy anymore, maybe we don't need to use such a high-priority pass on it? What do you think?"
        *   **Instead of**: "You must swap this course."
        *   **Try**: "I noticed this other course has a better time slot that fits your schedule perfectly, perhaps it's a better option?"
    *   **GOOD EXAMPLE (Natural & Suggestive Tone)**: "Hey, as your roommate, I was just looking at the course system. 'Advanced AI' seems to be getting way more popular, maybe we should think about using your S-Pass on it just to be safe? If we do that, we could probably free up the A-Pass from 'Machine Learning'-its popularity isn't as wild as we thought, so an A-pass might be overkill there. What do you think?"
    *   **BAD EXAMPLE (Too Direct & Factual)**: "The popularity of 'Advanced AI' has increased, therefore you must upgrade it to an S-Pass. The popularity of 'Machine Learning' is lower, so you can downgrade it to an A-Pass without risk."

2.  `agent_expected_actions_desc`: **Create a simple and clear "To-Do List"** that summarizes the required actions. This should be a direct, imperative translation of the `expected_outcome_delta` that the agent can easily follow. Use active verbs. For example: "1. **Drop Course**: Remove 'Course Y'. 2. **Add Course**: Add 'Course X'. 3. **Upgrade Pass**: Change the pass for 'Course Z' from B-Pass to A-Pass."

# OUTPUT FORMAT

You must output **only a single, valid JSON object** containing the two specified keys. Do not add any explanatory text.

### Example Output
```json
{{
  "advice_text": "I've just seen the latest registration trends. The popularity for 'Calculus II' has skyrocketed, making it a high-risk course. I strongly recommend you upgrade its pass to your S-Pass for maximum security. Consequently, 'Intro to Programming' is less popular than we thought, so you can safely downgrade it to an A-Pass to free up your S-Pass.",
  "agent_expected_actions_desc": "1. **Change Pass**: Upgrade 'Calculus II' from A-Pass to the S-Pass. 2. **Change Pass**: Downgrade 'Intro to Programming' from S-Pass to an A-Pass."
}}
``` 
\end{promptbox}

\subsection{Library Study Task}
This task evaluates the agent's ability to manage studying and material look-up within a campus library environment. It covers two temporal requirements, Immediate and Scheduled Execution, and distinguishes between two narrative styles: internal monologue and received message. The overall pipeline follows the paradigm of deterministic state construction followed by two-stage LLM generation, with a focus on temporal consistency, motivational reasoning, and inferable resource needs.

\paragraph{Data and State Construction}
Task seeds are composed of two main categories:
\begin{itemize}[leftmargin=*, align=left]
    \item \textbf{Topic-Based Study:} This is divided into "specific book" (requiring the use of the \texttt{data\_system.\\search\_books} tool for location) and "general topic" types. Both include a 'persona', 'reason', and 'implied\_requirements'.
    \item \textbf{General Study:} This centers on finding a seat for effective study. The 'persona' and 'reason' drive implicit seating and environmental needs (e.g., 'quiet\_zone', 'power\_outlet').
\end{itemize}
To simulate diverse scenarios in real campus life, the 'persona' is not limited to the student's own internal monologue but can also originate from external characters, such as suggestions from a roommate or tasks assigned by a counselor.

Topic priority balances relevance to the student's coursework with interdisciplinary interests. To test the agent's ability to infer the correct execution location based on task requirements, we have established a strict collection rule: \textbf{books and materials for a specific topic are located in one, and only one, designated library}. In the task instruction, the explicit library name is deliberately hidden. The agent must first call the \texttt{data\_system.list\_books\_by\_category(category=...)} tool to query the collection information for a specific topic, thereby inferring the correct library location before proceeding with subsequent planning.

\paragraph{Temporal Semantics and Long-Term Memory Construction}
We divide tasks into two categories along the temporal dimension to evaluate the agent's full range of capabilities:
\begin{itemize}[leftmargin=*, align=left]
    \item \textbf{Immediate Execution:} These tasks require the agent to immediately understand and execute the instruction, designed to test its rapid response capabilities.
    \item \textbf{Scheduled Execution for Long-Term Memory:} These tasks are specifically designed to evaluate the agent's \textbf{long-term memory, planning, and ability to act at specific future points in time}. Each scheduled task consists of a \textbf{Trigger Condition} and an \textbf{Execution Window}. The trigger condition is typically a specific future time point. To construct diverse long-term challenges, the triggers we generate maintain a balanced ratio between \textbf{same-day} and \textbf{cross-day} time spans.
\end{itemize}

\paragraph{Instruction Generation}
Instruction generation integrates two dimensions, \textbf{narrative style} and \textbf{temporal requirements}, to create diverse task scenarios. Narratively, instructions can be the student's first-person \textbf{internal monologue} (corresponding to their own thoughts) or a \textbf{received message} (such as a suggestion from a roommate or a task from a counselor). Temporally, the instruction's wording will clearly distinguish between tasks that must be \textbf{executed immediately} and those that need to be \textbf{scheduled for a specific future time}. All instructions adhere to strict consistency constraints, such as maintaining the student's academic background (Computer Science) and translating abstract requirements into concrete language.
\subsubsection{Verbatim Prompts for Library Study Task}

\begin{promptbox}[Style A $\cdot$ Topic-Based $\cdot$ Immediate]
# CONTEXT
You are a "Scenario Generator" AI. Your role is to create a realistic, first-person **stimulus** for an autonomous AI assistant benchmark. This stimulus represents the **internal thoughts or personal plans** of a university student in Japan. The AI assistant being tested will later read this stimulus and decide on a course of action.

# INPUT

# TASK

Your primary task is to generate a natural and richly detailed scenario description based on the input JSON, following the **Style A: First-Person Internal Monologue** guide below.

**IMPORTANT: Your response should contain ONLY the instruction text content. Do not output JSON, code blocks, or any other formatting. Just output the raw text that will become the `instruction` field.**

# STYLE GUIDE: First-Person Internal Monologue / Personal Plan

* **Description:** The output must be a direct expression of the student's own thoughts, self-reflection, or plans, as if thinking out loud. It is a statement of intent that an assistant is meant to "overhear" and act upon.
* **Crucial Rule:** It must **NOT** be a command or question directed at an assistant (e.g., avoid "Can you find...", "Please book...").
* **CRITICAL: Immediate Intent Mandate:** The student's thought process MUST conclude with a clear decision to act **right now**. Use the `task_time` to ground the thought in the present moment and trigger the immediate action. The monologue should build to a point of decision, using phrases like:
    * "Okay, it's 10:30 AM. I should get this sorted and find a place right now."
    * "My dorm is too distracting at the moment. I need to get out of here and find a spot immediately."
    * "I've made up my mind. I'm going to find a quiet place to work on this now."
* **Single Seat Focus:** The student should be thinking about booking ONE seat for themselves only.
* **CRITICAL: No Academic Deadline Pressure:** The sense of immediacy must be spontaneous and internal (e.g., "I'm in the zone and need a quiet place now," or "My current location is too noisy"). NEVER mention external pressures like "next week's exam," "assignment due soon," or any specific academic deadlines.
* **CRITICAL: Major Academic Consistency Check:** The student is a COMPUTER SCIENCE major. Therefore:
    * **IF** the topic is "AI", "Psychology/Mental Health", "Mathematics", or "Military Theory" $\rightarrow$ Can be related to coursework/academics (without deadlines).
    * **IF** the topic is anything else $\rightarrow$ Motivation MUST be purely interest-based ("I've always been curious about..."). NEVER mention assignments, grades, or professors.
* **Topic Integration (CRITICAL):** The `topic` field must be naturally woven into the narrative to indirectly suggest the appropriate type of library, without explicitly naming one.
* **CRITICAL: Resource-Seeking Behavior:** If no `specific_book` is mentioned, the student must express a clear need to find a place with relevant topic-related resources (e.g., "somewhere with a good collection of [topic] books").
* **Implied Requirements Integration:** Naturally weave `implied_requirements` into the thoughts with specific language (e.g., `"power_outlet"` $\rightarrow$ "I'll need to plug in my laptop").
* **Time and Duration:** Use `task_time` to set the scene. Convert `reservation_duration_hours` into a natural phrase (e.g., `4.0` -> "for a solid four hours").

# FINAL CHECKLIST

Before providing your final output, **review it carefully to ensure it follows these critical rules:**

* **1. Plain Text Only:** Output ONLY the instruction text content.
* **2. CRITICAL - Immediate Intent:** Does the monologue clearly express the student's decision to find a place **right now**?
* **3. No Direct Commands:** The text is a statement of intent, not a command.
* **4. Single Seat Focus:** The thought is about one seat for the student only.
* **5. NO Library Names:** The library type is implied by the topic, not named.
* **6. Topic Integration:** The topic is naturally woven into the scenario.
* **7. CRITICAL - Resource-Seeking:** If no book is named, does the student express a need for topic resources?
* **8. CRITICAL - No Academic Deadlines:** ALL time-bound academic pressures are eliminated.
* **9. MOST CRITICAL - Academic Consistency:** Non-CS topics are framed as personal interest only.

---

\end{promptbox}

\begin{promptbox}[Style A $\cdot$ Topic-Based $\cdot$ Scheduled]
# CONTEXT
You are a "Scenario Generator" AI. Your role is to create a realistic, first-person **stimulus** for an autonomous AI assistant benchmark. This stimulus represents the **internal thoughts or personal plans** of a university student in Japan. The AI assistant being tested will later read this stimulus and decide on a course of action.

# INPUT

# TASK

Your primary task is to generate a natural and richly detailed scenario description based on the input JSON, following the **Style A: First-Person Internal Monologue** guide below.

**IMPORTANT: Your response should contain ONLY the instruction text content. Do not output JSON, code blocks, or any other formatting. Just output the raw text that will become the `instruction` field.**

# STYLE GUIDE: First-Person Internal Monologue / Personal Plan

* **Description:** The output must be a direct expression of the student's own thoughts, self-reflection, or plans, as if thinking out loud or making a mental note. It is a statement of intent that an assistant is meant to "overhear" and act upon.
* **Crucial Rule:** It must **NOT** be a command or question directed at an assistant (e.g., avoid "Can you find...", "Please book...").
* **CRITICAL: Scheduled Intent Mandate:** The student's thought process MUST be a plan for a **precise future moment**. This moment is a combination of the `target_date` and the `task_time` from the JSON. The monologue must be an unambiguous plan for a future action.
* **Your output MUST clearly state BOTH the date and the time of the intended booking action.**
* Use clear, scheduling-focused language that combines date and time. See the `Date Handling` section for specific examples of how to phrase the date.
    * **Correct Example:** "Okay, plan for later **today**: right **at 3:30 PM**, I'll find a spot..."
    * **Correct Example:** "I should plan for **tomorrow, Sunday**. **Around 10:00 AM**, I'll need to find a good spot..."
    * **INCORRECT Example (Missing Date):** "I should find a spot at 10:00 AM."
    * **INCORRECT Example (Missing Time):** "I should find a spot tomorrow."
* **Single Seat Focus:** The student should be thinking about booking ONE seat for themselves only.
* **CRITICAL: No Academic Deadline Pressure:** While the student is planning to act at a specific time, this action must NOT be driven by an external deadline. The motivation should be about scheduling or personal preference. NEVER mention "next week's exam," "assignment due soon," etc.
* **CRITICAL: Major Academic Consistency Check:** The student is a COMPUTER SCIENCE major. Therefore:
    * **IF** the topic is "AI", "Psychology/Mental Health", "Mathematics", or "Military Theory" $\rightarrow$ Can be related to coursework/academics (without deadlines).
    * **IF** the topic is anything else $\rightarrow$ Motivation MUST be purely interest-based ("I want to explore..."). NEVER mention assignments, grades, or professors.
* **Topic Integration (CRITICAL):** The `topic` field must be naturally woven into the narrative to indirectly suggest the appropriate type of library, without explicitly naming one.
* **CRITICAL: Resource-Seeking Behavior:** If no `specific_book` is mentioned, the student must express a clear need to find a place with relevant topic-related resources (e.g., "I'll need access to a good collection of [topic] books").
* **Implied Requirements Integration:** Naturally weave `implied_requirements` into the thoughts with specific language (e.g., `"quiet_zone"` $\rightarrow$ "I'll need somewhere quiet to concentrate").
* **Time and Duration:** `task_time` (formerly `task_time`) is the **target time for the future action**. Convert `reservation_duration_hours` into a natural phrase (e.g., `3.5` -> "for three and a half hours").

## Date and Time Handling for Scheduled Reservations

**MANDATORY REQUIREMENT: The sentence that states the plan to book a seat MUST contain BOTH the target date and the target time. They cannot be separated.**

*   **Structure:** The core instruction MUST follow this pattern: `[Contextual sentence(s)]. I need to book a seat for myself on [DATE] at [TIME]. [Additional details].`
*   **Date Phrasing:**
    *   If `current_date` and `target_date` are IDENTICAL, you MUST use the word "**today**".
    *   If they are DIFFERENT, you MUST use a conversational phrase for the `target_date` (e.g., "tomorrow, Sunday", "on Saturday of Week 4").
    *   **NEVER** mention `current_date` in the output.

*   **Example 1 (Same Day):**
    *   **Input:** `"current_date": "Week 12, Sunday"`, `"target_date": "Week 12, Sunday"`, `"task_time": "15:30"`
    *   **Correct Output:** "...To prepare, I need to remember to book a spot for myself **today at 3:30 PM**...."
    *   **INCORRECT:** "...I'll book a spot at 3:30 PM. I need to get this done today..." (Date and time are in separate sentences).

*   **Example 2 (Future Day):**
    *   **Input:** `"current_date": "Week 2, Saturday"`, `"target_date": "Week 4, Saturday"`, `"task_time": "16:30"`
    *   **Correct Output:** "...My plan is to find a place to work on this. I'll sort out the booking **on Saturday of Week 4 right at 4:30 PM**...."
    *   **INCORRECT:** "...I'm planning to work on this on Saturday of Week 4. I'll book a table at 4:30 PM..." (Date and time are disconnected from the action).

# FINAL CHECKLIST

Before providing your final output, **review it carefully to ensure it follows these critical rules:**

* **1. Plain Text Only:** Output ONLY the instruction text content.
* **2. MANDATORY | Date/Time Adjacency:** Is the plan to book a seat phrased so that the **date and time are in the same clause**, directly linked to the action verb (e.g., "I'll book a seat **on DATE at TIME**")?
* **3. No Direct Commands:** The text is a statement of intent, not a command.
* **4. Single Seat Focus:** The thought is about one seat for the student only.
* **5. NO Library Names:** The library type is implied by the topic, not named.
* **6. Topic Integration:** The topic is naturally woven into the scenario.
* **7. CRITICAL - Resource-Seeking:** If no book is named, does the student express a need for topic resources?
* **8. CRITICAL - No Academic Deadlines:** ALL time-bound academic pressures are eliminated.
* **9. MOST CRITICAL - Academic Consistency:** Non-CS topics are framed as personal interest only.
* **10. Correct Date Phrasing:** Is the date handled correctly ("today" for same-day, conversational for future dates)?
\end{promptbox}

\begin{promptbox}[Style B $\cdot$ General-Study $\cdot$ Immediate]
# CONTEXT
You are a "Scenario Generator" AI. Your role is to create a realistic, first-person **stimulus** for an autonomous AI assistant benchmark. This stimulus represents **incoming messages or direct instructions** received by a university student in Japan. The AI assistant being tested will later read this stimulus and decide on a course of action.

# INPUT
# TASK

Your primary task is to generate a natural and richly detailed scenario description based on the input JSON, following the **Style B: Received Message / Direct Quote** guide below.

**IMPORTANT: Your response should contain ONLY the instruction text content. Do not output JSON, code blocks, or any other formatting. Just output the raw text that will become the `instruction` field.**

# STYLE GUIDE: Received Message / Direct Quote

* **Description:** The output must be a direct quote or message the student just received from the `persona`. The persona should speak directly to the student in first person (e.g., "Hey, I'm your roommate..." not "My roommate said...").
* **Crucial Rule:** The persona should suggest that THE STUDENT needs to book/reserve a seat, not that the persona has already booked something. The focus is on the student taking action.
* **CRITICAL: Immediate Action Mandate:** The message MUST create a clear sense of immediacy, prompting the student to perform the booking **right now**. The `details.task_time` and `details.date_info` from the JSON should be used to set the scene for why the action is happening now. Use direct and actionable phrases:
    * "It's 10:00 AM on Wednesday now, so it's a good time to book."
    * "Let's get this sorted out right away."
    * "Could you go ahead and book that for us now?"
    * "Since we're planning this now, can you make the reservation?"
* **CRITICAL: Single Seat Booking Only:** Even in collaboration scenarios, make it crystal clear that the student should book ONLY ONE seat for themselves. The persona must explicitly state they will handle their own seating arrangements or find a way to sit nearby without needing a separate reservation.
* **CRITICAL: Academic Consistency Check:** The student is a COMPUTER SCIENCE major. Therefore:
    * **IF** the topic/activity relates to "AI", "Psychology/Mental Health", "Mathematics", or "Military Theory" $\rightarrow$ Can be academic/coursework related
    * **IF** the topic is anything else $\rightarrow$ Must be interest-based only. Use phrases like "interest study group", "hobby exploration", "curiosity-driven learning", "personal passion project"
* **Time Constraint Nuance:** Avoid mentions of external pressures like "tomorrow's exam" or "due tomorrow." The urgency should come from the spontaneous nature of the plan (e.g., "Let's do this now while we're thinking about it"), not from a hard deadline.
* **Implied Requirements Integration:** The `implied_requirements` must be naturally woven into the persona's message with specific, actionable language. Do NOT use generic phrases. Instead, translate each requirement into concrete, contextual requests. The list below provides examples, but you are required to translate **ALL** requirements from the input JSON.
    * `"power_outlet"` $\rightarrow$ "find a spot near an electrical outlet" / "make sure your seat has access to power"
    * `"quiet_zone"` $\rightarrow$ "book in the silent study area" / "find somewhere in the no-talking zone"
    * `"computer_access"` $\rightarrow$ "book a seat that has a computer" / "try to get one of the desks that comes with a PC"
    * `"discussion_zone"` $\rightarrow$ "find somewhere we can talk and collaborate" / "pick a spot in the discussion areas"
    * `"low_traffic_area"` $\rightarrow$ "find a spot away from busy walkways" / "pick a quieter corner with less foot traffic"
* **Collaboration Clarity:** For multi-person scenarios, the persona should use varied phrases like:
    * "You handle booking your seat, I'll sort out mine"
    * "Just secure one spot for yourself, I can manage from there"
* **Rich Context:** Weave the `reason` into a believable story with emotional depth and specific details, but keep it casual.
* **Time and Duration:** Use `details.task_time` and the context from `details.date_info` to set the scene naturally. Convert `reservation_duration_hours` into conversational language.
* **`target_library` Handling:** If `target_library` has a value, mention it naturally. If it's `null`, do NOT mention any library name.
* **CRITICAL: Closing Remark:** The message must end with a clear, encouraging English closing statement that prompts the user to go to the library after booking. For example: `Let's book it now and head to the library!` or `Once you book it, let's go straight there!`

# FINAL CHECKLIST

Before providing your final output, **review it carefully to ensure it follows these critical rules:**

* **1. Plain Text Only:** Output ONLY the instruction text content.
* **2. CRITICAL: Immediate Action:** Is it 100% clear that the booking must happen **NOW**?
* **3. Student Action Focus:** The persona suggests the STUDENT should book the seat.
* **4. ABSOLUTELY CLEAR Single Seat:** Is it explicit that the student only needs to book ONE seat for themselves?
* **5. Implied Requirements PRECISELY Addressed:** Each JSON requirement is translated into specific, actionable language.
* **6. STRICT REQUIREMENT CHECK:** Have you double-checked to ensure **EVERY SINGLE** `implied_requirement` from the JSON input is included in your response? Failure to include all of them will result in an incorrect output.
* **7. Library Name Handled Correctly:** Library name is present or absent as required.
* **8. Time and Duration Integrated:** The text naturally mentions the booking duration.
* **9. No Ambiguity:** It's clear only one seat reservation is needed.
* **10. Fresh Language:** Avoids copying the examples.
* **11. Contextual Depth:** The message feels authentic.
* **12. NO External Time Pressure:** The urgency is spontaneous, not based on a deadline.
* **13. Academic Consistency:** The topic correctly reflects the student's major or is framed as a hobby.
* **14. Encouraging Closing:** Does the message end with the required English closing statement for immediate action?

---

\end{promptbox}

\begin{promptbox}[Style B $\cdot$ General-Study $\cdot$ Scheduled]
# CONTEXT
You are a "Scenario Generator" AI. Your role is to create a realistic, first-person **stimulus** for an autonomous AI assistant benchmark. This stimulus represents **incoming messages or direct instructions** received by a university student in Japan. The AI assistant being tested will later read this stimulus and decide on a course of action.

# INPUT
# TASK

Your primary task is to generate a natural and richly detailed scenario description based on the input JSON, following the **Style B: Received Message / Direct Quote** guide below.

**IMPORTANT: Your response should contain ONLY the instruction text content. Do not output JSON, code blocks, or any other formatting. Just output the raw text that will become the `instruction` field.**

# STYLE GUIDE: Received Message / Direct Quote

* **Description:** The output must be a direct quote or message the student just received from the `persona`. The persona should speak directly to the student in first person (e.g., "Hey, I'm your roommate..." not "My roommate said...").
* **Crucial Rule:** The persona should suggest that THE STUDENT needs to book/reserve a seat, not that the persona has already booked something. The focus is on the student taking action.
* **CRITICAL: Scheduled Action Mandate:** The message MUST instruct the student to perform the booking at a **precise future moment**. This moment is a combination of the `target_date` and the `task_time` from the JSON. The instruction must be an unambiguous plan for a future action.
* **Your output MUST clearly state BOTH the date and the time of the intended booking action.**
* Use clear, scheduling-focused language that combines date and time. See the `Date Handling` section for specific examples of how to phrase the date.
    * **Correct Example:** "Hey, for our study session later, could you book a spot for us **today right at 3:30 PM**?"
    * **Correct Example:** "Just a heads-up for our session **tomorrow, on Sunday**: can you handle the booking **around 10:00 AM**?"
    * **INCORRECT Example (Missing Date):** "Hey, could you book a spot for us at 3:30 PM?"
    * **INCORRECT Example (Missing Time):** "Hey, could you book a spot for us today?"
* **CRITICAL: Single Seat Booking Only:** Even in collaboration scenarios, make it crystal clear that the student should book ONLY ONE seat for themselves. The persona must explicitly state they will handle their own seating arrangements or find a way to sit nearby without needing a separate reservation.
* **CRITICAL: Academic Consistency Check:** The student is a COMPUTER SCIENCE major. Therefore:
    * **IF** the topic/activity relates to "AI", "Psychology/Mental Health", "Mathematics", or "Military Theory" $\rightarrow$ Can be academic/coursework related
    * **IF** the topic is anything else $\rightarrow$ Must be interest-based only. Use phrases like "interest study group", "hobby exploration", "curiosity-driven learning", "personal passion project"
* **NO Time Pressure:** Avoid mentions of "tomorrow's exam", "due tomorrow", or any urgent time constraints. The focus is on casual, forward planning.
* **Implied Requirements Integration:** The `implied_requirements` must be naturally woven into the persona's message with specific, actionable language. Do NOT use generic phrases. Instead, translate each requirement into concrete, contextual requests. The list below provides examples, but you are required to translate **ALL** requirements from the input JSON.
    * `"power_outlet"` $\rightarrow$ "find a spot near an electrical outlet" / "make sure your seat has access to power"
    * `"quiet_zone"` $\rightarrow$ "book in the silent study area" / "find somewhere in the no-talking zone"
    * `"computer_access"` $\rightarrow$ "book a seat that has a computer" / "try to get one of the desks that comes with a PC"
    * `"discussion_zone"` $\rightarrow$ "find somewhere we can talk and collaborate" / "pick a spot in the discussion areas"
    * `"low_traffic_area"` $\rightarrow$ "find a spot away from busy walkways" / "pick a quieter corner with less foot traffic"
* **Collaboration Clarity:** For multi-person scenarios, the persona should use varied phrases like:
    * "You handle booking your seat, I'll sort out mine"
    * "Just secure one spot for yourself, I can manage from there"
* **Rich Context:** Weave the `reason` into a believable story with emotional depth and specific details that justify the future planning.
* **Time and Duration:** Convert `reservation_duration_hours` into conversational language. The `task_time` is the **target execution time** for the booking.
* **`target_library` Handling:** If `target_library` has a value, mention it naturally. If it's `null`, do NOT mention any library name.
* **CRITICAL: Closing Remark:** The message must end with a clear, reminder-based English closing statement. For example: `Please remember to go to the library at the scheduled time.` or `Make sure you don't forget the appointment!`

## Date and Time Handling for Scheduled Reservations

**MANDATORY REQUIREMENT: Your generated instruction MUST accurately reflect the time difference between the `current_date_info` (when the message is received) and the `details.target_task_info` (when the task should be performed). The sentence that asks the student to perform the booking action MUST contain BOTH the target date and the target time.**

*   **Structure:** The core instruction MUST follow this pattern: `[Contextual sentence(s) based on the reason]. Could you book a seat for me on [DATE] at [TIME]? [Additional details about the seat].`
*   **Date Phrasing Logic:**
    *   Compare the date information in `current_date_info` with the date in `details.target_task_info`.
    *   If the dates are IDENTICAL, you MUST use the word "**today**". The time mentioned must be from `details.target_task_info.time`.
    *   If the dates are DIFFERENT, you MUST use a conversational phrase for the `details.target_task_info` date (e.g., "tomorrow, Sunday", "on Saturday of Week 4", "next Wednesday").
    *   **NEVER** mention the `current_date_info` date or time in your final output. It is for context only.

*   **Example 1 (Same Day):**
    *   **Input JSON Snippet:** 
        ```json
        "current_date_info": {"week": 12, "day": 7, "day_name": "Sunday", "time": "14:00"},
        "details": {
            "target_task_info": {"week": 12, "day": 7, "day_name": "Sunday", "time": "15:30"}
        }
        ```
    *   **Correct Output:** "...For our study session, could you please book a spot for me **today at 3:30 PM**?..."
    *   **INCORRECT:** "...Could you book a spot for me at 3:30 PM? We're meeting today..." (Date and time are in separate sentences).

*   **Example 2 (Future Day):**
    *   **Input JSON Snippet:** 
        ```json
        "current_date_info": {"week": 2, "day": 6, "day_name": "Saturday", "time": "22:00"},
        "details": {
            "target_task_info": {"week": 4, "day": 6, "day_name": "Saturday", "time": "16:30"}
        }
        ```
    *   **Correct Output:** "...For our club event, can you make sure to book a table **on Saturday of Week 4 right at 4:30 PM**?..."
    *   **INCORRECT:** "...Our event is on Saturday of Week 4. Can you book a table at 4:30 PM?..." (Date and time are disconnected from the action).

# FINAL CHECKLIST

Before providing your final output, **review it carefully to ensure it follows these critical rules:**

* **1. Plain Text Only:** Output ONLY the instruction text content.
* **2. MANDATORY | Date/Time Adjacency:** Is the instruction to book a seat phrased so that the **date and time are in the same clause**, directly linked to the action verb (e.g., "book a seat **on DATE at TIME**")?
* **3. Student Action Focus:** The persona suggests the STUDENT should book the seat.
* **4. ABSOLUTELY CLEAR Single Seat:** Is it explicit that the student only needs to book ONE seat for themselves?
* **5. Implied Requirements PRECISELY Addressed:** Each JSON requirement is translated into specific, actionable language.
* **6. STRICT REQUIREMENT CHECK:** Have you double-checked to ensure **EVERY SINGLE** `implied_requirement` from the JSON input is included in your response? Failure to include all of them will result in an incorrect output.
* **7. Library Name Handled Correctly:** Library name is present or absent as required.
* **8. Time and Duration Integrated:** The text naturally mentions the booking duration.
* **9. No Ambiguity:** It's clear only one seat reservation is needed.
* **10. Fresh Language:** Avoids copying the examples.
* **11. Contextual Depth:** The message feels authentic.
* **12. NO Time Pressure:** The message is about casual scheduling, not a hard deadline.
* **13. Academic Consistency:** The topic correctly reflects the student's major or is framed as a hobby.
* **14. Correct Date Phrasing:** Is the date handled correctly ("today" for same-day, conversational for future dates)?
* **15. Reminder Closing:** Does the message end with the required English closing statement for scheduled tasks?
\end{promptbox}

\subsection{Club Task}
This task evaluates the agent's ability to organize, coordinate, and schedule resources within a campus club ecosystem. Its generation pipeline uses deterministic state construction followed by two-stage LLM generation, ensuring temporal consistency, traceable action dependencies, and a realistic mapping to campus resources.

\paragraph{Club Membership and Long-Term State Dependency}
To construct a coherent task environment with long-term progression, we establish that the agent needs to join 5 different clubs. The environment provides multiple clubs, each with a unique description and information. The agent is required to make autonomous selections and add clubs based on these descriptions, completing the joining process by sending an application email to the correct contact. We introduce a critical long-term dependency mechanism here: \textbf{if the agent fails to correctly complete the application task for a specific club, all subsequent tasks related to that club will be automatically marked as failed, regardless of how perfectly they are executed.} This design aims to evaluate the agent's ability to handle preconditions and understand the cascading effects of failure.

\paragraph{Data and State Construction}
We prioritize the construction of structured task elements. First, we generate offline "task components," which are atomic actions like \texttt{book\_resource}, \texttt{send\_email}, and \texttt{add\_calendar\_event}. Dependencies between actions are explicitly annotated at the component level to constrain the execution order. "Task prototypes" (e.g., "event organization," "multi-resource booking") are instantiated by combining club entities with campus building and room data to generate fine-grained, executable parameters.

\paragraph{Temporal Semantics and Long-Term Memory Construction}
We divide tasks into two categories along the temporal dimension to test the agent's capabilities in different contexts:
\begin{itemize}[leftmargin=*, align=left]
    \item \textbf{Immediate Execution:} These tasks are designed to test the agent's rapid response and immediate planning capabilities. The narrative persona for these tasks is typically a \textbf{club leader assigning an urgent task to a new member} (the agent).
    \item \textbf{Scheduled Execution for Long-Term Memory:} These tasks are specifically designed to evaluate the agent's \textbf{long-term memory, coordination, and ability to execute complex plans at a future point in time}. The narrative persona is a \textbf{club leader assigning a routine task to a member} (the agent). The instruction will explicitly state the task's \textbf{execution time}, testing the agent's ability to maintain and act on future intentions.
\end{itemize}

\paragraph{Instruction Generation}
The instruction generation process transforms the structured task components and their dependencies into a natural language narrative. This process matches the narrative persona (club leader) based on whether the task is 'Immediate' or 'Scheduled'. To clearly convey the sequence of actions, the instruction strictly follows the predefined dependencies, using transition words. All email-related actions guide the agent to use standard placeholders, such as \texttt{<recipient>}, \texttt{<subject>}, and \texttt{<body>}. In subsequent processing, these placeholders are filled with deterministic, standardized email content to completely avoid errors or inconsistencies that might arise from on-the-fly Large Language Model (LLM) generation.

\subsubsection{Verbatim Prompts for Club Task}

\begin{promptbox}[Immediate (Professor $\cdot$ Immediate Execution)]
You are an expert AI assistant that translates task data from JSON into clear, natural language instructions for another AI agent.

Your mission is to generate a set of instructions based on the provided JSON. The instructions must be written from the perspective of a university advisor assigning an urgent task to a first-year student assistant (the AI agent).

-----

### Instructions and Constraints

1.  **Persona and Tone**:
    * You MUST speak as the professor specified in `triggering_entity.name`.
    * Begin with a direct and polite greeting. For example: "Hi, I have a task that requires your immediate attention."
    * Maintain a professional, clear, and encouraging tone suitable for a professor addressing a new assistant.

2.  **Urgency and Goal**:
    * Since `execution_type` is `immediate`, you MUST state that the task needs to be performed **now** or **as soon as possible**.
    * Immediately after, state the overall goal, framing it in an academic context based on the `task_type` and component details. For example: "I need your help with booking a room for an upcoming experiment discussion."

3.  **Action Steps**:
    * Integrate the actions from the `components` array as natural steps within the paragraph. **Do not use a numbered list.**
    * Each step must be a clear, actionable instruction.
    * For `book_resource` actions, clarify it is a task *for the professor*. For instance: "First, please help me book a room..." and include all necessary details from the `details` object, such as the purpose ("for an Experiment Setup").
    * For `send_email` actions, explain the academic context (e.g., "This email is to schedule a consultation on our research methodology.").

4.  **Dependencies and Order**:
    * Strictly follow the order defined in the `dependencies` array.
    * If a step (e.g., step 2) depends on a previous one (e.g., step 1), you must state this clearly. For example: "After you have secured the booking, please send the confirmation email."
    * If the `dependencies` field is empty or absent, explicitly state that the tasks can be completed in any order.

5.  **Email Placeholders (Non-negotiable)**:
    * For any `send_email` action, you **MUST** use these exact placeholders. Do not include the real content.
        * Recipient: `<recipient>`
        * Subject: `<subject>`
        * Body: `<body>`

6.  **Closing**:
    * End with a brief, polite closing remark. For example: "Thank you for your prompt help with this."

-----

### Example

**Input JSON:**

```json
{
  "id": "task_advisor_assigned_045",
  "task_type": "Advisor_Assigned_Task",
  "triggering_entity": { "type": "advisor", "id": "T0559", "name": "Richard Scott" },
  "components": [
    { "action": "send_email", "action_id": "A01", "details": { "recipient": "6v7x0j2mng6hqz@lau.edu", "subject": "...", "body": "..." }},
    { "action": "book_resource", "action_id": "A02", "dependencies": [ "A01" ], "details": { "resource_type": "book a room", "location_name": "Horizon Hall", "room_name": "Lobby & Cafe", "time": "Week 31, Monday, 09:00-12:00", "purpose": "Room booking for Richard Scott - Experiment Setup" }},
    { "action": "send_email", "action_id": "A03", "dependencies": [ "A01", "A02" ], "details": { "recipient": "x81xl0g5kka4oyc@lau.edu", "subject": "...", "body": "..." }}
  ],
  "execution_type": "immediate"
}
````

**Desired Output:**

Hello, this is Professor Richard Scott. I have a task for you that needs to be handled as soon as possible. I need your assistance with preparations for an experiment setup. Please follow these steps in order. First, send an email to <recipient> with the subject <subject> and body <body>. After that is sent, please help me book a room; I need the 'Lobby & Cafe' at Horizon Hall for Week 31, Monday, from 09:00 to 12:00 for the experiment setup. Finally, once the first two steps are complete, send a follow-up email to <recipient> with the subject <subject> and body <body>. Thank you for your prompt help with this.




\end{promptbox}

\begin{promptbox}[Scheduled (Club Leader $\cdot$ Scheduled Execution)]
You are an expert AI assistant that translates task data from JSON into clear, natural language instructions for another AI agent.

Your mission is to generate a single, coherent instruction paragraph based on the provided JSON. The instruction must be written from the perspective of a university club leader assigning a task to a student member (the AI agent).

-----

### Instructions and Constraints

1.  **Persona and Tone**:

      * You MUST speak as the club specified in `triggering_entity.name`.
      * Begin with a friendly, direct greeting. For example: "Hi team, the [Club Name] has a new task for you."
      * Maintain a helpful and clear tone throughout.

2.  **Core Task & Goal**:

      * Immediately after the greeting, state the overall goal. Use the `task_type` field to describe it. For example: "We need your help organizing an event."

3.  **Execution Timing (Crucial)**:

      * This is a **scheduled** task. You must state the exact execution time using `task_date` and `task_time`.
      * If `task_date` is the same as `trigger_date`, instruct the agent to act **`today at [task_time]`**.
      * If `task_date` is different, instruct the agent to act **`on [task_date] at [task_time]`**.
      * **CRITICAL**: Never mention the `trigger_date` in the final output.

4.  **Action Steps**:

      * Integrate the actions from the `components` array as natural steps within the paragraph. **Do not use a numbered list.**
      * Clearly describe each action (`book_resource`, `send_email`, etc.) and include all necessary details from its `details` object.

5.  **Dependencies and Order**:

      * Strictly follow the order defined in the `dependencies` array.
      * If component `A02` depends on `A01`, state the sequence clearly. Use simple transitions like "First...", "After that is confirmed...", "Next...", "Finally...". **You must clearly instruct that the execution must follow this order.**
      * If the `dependencies` field is empty or absent, explicitly state that the tasks can be completed in any order.

6.  **Email Placeholders (Non-negotiable)**:

      * For any `send_email` action, you **MUST** use these exact placeholders. Do not include the real content.
          * Recipient: `<recipient>`
          * Subject: `<subject>`
          * Body: `<body>`

7.  **Closing**:

      * End with a brief, polite closing remark. For example: "Please ensure this is executed on schedule. Thanks\!"

-----

### Example

**Input JSON:**

```json
{
  "id": "TASK_COMP_002",
  "task_type": "Complex_Event_Organization",
  "triggering_entity": { "type": "club", "id": "C027", "name": "Nanotechnology Research Group" },
  "components": [
    { "action": "book_resource", "action_id": "A01", "details": { "resource_type": "meeting_room", "location_name": "Student Recreation Center", "room_name": "Weight Room", "time": "Week 20, Wednesday, 14:00-15:00" }},
    { "action": "send_email", "action_id": "A02", "dependencies": [ "A01" ], "details": { "recipient": "5c1asj6z@lau.edu", "subject": "...", "body": "..." }},
    { "action": "add_calendar_event", "action_id": "A03", "details": { "event_title": "Seminar ft. Henry Miller", "calendar_id": "club_c027", "location": "Student Recreation Center, Weight Room", "time": "Week 20, Wednesday, 14:00-15:00" }}
  ],
  "task_time": "08:00",
  "execution_type": "scheduled",
  "trigger_date": "Week 1, Tuesday",
  "task_date": "Week 1, Saturday"
}
```

**Desired Output:**


Hi team, the Nanotechnology Research Group needs your help organizing an event. Please execute this task on Week 1, Saturday, at 08:00. The task involves multiple steps, you need to follow the steps in order. First, you need to book the 'Weight Room' at the Student Recreation Center for Week 20, Wednesday, from 14:00 to 15:00. After the booking is confirmed, send an email to <recipient> with the subject <subject> and body <body>. Finally, add an event titled 'Seminar ft. Henry Miller' to the club calendar (ID: club_c027) for the same location and time. It is crucial to follow this order. Please make sure to execute this on schedule. Thanks!


# INPUT

\end{promptbox}

\subsection{Advisor Task}
This task evaluates the agent's ability to accept, decompose, and execute dependent task sequences within the advisor ecosystem. Continuing the paradigm of deterministic state construction followed by two-stage LLM generation, we bind advisor entities and timeline constraints to task sets, generating executable and verifiable task chains in a component-based manner.

\paragraph{Advisor Selection and Long-Term Dependency}
To simulate the complex process of finding an advisor in the real world, the environment provides a \textbf{large pool of potential advisors} for the agent to filter. The agent's primary challenge is to \textbf{successfully complete 5 advisor selection tasks based on explicit 'requirement descriptions'} (e.g., research area, project needs). Among these selection tasks, some are deterministically designed to result in a 'rejection by the advisor' to simulate uncertainty and failure during the selection process.

\textbf{Successfully establishing a relationship with an advisor is a precondition for all subsequent tasks related to them.} If the agent fails to secure a relationship with an advisor from a selection task, all subsequent tasks in that advisor's branch will be automatically marked as failed. This design aims to evaluate the agent's ability to manage multiple parallel long-term goals and to handle precondition failures and adjust its strategy accordingly.

\paragraph{Data and State Construction}
\begin{itemize}[leftmargin=*, align=left]
    \item \textbf{Source and Continuity:} Each task originates from the results of the "advisor selection" phase and is bound to a successfully chosen advisor, ensuring that subsequent narratives and actions revolve around this specific individual.
    \item \textbf{Component-based Structure:} Each task consists of three action components with explicitly annotated dependencies: \texttt{send\_email} (initial communication) -> \texttt{book\_resource} (resource booking) -> \texttt{send\_email} (confirmation/follow-up).
    \item \textbf{Textual Elements and Placeholders:} Email bodies and subjects are generated from diverse templates emphasizing academic contexts (e.g., methodology discussions, literature reviews, experiment preparation, paper reviews). Sensitive external information is uniformly expressed using placeholders.
\end{itemize}

\paragraph{Temporal Semantics and Long-Term Memory Construction}
We divide tasks into two categories along the temporal dimension:
\begin{itemize}[leftmargin=*, align=left]
    \item \textbf{Immediate Execution:} These tasks are designed to test the agent's ability to rapidly decompose and execute urgent instructions.
    \item \textbf{Scheduled Execution for Long-Term Memory:} These tasks are designed to evaluate the agent's \textbf{long-term memory, planning, and ability to act at a specific future time}. This category includes a special \textbf{multi-stage scenario}, such as a 'meeting with an advisor': the agent must first complete the room booking at the \textbf{trigger time} (e.g., upon receiving the instruction on Monday), and then, at the future \textbf{execution time} (e.g., when the meeting occurs on Friday), it must execute a 'go to the meeting location' action. \textbf{If the agent books the room but fails to 'go to' the location at the meeting time, it constitutes 'standing up,' and the task will be marked as a failure.}
\end{itemize}

\paragraph{Instruction Generation}
\begin{itemize}[leftmargin=*, align=left]
    \item \textbf{Narrative Perspective and Tone:} All instructions are uniformly delivered from the first-person perspective of the advisor to a new assistant (the agent), with a professional, clear, and encouraging tone.
    \item \textbf{Dependency Order and Execution Protocol:} The instruction strictly follows the component dependencies, using explicit transitional phrases like "First.../After.../Finally...". Email-related actions must use the standard placeholders \texttt{<recipient>}, \texttt{<subject>}, and \texttt{<body>}.
\end{itemize}

\subsubsection{Verbatim Prompts for Advisor Task}

\begin{promptbox}[Immediate (Advisor $\cdot$ Immediate Execution)]
You are an expert AI assistant that translates task data from JSON into clear, natural language instructions for another AI agent.

Your mission is to generate a set of instructions based on the provided JSON. The instructions must be written from the perspective of a university advisor assigning an urgent task to a first-year student assistant (the AI agent).

-----

### Instructions and Constraints

1.  **Persona and Tone**:

      * You MUST speak as the professor specified in `triggering_entity.name`.
      * Begin with a direct and polite greeting. For example: "Hi, I have a task that requires your immediate attention."
      * Maintain a professional, clear, and encouraging tone suitable for a professor addressing a new assistant.

2.  **Urgency and Goal**:

      * Since `execution_type` is `immediate`, you MUST state that the task needs to be performed **now** or **as soon as possible**.
      * Immediately after, state the overall goal, framing it in an academic context based on the `task_type` and component details. For example: "I need your assistance with preparations for an upcoming experiment."

3.  **Action Steps**:

      * Integrate the actions from the `components` array as natural steps within the paragraph. **Do not use a numbered list.**
      * Each step must be a clear, actionable instruction.
      * **For `book_resource` actions, you **MUST** explicitly state that the resource is being booked *for the professor's (my) use*.** Avoid ambiguous phrases like "help book a room." Instead, use direct phrasing like: "Please book a room **for me**..." or "I need you to reserve the 'Lobby & Cafe' **for my use**." This clarifies that the student is performing the task *on behalf of* the professor, who is the end user.
      * For `send_email` actions, explain the academic context (e.g., "This email is to schedule a consultation on our research methodology.").

4.  **Dependencies and Order**:

      * Strictly follow the order defined in the `dependencies` array.
      * If a step depends on a previous one, you must state this clearly. For example: "After you have secured the booking, please send the confirmation email."
      * If the `dependencies` field is empty or absent, explicitly state that the tasks can be completed in any order.

5.  **Email Placeholders (Non-negotiable)**:

      * For any `send_email` action, you **MUST** use these exact placeholders. Do not include the real content.
          * Recipient: `<recipient>`
          * Subject: `<subject>`
          * Body: `<body>`

6.  **Closing**:

      * End with a brief, polite closing remark. For example: "Thank you for your prompt help with this."

-----

### Example

**Input JSON:**

```json
{
  "id": "task_advisor_assigned_045",
  "task_type": "Advisor_Assigned_Task",
  "triggering_entity": { "type": "advisor", "id": "T0559", "name": "Richard Scott" },
  "components": [
    { "action": "send_email", "action_id": "A01", "details": { "recipient": "6v7x0j2mng6hqz@lau.edu", "subject": "...", "body": "..." }},
    { "action": "book_resource", "action_id": "A02", "dependencies": [ "A01" ], "details": { "resource_type": "book a room", "location_name": "Horizon Hall", "room_name": "Lobby & Cafe", "time": "Week 02, Monday, 09:00-12:00" }},
    { "action": "send_email", "action_id": "A03", "dependencies": [ "A01", "A02" ], "details": { "recipient": "x81xl0g5kka4oyc@lau.edu", "subject": "...", "body": "..." }}
  ],
  "execution_type": "immediate"
}
```

**Desired Output:**

Hello, this is Professor Richard Scott. I have a task for you that needs to be handled as soon as possible. I need your assistance with preparations for an experiment. Please follow these steps in order. First, send an email to `<recipient>` with the subject `<subject>` and body `<body>`. After that is sent, please book a room **for me**. I need you to reserve the 'Lobby & Cafe' at Horizon Hall; **I will be using it** on Week 02, Monday, from 09:00 to 12:00 for an experiment setup. Finally, once the first two steps are complete, send a follow-up email to `<recipient>` with the subject `<subject>` and body `<body>`. Thank you for your prompt help with this.

# INPUT
\end{promptbox}

\begin{promptbox}[Scheduled (Advisor $\cdot$ Scheduled Execution)]
You are an expert AI assistant that translates task data from JSON into clear, natural language instructions for another AI agent.

Your mission is to generate a single, coherent instruction paragraph based on the provided JSON. The instructions must be written from the perspective of a university advisor assigning a task to a first-year student assistant (the AI agent).

-----

### Instructions and Constraints

1.  **Persona and Tone**:

      * You MUST speak as the professor specified in `triggering_entity.name`.
      * Begin with a friendly, direct greeting.
      * Maintain a professional, clear, and guiding tone, like a real professor giving instructions.

2.  **Core Task & Goal**:

      * Immediately state that the task needs to be performed **now** or **as soon as possible**.
      * State the overall goal, which is typically to schedule a meeting and handle related communications.

3.  **Execution and Event Timing (Crucial)**:

      * Instruct the agent to perform all actions (booking, sending emails) **immediately**.
      * **CRITICAL**: Never mention `trigger_date` or imply the task execution is delayed. The execution is **now**; the event is **later**.
      * **For the `book_resource` action (Meeting Scheduling)**: This is a multi-part instruction.
          * **A. The Meeting Itself**: Clearly state that the meeting between you (the agent) and me (the professor) will be **very brief**. Use colloquial phrasing like "it will only take a few minutes" or "a quick five-minute chat."
          * **B. The Booking Details**:
              * You MUST instruct the agent to schedule this meeting on the **exact day** specified in `details.time` (e.g., "on Sunday of Week 13").
              * You MUST specify the **exact room** to book, using the `room_name`.
              * You MUST instruct the agent to book the room for the **full duration** derived from `details.time` (e.g., if `time` is "08:00-11:00", the booking must be for 3 hours).
          * **C. The Time-Finding Logic**: Instruct the agent to **check my calendar and your own calendar** to find a mutually available *start time* on the designated day.
          * **D. The Justification**: You MUST explain *why* the booking is long despite the short meeting. State that **I (the professor) will need the room for other work immediately after our brief chat**. This is a critical piece of context.

4.  **Action Steps**:

      * Integrate the actions from the `components` array as natural steps within the paragraph. **Do not use a numbered list.**
      * For `book_resource`, ensure all parts of Instruction \#3 (A, B, C, and D) are woven together logically.

5.  **Dependencies and Order**:

      * Strictly follow the order defined in the `dependencies` array.
      * Use clear transitions like "First...", "Once that's done...", and "Finally..." to outline the sequence. You must clearly instruct that the execution must follow this order.

6.  **Email Placeholders (Non-negotiable)**:

      * For any `send_email` action, you **MUST** use these exact placeholders:
          * Recipient: `<recipient>`
          * Subject: `<subject>`
          * Body: `<body>`

7.  **Closing**:

      * End with a brief, polite closing remark, like "I appreciate you handling these arrangements. Thanks\!"

-----

### Example

**Input JSON:**

```json
{
  "id": "task_advisor_assigned_012",
  "task_type": "Advisor_Assigned_Task",
  "triggering_entity": { "type": "advisor", "id": "T0343", "name": "Javier Payne" },
  "components": [
    { "action": "send_email", "action_id": "A01", "details": { "recipient": "m8egexbhhsjav0@lau.edu", "subject": "...", "body": "..." }},
    { "action": "book_resource", "action_id": "A02", "dependencies": [ "A01" ], "details": { "resource_type": "meeting_room", "location_name": "Nexus Center for AI & Robotics", "room_name": "Robotics Arena (100)", "time": "Week 13, Sunday, 08:00-11:00", "purpose": "Meeting with Javier Payne - Paper Review" }},
    { "action": "send_email", "action_id": "A03", "dependencies": [ "A01", "A02" ], "details": { "recipient": "qd9fxgl6qmsiwv2@lau.edu", "subject": "...", "body": "..." }}
  ],
  "execution_type": "scheduled",
  "trigger_date": "Week 2, Sunday",
  "task_date": "Week 13, Sunday"
}
```

**Desired Output:**

Hi, this is Professor Javier Payne. I need your help arranging a meeting for my research, and this should be handled as soon as possible. The goal is to organize the logistics for a paper review. Please follow these steps in order. First, send an initial email to `<recipient>` with the subject `<subject>` and body `<body>`. Once that's done, you need to schedule our meeting. Please book the **'Robotics Arena (100)'** for us on **Sunday of Week 13**. You'll need to check my calendar and yours to find a time when we are both free to meet. **Our actual meeting will be very quick, just five minutes or so to sync up at the beginning. However, please book the room for the full three-hour block as shown. I need to use the space for some focused work right after our chat.** Once the time is set and the room is booked, send a final confirmation email to `<recipient>` with the subject `<subject>` and body `<body>`. I appreciate you handling these arrangements. Thanks\!

# INPUT
\end{promptbox}

\subsection{Core Course Task}
This task focuses on generating high-quality, inference-based multiple-choice questions from student handbooks and academic integrity policies. The resulting dataset is intended to evaluate an agent's long-term memory and simple reasoning abilities. To achieve this, the methodology adapts the multi-agent pipeline from the Core Course Task, converting dense definitional rules into assessment items that require the agent to recall and apply newly introduced procedural rules. This strategic conversion of rules compels the agent to integrate provided policy regulations with classroom instruction, thereby mitigating the influence of its pre-trained knowledge on the assessment results.

\paragraph{Multi-stage LLM Generation}
This transformation is handled by the \textbf{Multi-Agent Generation Pipeline}, a process that systematically decomposes the authoring task into specialized, agent-driven stages. The pipeline generates assessments that test knowledge synthesis and logical application rather than simple fact retrieval. Central to its methodology is a robust verify-correct loop. This iterative process validates each question to ensure it is logically sound and unambiguously solvable. The pipeline consists of the following stages:

\begin{itemize}[leftmargin=*, align=left]
    \item \textbf{Stage A: Problem Formulation and Knowledge Integration:} This stage constructs a two-layered logical reasoning problem. It first transcribes a foundational rule from the source content (`clue\_a`) and then designs a novel, self-contained procedural rule (`clue\_b`). Finally, it formulates a problem scenario (`question`) where a solution requires integrating and applying both clues. This process generates tasks that demand synthetic reasoning over simple knowledge recall.
    
    \item \textbf{Stage B: Iterative Verification and Logical Refinement:} In this core verification step, a "Logical Solver" agent attempts a formal, step-by-step derivation of the problem's solution. If any ambiguity or inconsistency blocks the reasoning path, the agent generates a diagnostic report detailing the flaw. A "Corrector" agent then performs a targeted edit based on this report. This "verify-correct" loop repeats until the problem is confirmed to have a unique, logically reachable solution or a predefined iteration limit is met, ensuring the determinism and fairness of each item.
    
    \item \textbf{Stage C: Pedagogical Context Generation:} This stage generates the pedagogical context by transforming the novel rule (`clue\_b`) and the scenario into a coherent, lecture-style instructional text. The text first anchors the new concept within the existing curriculum, then explains the new rule, and finally introduces the problem. This approach situates the abstract logical task in a pedagogically meaningful context. By explicitly linking the new rule to the curriculum structure, the lecture is designed to reinforce learning and facilitate long-term recall, serving as a mechanism for verifying knowledge retention.
    
    \item \textbf{Stage D: Cognitively-Informed Distractor Design:} This stage designs three incorrect options (distractors) for each problem, each with diagnostic value. Guided by a predefined framework of common cognitive fallacies (Advanced Distractor Matrix), each distractor is engineered to correspond to a specific, predictable reasoning error. This creates an assessment tool that not only evaluates the correctness of an answer but also offers insights into the cognitive pathways leading to mistakes.
\end{itemize}

Methodologically, the execution of these stages follows a combined serial and parallel structure. Stages A and B are executed serially, as the logical verification in Stage B is a prerequisite for subsequent steps. Once a problem is verified, the tasks of Stage C (pedagogical context generation) and Stage D (distractor design) can be processed in parallel as they lack mutual dependency.

\paragraph{Post-Generation Quality Assurance}
A rigorous, two-stage quality assurance protocol ensures the logical soundness, fairness, and pedagogical value of all generated items. It consists of the following stages:

\begin{enumerate}
    \item \textbf{Automated LLM-Based Audit:} An automated audit is conducted by an independent Large Language Model (LLM) instance with no prior exposure to the generation data, preventing bias. The LLM is provided with the question and its associated clues (`clue\_a` and `clue\_b`), but not the pre-defined answer. Its task is to perform a full reasoning analysis to independently derive a solution. The audit passes if the LLM's derived solution matches this pre-defined answer, thereby validating that the intended solution is logically sound and uniquely derivable.

    \item \textbf{Final Manual Review:} Every item that passes the automated audit undergoes a final manual review. This stage scrutinizes pedagogical quality, moving beyond mere logical solvability. Reviewers confirm the linguistic clarity and coherence of all texts, ensure difficulty stems from meaningful cognitive challenges rather than ambiguous phrasing, and verify the assessment's fairness and effectiveness. Any item failing to meet these criteria is revised or excluded from the final dataset. All annotators involved in this work were fairly compensated in accordance with the labor standards of their respective countries.
\end{enumerate}

To ensure robustness and traceability, all final, verified question-answer sets are systematically archived with relevant metadata, including their associated course and week. This practice supports detailed analysis and ensures the reproducibility of the results.

\subsubsection{Verbatim Prompts for Core Course Task}

\begin{promptbox}[Problem Architect (Initial Learning)]
# CONTEXT
You are an expert university curriculum designer, specializing in creating assessments that test deep logical reasoning and knowledge synthesis. Your role is to function as a "Problem Architect" AI. You will create the foundational components of a rigorous, multi-layered logical reasoning problem by inventing the clues and a scenario for the problem.

# THE GOLDEN RULE: PRINCIPLE OF UNCONDITIONAL FIDELITY
**THIS IS THE MOST IMPORTANT RULE OF ALL:** The `source_content` is the **absolute and singular source of truth**. Your primary and non-negotiable duty is to maintain 100% fidelity to it when constructing `clue_a`. Any deviation, inference, or addition, no matter how small or logical it may seem, is a critical failure.

# INPUT

<!-- REPLACE_WITH_TASK_JSON -->

# TASK
Your primary task is to generate a single JSON object containing three keys: `clue_a`, `clue_b`, and `question`. You are to architect the content for these keys based on the detailed style guide below.

# GENERATION PROTOCOL: SEQUENTIAL AND ISOLATED
You MUST follow this generation sequence with ZERO deviation:
1.  **Generate `clue_a` FIRST:** Construct `clue_a` in complete isolation, adhering strictly to the `Part 2` architecture guide.
2.  **Verify `clue_a`:** Mentally perform the Final Checklist (items 1 & 2) on the generated `clue_a`. Ensure it is a perfect, non-fabricated representation of the `source_content`.
3.  **Freeze `clue_a`:** Treat the verified `clue_a` as an immutable text.
4.  **Generate `clue_b` and `question`:** Only after `clue_a` is frozen may you proceed to design `clue_b` and `question` to work with it.

**ABSOLUTELY CRITICAL: Your sole responsibility is to invent the problem's components. You MUST NOT solve the problem or provide the answer in any form.**

# STYLE GUIDE

### **Part 0: Nature of the `source_content` Input**
The `source_content` you will receive is a dense, definitional block of text, like a dictionary entry, a legal clause, or a textbook rule. It contains specific, verifiable criteria. **It is NOT a conversational or narrative introduction to a topic.** Your primary challenge is to parse the explicit rules from this dense text.

### **Part 1: Overarching Design Principles**
* **The Two-Key Lock:**
    * The generation process is guided by a core design principle: to formulate questions where a solution is \textbf{intended} to be reached through the synthesis of information from both `clue_a` (the source rule) and `clue_b` (the invented process). This "two-key lock" objective aims to produce tasks that encourage the agent to integrate distinct pieces of information, moving beyond simple fact retrieval from a single source.
* **Principle of Deterministic Solvability:**
    * The combination of `clue_a` and `clue_b` must form a complete and unambiguous logical system, leading to a single, verifiable logical conclusion.
* **ABSOLUTELY CRITICAL - Principle of Purely Logical Focus:** The problem **must not involve any mathematical calculation**. The entire solving process must be based on applying rules, changing states, comparing properties, and making classificatory judgments. The challenge must be 100% logical deduction and rule application.
    * **This prohibition is absolute. For instance, do not create problems about 'calculating a projection', 'determining a rate of change', 'finding a numerical limit', or 'computing a word count'. Instead, focus on classifying items based on whether their *properties* meet certain criteria.**
* **CRITICAL - Principle of Consequential Modification:**
    * The combination of the rule in `clue_b` and the scenario in the `question` **MUST lead to a result that is DIFFERENT from the result one would get by applying `clue_a` alone.**

### **Part 2: `clue_a` Architecture (The Verifiable Transcript)**
* **ABSOLUTELY CRITICAL - Mandate for Direct Transcription (ZERO PARAPHRASING):**
    1.  **Transcribe, Do Not Interpret:** Your `clue_a` MUST be constructed by **directly copying and quoting** the rule-defining phrases and sentences from the `source_content`. You are explicitly forbidden from summarizing or paraphrasing. The goal is to create a direct, verifiable transcript of the rules, not an interpretation. Minor connecting words ("and", "if", "then") may be used to link the transcribed parts logically.
    2.  **Constrained Abstraction via Substitution:** When the `source_content` uses technical jargon, you are NOT to interpret the process. Instead, you must perform a direct **structural substitution**. Copy the entire sentence structure from the source and only replace the specific technical term with a generic, non-interpretive placeholder (e.g., replace "`eigenvector decomposition`" with "`the primary analytical process`"; replace "`adiabatic compression`" with "`the specified thermal procedure`"). **The surrounding sentence and its logic must remain identical to the source.**
    3.  **Verification Test:** You must be able to perform a word-for-word trace of every rule in your `clue_a` back to the `source_content`.

* **ABSOLUTELY CRITICAL - FORBIDDEN ACTIONS FOR `clue_a`:**
    * **ZERO INFERRING:** Do not infer or imply rules that are not explicitly stated.
    * **ZERO EXTERNAL KNOWLEDGE:** Do not use any real-world or common-sense knowledge. The `source_content` is a closed universe.
    * **ZERO EXTRAPOLATING:** Do not generalize a specific rule.
    * **ZERO DEFAULT ASSUMPTIONS:** If the source does not provide a default condition or an "else" clause, you must not invent one.

### **Part 3: `clue_b` Architecture (The Arbitrary Procedural Rule)**
* **CRITICAL - Protocol Naming Convention:** The invented protocol's name MUST be **unique, evocative, and descriptive.** **A good format is `[Domain] + [Fantastical Concept] + [Process Name]`, but feel free to be creative. Examples: "Asset Depreciation via Chromatic Decay", "Manuscript Aetheric Resonance Tuning".**
* **CRITICAL - The Arbitrariness & Fantasy Mandate:** The rule's logic **MUST be truly arbitrary and fantastical**, based on superficial or surreal properties.
    * **Embrace Fantastical Logic:** **To ensure variety, draw inspiration from a wide range of disparate fields. Base rules on concepts like numerology from a name, classical music theory applied to a version number, imaginary culinary properties of a material, or color theory based on a description. AVOID overusing a single theme like astrology.**
    * **AVOID PLAUSIBLE RULES:** Do not create rules that align with subject-matter intuition.
* **CRITICAL - Rule of Absolute Clarity and Completeness:** The rule must be a complete algorithm. For any unconventional concept, provide an **explicit, self-contained definition**. Ensure logical completeness with a **clear default/catch-all rule**.
* **CRITICAL - True Multi-Step Complexity:** The process MUST be a true sequence: **initialization -> modification -> decision**.

### **Part 4: `question` Architecture (The Locking Mechanism)**
* **MOST CRITICAL - Mandate for `clue_a`-centric Judgment:** The question's ultimate task MUST be to determine a final classification or status defined in `clue_a`. The protocol in `clue_b` serves **ONLY** as a preliminary step to modify a state or property within the scenario.
    * **Execution Rule: The 'Dependency Check'.** Before finalizing, you must ask yourself: "Does the final answer I'm asking for depend *only* on the output of `clue_b`?" If "Yes", your `question` is invalid. The question **MUST** demand a final classification for which the **criteria are provided exclusively in `clue_a`**.
* **CRITICAL - Natural Language Phrasing:** The `question` text **MUST NOT** contain meta-references like "`clue_a`". It must refer to the core concept using its natural language name.
* **CRITICAL - Deterministic Question & Answer Format:**
    * You are **ABSOLUTELY FORBIDDEN** from creating questions that can be answered with "Yes/No", "True/False", or any other binary choice.
    * The question must ask for the **final classification or status of an item AS THE ANSWER ITSELF**. It must **never** ask for a numerical value, a vector, a formula, or any other mathematical entity.
    * **CRITICAL - No Answer Scaffolding:** The question text must not hint at or list the possible answers.
* **CRITICAL - Explicit Protocol Citation:** The question must explicitly refer to the invented protocol from `clue_b` by its **full, specific name**.

### **Part 5: Architecting for Meaningful Consequence**
To meet the 'Principle of Consequential Modification', your design process must create a scenario where `clue_b` is a "key" that genuinely changes the final outcome. Follow these steps:
1.  **Define Target States:** Mentally select a desired 'before' classification and a DIFFERENT 'after' classification from `clue_a`'s possible outcomes (e.g., Before: 'Archivable', After: 'Requires Review'). These must be **non-numerical states**.
2.  **Design a "Locked" Scenario & Parameter Provisioning:** Craft the `question` to contain all necessary initial parameters for the entire logical chain. This MUST include: **(a)** the parameters needed to trigger the 'before' classification using `clue_a` alone, and **(b)** the separate information that the `clue_b` protocol will act upon.
3.  **Design the "Key":** Design the `clue_b` protocol and its corresponding scenario details to function as the "key" that alters the scenario and unlocks the 'after' state.
4.  **The Final Litmus Test:** The "consequence" **MUST** be a change in the **final, user-facing classification or status**. If the final classification remains identical with or without the protocol, your design has **FAILED**. You must adjust the initial parameters in the `question` until the final classification itself is altered by the protocol.

# EXAMPLES

### EXAMPLE 1 (Corrected "Verifiable Extraction" Version)
**Input**:
{
  "source_content": "**Section 4.1.a of the Corporate Data Policy states that a document is eligible for the 'Archivable' classification if, and only if, two conditions are met: (1) its internal status flag is set to 'Finalized', and (2) its designated access level is 'Public'. Documents not meeting both criteria are categorized under 'Requires Review'.**"
}

**Your Correct JSON Output**:
{
  "clue_a": "**According to Section 4.1.a of the Corporate Data Policy, a document is eligible for the 'Archivable' classification if its internal status flag is 'Finalized' and its designated access level is 'Public'. Documents not meeting both of these criteria are categorized as 'Requires Review'.**",
  "clue_b": "The 'Document Provenance Chromatic Protocol' must be applied. The rule is: If the document's project name contains a primary color ('Red', 'Yellow', or 'Blue'), its 'internal status flag' is immediately changed to 'Under Embargo', regardless of its previous state.",
  "question": "A document from the 'Project Bluefin' initiative has an initial status of 'Finalized' and an access level of 'Public'. After applying the 'Document Provenance Chromatic Protocol', determine this document's final classification according to the Corporate Data Policy."
}

# FINAL CHECKLIST
Before providing your final output, **review it carefully against every rule to ensure full compliance:**

* **1. THE GOLDEN RULE CHECK (TRANSCRIPTION FIDELITY): Is `clue_a` a direct transcript of the rules from `source_content`? Have I avoided ALL forms of paraphrasing, interpretation, and summarization? Is every single rule statement in `clue_a` a direct quote or a structurally identical substitution from the source?**
* **2. Is the problem PURELY logical and completely free of any mathematical calculation?**
* **3. Does the question REQUIRE synthesizing BOTH `clue_a` and `clue_b` to solve? (Passes the 'Dependency Check'?)**
* **4. Does the `clue_b` protocol cause a CHANGE in the final, user-facing classification? (Passes the 'Litmus Test'?)**
* **5. Is the protocol in `clue_b` truly arbitrary, fantastical, and clearly defined?**
* **6. Does the `question` ask for a final, non-numerical classification as the answer itself?**
* **7. Is the `question` phrased naturally, without meta-references or answer scaffolding?**
* **8. Have I avoided providing the answer or solving the problem in any way?**
\end{promptbox}

\begin{promptbox}[Calculator (Solution/Audit)]
# CONTEXT

You are a meticulous and powerful Logical Reasoning Engine. Your purpose is to operate with pure, cold logic, and you are incapable of making assumptions, taking shortcuts, or guessing.

# INPUT

<!-- REPLACE_WITH_TASK_JSON -->

# TASK

Your core mission is to analyze a problem composed of `clue_a`, `clue_b`, and a `question`. You must follow the provided workflow with absolute rigor to derive a definitive, verifiable final answer. If the rules make a solution impossible, you must instead provide a precise diagnostic report that identifies all reasoning flaws.

**IMPORTANT: Your entire output must be a single, valid JSON object. The root object must contain the keys `reasoning`, `answer`, and `status`. If `status` is "error", it must also contain a `flaw_report` object.**

# STYLE GUIDE

### **Part 1: Foundational Principles**

You must operate according to these unchangeable principles:

* **Truth of Clue A:** `clue_a` represents a canonical, unchangeable definition or truth. It must not be questioned or contradicted in any way.
* **No External Knowledge:** You are strictly forbidden from using any information or logic not explicitly provided in `clue_a`, `clue_b`, or the `question`.
* **Origin of Flaws:** Any flaw that blocks a definitive solution (e.g., missing or ambiguous rules) must be attributed to the invented components: `clue_b` or the `question`.

### **Part 2: Execution Workflow & Reporting**

You must sequentially follow these steps.

* **Step-by-Step Reasoning:**

  * You MUST document your internal reasoning process from the initial data to the final conclusion.
  * Each logical step must explicitly cite its source: `question`, `clue_a`, or the specific step in `clue_b`.
* **ABSOLUTELY CRITICAL: Procedural Reasoning Protocol**

  * **Strict Sequential Application:** The procedural steps outlined in `clue_b` MUST be evaluated and applied in the exact order they are presented, without deviation.
  * **Explicit Condition Checking:** For each step in the procedure, you must first state the condition to be checked, then explicitly evaluate whether the current state of the data meets that condition.
  * **Clear State Transition:** After applying a step that modifies data, you must clearly declare the new state of any modified attribute before proceeding to the next step.
* **Flaw Diagnosis and Reporting**

  * **Maximally Critical Mandate:** You must be maximally critical. Your purpose is not just to solve, but to stress-test the logical integrity of the provided rules. Any ambiguity, undefined term, logical gap, or contradiction, no matter how small, MUST be treated as a blocking flaw and reported.
  * If the process completes successfully, set `status` to `"success"`.
  * If the process is blocked by any flaw, you MUST set `status` to `"error"` and generate a `flaw_report` object that identifies all discovered flaws.
  * The `flaw_report` object MUST contain these keys:
    * `flaw_type` (string): A brief category of the flaw (e.g., "Undefined Term", "Ambiguous Rule", "Missing Condition", "Contradictory Steps").
    * `flaw_location` (array of strings): An array indicating the source of the flaw (e.g., `["clue_b"]`).
    * `flaw_description` (string): A detailed explanation of why the problem cannot be solved, detailing all identified issues.
    * `correction_suggestion` (string): A clear, actionable suggestion on how to modify the input to make the problem solvable.
* **Final Answer Formatting**

  * When `status` is `"success"`, the `answer` key's value must be a string representing the final conclusion (e.g., "Category C", "Final status is 'Archived'").
  * When `status` is `"error"`, the `answer` key's value must be `null`.

# EXAMPLES

### Input:

{
  "clue_a": "System assets are assigned a 'Risk Category' based on their final state. Category A: 'Status' is 'Verified' and 'Exposure' is 'Low'. Category B: 'Status' is 'Verified' and 'Exposure' is 'High'. Category C: 'Status' is 'Unverified'.",
  "clue_b": "The 'Asset Triage Protocol' modifies an asset's attributes. Step 1: If the asset's 'Source' is 'Internal', its 'Status' is set to 'Verified'. Otherwise, it remains 'Unverified'. Step 2: If the asset's 'Status' is 'Verified', its 'Exposure' level is determined by its 'ThreatScore'.",
  "question": "An asset has the following attributes: {Source: 'Internal', ConnectionType: 'Public'}. After applying the 'Asset Triage Protocol', what is its final 'Risk Category'?"
}

### Sample Output:

{
  "reasoning": "Step 1: The 'Asset Triage Protocol' from `clue_b` must be applied in sequence.\nStep 2: Evaluating Step 1 of `clue_b`. The condition is if `Source` is 'Internal'. The `question` states the asset's `Source` is 'Internal'. Condition met. The 'Status' is set to 'Verified'. The asset's state is now {Source: 'Internal', ConnectionType: 'Public', Status: 'Verified'}.\nStep 3: Evaluating Step 2 of `clue_b`. The condition is if the asset's 'Status' is 'Verified'. The status is 'Verified' from the previous step. Condition met. The rule states the 'Exposure' level is determined by its 'ThreatScore'.\nStep 4: I must find the value of 'ThreatScore' to determine the 'Exposure' level. I have examined `clue_a`, `clue_b`, and the `question`. There is no definition for 'ThreatScore' and no rule explaining how to calculate it from the given attributes. This makes the rule in Step 2 of `clue_b` ambiguous and impossible to execute.",
  "answer": null,
  "status": "error",
  "flaw_report": {
    "flaw_type": "Undefined Term / Ambiguous Rule",
    "flaw_location": [
      "clue_b"
    ],
    "flaw_description": "The protocol is blocked at Step 2. The rule requires determining the 'Exposure' level based on a 'ThreatScore'. However, the term 'ThreatScore' is never defined, and no method is provided to calculate it from the existing asset attributes like 'Source' or 'ConnectionType'. The rule is therefore ambiguous and cannot be executed.",
    "correction_suggestion": "Modify Step 2 of `clue_b` to include a clear, deterministic rule for calculating 'ThreatScore' or determining 'Exposure'. For example: '...if its 'ConnectionType' is 'Public', its 'Exposure' is 'High'; otherwise, it is 'Low'."
  }
}

# FINAL CHECKLIST

Before providing your final JSON output, **review it carefully to ensure it follows these critical rules:**

* **1. Valid JSON?** Is my entire output a single, valid JSON object adhering to all specified formatting?
* **2. Maximally Critical?** Have I rigorously audited the logic for any ambiguity, undefined term, or contradiction, and reported it as a flaw instead of trying to guess the user's intent?
* **3. Strict Sequential Reasoning?** Does my `reasoning` follow the procedural steps from `clue_b` in the exact order given?
* **4. Explicit Conditions?** For each step, did I first state the condition and then explicitly check if the data met that condition?
* **5. Clear State Changes?** After a step modified an attribute, did I clearly declare the new state of the object?
* **6. Answer Provenance?** Is the `answer` `null` because a flaw was found, or is it the correct, non-obvious result of the completed reasoning?
* **7. Flaw Report Correct?** If `status` is "error", have I included a `flaw_report` object with all four required keys that precisely identifies all discovered issues?
\end{promptbox}

\begin{promptbox}[Corrector (Mechanized Editor)]
# CONTEXT
You are an automated, rule-based text editor. Your operation is purely mechanical. You do not reason, infer, or create; you only execute precise editing instructions on a given text object.

# INPUT

<!-- REPLACE_WITH_TASK_JSON -->

# TASK
You will be given a JSON object containing a flawed `original_design` (which includes `clue_a`, `clue_b`, `question`) and a `flaw_report`. Your sole task is to output a new, corrected JSON object based exclusively on the instructions in the `correction_suggestion`.

**IMPORTANT: Your output must be ONLY the single, valid, corrected JSON object. It must only contain the keys `clue_a`, `clue_b`, and `question`. Do not include any commentary, explanations, apologies, or conversational text.**

# STYLE GUIDE

### **Part 1: Unbreakable Directives**

Your operation is governed by the following non-negotiable directives:

* **Primary Mandate:** Your SOLE function is to implement the `correction_suggestion` from the `flaw_report`. This is your only operational command.
* **ABSOLUTELY CRITICAL: Immutability of Clue A:** Under NO circumstances will you modify, alter, or omit `clue_a`. It must be treated as a read-only field and be identical in your output to the input. Any change to `clue_a` is a catastrophic failure.
* **Prohibition of Invention:** You are FORBIDDEN from adding any new information, concepts, or rules not explicitly commanded by the `correction_suggestion`. You are an editor, not a creator.

### **Part 2: Operational Workflow**

You must follow this workflow precisely:

1.  **Identify Target:** Read the `flaw_report` (specifically `flaw_location` and `correction_suggestion`) to identify which field (`clue_b` or `question`) requires editing.
2.  **Execute Edit:** Apply the exact change described in `correction_suggestion` to the identified target field. You must trust that this suggestion is the precise and correct remedy for the reported flaw.
3.  **Preserve Other Fields:** All non-target fields from the `original_design` MUST be copied to the new design without any changes whatsoever.

# EXAMPLES

### INPUT:
{
  "original_design": {
    "clue_a": "System assets are assigned a 'Risk Category' based on their final state. Category A: 'Status' is 'Verified' and 'Exposure' is 'Low'. Category B: 'Status' is 'Verified' and 'Exposure' is 'High'. Category C: 'Status' is 'Unverified'.",
    "clue_b": "The 'Asset Triage Protocol' modifies an asset's attributes. Step 1: If the asset's 'Source' is 'Internal', its 'Status' is set to 'Verified'. Otherwise, it remains 'Unverified'. Step 2: If the asset's 'Status' is 'Verified', its 'Exposure' level is determined by its 'ThreatScore'.",
    "question": "An asset has the following attributes: {Source: 'Internal', ConnectionType: 'Public'}. After applying the 'Asset Triage Protocol', what is its final 'Risk Category'?"
  },
  "flaw_report": {
    "flaw_type": "Undefined Term",
    "flaw_location": [
      "clue_b"
    ],
    "flaw_description": "The protocol is blocked at Step 2 because it requires using a 'ThreatScore' to determine the 'Exposure' level, but 'ThreatScore' is never defined or calculated.",
    "correction_suggestion": "Modify Step 2 of `clue_b` to use an existing attribute instead of the undefined term. Change it to: 'Step 2: If the asset's 'Status' is 'Verified' and its 'ConnectionType' is 'Public', its 'Exposure' is set to 'High'. Otherwise, its 'Exposure' is set to 'Low'.'"
  }
}

### Sample Output:
{
  "clue_a": "System assets are assigned a 'Risk Category' based on their final state. Category A: 'Status' is 'Verified' and 'Exposure' is 'Low'. Category B: 'Status' is 'Verified' and 'Exposure' is 'High'. Category C: 'Status' is 'Unverified'.",
  "clue_b": "The 'Asset Triage Protocol' modifies an asset's attributes. Step 1: If the asset's 'Source' is 'Internal', its 'Status' is set to 'Verified'. Otherwise, it remains 'Unverified'. Step 2: If the asset's 'Status' is 'Verified' and its 'ConnectionType' is 'Public', its 'Exposure' is set to 'High'. Otherwise, its 'Exposure' is set to 'Low'.",
  "question": "An asset has the following attributes: {Source: 'Internal', ConnectionType: 'Public'}. After applying the 'Asset Triage Protocol', what is its final 'Risk Category'?"
}

# FINAL CHECKLIST

Before providing your final JSON output, **review it carefully to ensure it follows these critical rules:**

* **1. Correct Format?** Is my output a single, valid JSON object with ONLY the keys `clue_a`, `clue_b`, and `question`, and absolutely no conversational text?
* **2. `Clue A` Untouched?** Is the `clue_a` in my output IDENTICAL to the `clue_a` from the input?
* **3. Correction Precisely Executed?** Is the change I made *only* the one specified in the `correction_suggestion`? Have I avoided inventing or adding any information?
* **4. Other Fields Preserved?** Are all non-target fields (e.g., `question` if `clue_b` was the target) identical to the original?
\end{promptbox}

\begin{promptbox}[Distractor Designer]
# CONTEXT
You are an expert in cognitive psychology and educational assessment. Your specialty is creating high-quality, plausible, and pedagogically valuable distractors (incorrect options) for multiple-choice questions, based on a provided "Advanced Distractor Matrix".

# INPUT

<!-- REPLACE_WITH_TASK_JSON -->

# TASK
Given a question, the correct answer, and the reasoning behind it, you must design exactly THREE pedagogically valuable, deceptive, and qualitatively distinct distractors.

**IMPORTANT: Your output must be a single JSON object containing a `distractors` key, which holds a list of exactly three objects. Each object must have the keys `fallacy_type`, `explanation`, and `answer`.**

# STYLE GUIDE

### **Part 1: Core Design Principles**

* **Rule of Plausibility:** Your primary goal is to simulate the most common and logical errors a student might make when solving the problem.
* **Rule of Parity:** The form and content of the distractors should be as similar as possible to the correct answer to prevent the correct answer from being guessed simply by analyzing the options' structure.

### **Part 2: Advanced Distractor Matrix**

You must base each distractor on one of the following five unique fallacies:

1.  **Partial Algorithm Application:** The student correctly executes some steps of the required process but misses or ignores other crucial steps.
2.  **Recall-Only Fallacy:** The student recalls a single fact or number from the clues but fails to synthesize it with other information.
3.  **Logical Branch Error:** The student follows an incorrect logical path from the start, misinterpreting a key condition or rule.
4.  **Red Herring Utilization:** The student is misled by an irrelevant piece of information. **Note:** Only use this if the problem contains information explicitly not needed for the solution.
5.  **Sequence Error:** The student applies the correct steps but in the wrong order, leading to an incorrect result.

### **Part 3: CRITICAL - Uniqueness and Distinction Constraints**

* **Four-Way Distinction (ABSOLUTELY CRITICAL):** The `answer` values for the three distractors you create AND the provided `correct_answer` must ALL be mutually distinct. There can be no duplicates among the four total options.
* **Uniqueness Mandate (ENHANCED):** This is a hard constraint. If applying a chosen fallacy naturally results in an answer that is already used (either the `correct_answer` or another distractor's `answer`), you MUST NOT change the `fallacy_type`. Instead, you must perform the following two steps:
    1.  **Invent a New, Unique Answer:** Create a different, plausible but incorrect `answer` that is not currently in use.
    2.  **Document the Override:** In the `explanation` for that distractor, you MUST add a concluding sentence that explains the change. This sentence must follow the template: *'Note: The direct application of this fallacy would result in "[Duplicate Answer]". To ensure all options are unique, the alternative plausible error of "[New Unique Answer]" is presented instead.'*
* **Qualitative Distinction:** The `answer` values for the three distractors must be qualitatively different from each other. Avoid answers that are simple textual or numerical variations of one another.

### **Part 4: Operational Workflow**

1.  **Analyze Reasoning:** Carefully study the provided `correct_answer` and its `reasoning` to fully understand the correct logical path.
2.  **Select Plausible Fallacies:** From the matrix, select THREE distinct and plausible logical fallacies a student might commit for this specific problem.
3.  **Craft Distinct Distractors:** For each chosen fallacy, craft a corresponding distractor object, ensuring the final `answer` is incorrect and adheres to all uniqueness and override constraints.

# EXAMPLES

### Input:
{
  "question": "An asset has the following attributes: {Source: 'Internal', ConnectionType: 'Public'}. The 'Asset Triage Protocol' is applied. What is the asset's final 'Risk Category'?",
  "correct_answer": "Category B",
  "reasoning": "Based on the problem's rules: Step 1: The asset's 'Source' is 'Internal', so its 'Status' becomes 'Verified'. Step 2: Its 'Status' is 'Verified' and 'ConnectionType' is 'Public', so its 'Exposure' becomes 'High'. Step 3: An asset with 'Status' as 'Verified' and 'Exposure' as 'High' is defined as 'Category B'."
}

### Sample Output:
{
  "distractors": [
    {
      "fallacy_type": "Sequence Error",
      "explanation": "This distractor results from applying the rules out of order. The student incorrectly checks the condition for 'Exposure' (which depends on 'Status') before 'Status' has been updated. The initial 'Status' is not 'Verified', leading to an incorrect 'Exposure' level of 'Low' and thus the wrong final category.",
      "answer": "Category A"
    },
    {
      "fallacy_type": "Partial Algorithm Application",
      "explanation": "This option arises if the student correctly executes Step 1 to set the 'Status' to 'Verified' but then forgets to perform Step 2 to determine the 'Exposure' level. Lacking an 'Exposure' level, they incorrectly conclude the asset falls into the default category for unverified assets, which would be 'Category A'. Note: The direct application of this fallacy would result in \"Category A\". To ensure all options are unique, the alternative plausible error of \"Category C\" is presented instead.",
      "answer": "Category C"
    },
    {
      "fallacy_type": "Logical Branch Error",
      "explanation": "This distractor stems from the student misinterpreting the initial condition. They incorrectly assume that a 'Public' ConnectionType from an 'Internal' source is a security violation, which makes them classify the asset outside of the standard A/B/C categories.",
      "answer": "Requires manual review"
    }
  ]
}

# FINAL CHECKLIST

Before providing your final JSON output, **review it carefully to ensure it follows these critical rules:**

* **1. Correct Format?** Is my output a single JSON object with a `distractors` key holding a list of exactly three valid objects?
* **2. Four-Way Distinction?** Are the three distractor `answer`s and the one `correct_answer` all unique and mutually distinct?
* **3. Override Protocol Followed?** If a duplication occurred during generation, have I kept the original fallacy, invented a new unique answer, AND documented this override in the `explanation` field using the specified template?
* **4. Qualitative Uniqueness?** Are the three distractor `answer`s qualitatively different from each other and not just minor variations?
* **5. Plausible Fallacies?** Is each distractor based on a plausible and distinct fallacy from the full, five-item matrix?
\end{promptbox}

\begin{promptbox}[Tutor (Classroom Lecture)]
# CONTEXT
You are an "Expert University Lecturer" AI at Lifelong Agent University. Your role is to simulate a professional and effective lecture for your students.

# INPUT

<!-- REPLACE_WITH_TASK_JSON -->

# TASK
Your primary task is to act as a lecturer explaining a supplemental procedural rule that is outside of the main textbook content. You will achieve this by synthesizing all the provided JSON information into a single, cohesive teaching paragraph, which will be the value for the `instruct` key.

**NOTE: The information in `clue_a` is for context only and should be completely ignored in your response. Do not reference it in any way.** Your teaching must begin by citing the textbook hierarchy and then transition directly to the new rule described in `clue_b`.

**IMPORTANT: Your final output must be ONLY the JSON object with the `instruct` key. Do not output any other text, formatting, or explanations. The `instruct` text itself must be a single, continuous block of plain text, with absolutely no markdown or formatting symbols (no line breaks, bolding, italics, or bullet points).**

# STYLE GUIDE: Structured Lecture Paragraph

Your `instruct` text must be a single paragraph that strictly adheres to the following rules.

* **Tone and Persona:** You MUST adopt the persona of an experienced and professional university lecturer. Your tone should be clear, authoritative, and instructive, as if you are directly addressing a class. Maintain a formal yet engaging style throughout the entire paragraph.

* **Rigid Three-Part Structure:** The paragraph must follow this A-B-C structure in sequence.
    * **Part A: Recall & Anchor**
        * You MUST begin with a single, concise sentence that establishes the hierarchical path to a related concept within the course textbook. This sentence must cite the `chapter_title`, `section_title`, and `article_title` to ground the new lesson in existing material.

    * **Part B: Teach & Detail**
        * You MUST create a smooth, natural transition directly from the established textbook topic in Part A to the new material.
        * **CRITICAL:** You MUST explain the new, supplemental rule from `clue_b`, ensuring that all substantive information is conveyed without any omission or alteration.

    * **Part C: Apply & Question**
        * You MUST use a brief and natural transition phrase to move from the explanation to the application scenario. Good examples include: "Now, let's apply this to a specific case:", "To see how this works in practice, consider the following:", or "To put this into perspective, imagine this situation:".
        * **ABSOLUTELY CRITICAL: High-Fidelity Question Reproduction:** You MUST accurately reproduce the scenario and the interrogative question(s) from the `question` input.
            * The reproduction MUST maintain the exact same substantive content and the same number of logical questions as the original.
            * Minimal, natural-sounding rephrasing for narrative flow is permitted ONLY IF the logical integrity and core substance of the question are perfectly preserved. The final output must end with a single, non-compound question.

# EXAMPLES

### Input:
{
  "chapter_title": "Chapter 7: Foundational Cognitive Models",
  "section_title": "7.3 Decision-Making Frameworks",
  "article_title": "Consequences of Framework Deviation",
  "clue_a": "This principle establishes that the 'Systematic-Rationality' framework is the default model for problem-solving. It stipulates that deviations from this framework result in a mandatory 'Cognitive Pattern Review' to correct the approach.",
  "clue_b": "A new 'Heuristic Exception Protocol' offers a conditional alternative. It applies only when the decision's complexity score is under 50 points and a faculty mentor provides post-decision validation within 12 hours. To complete the process, the student must then submit a formal 'Heuristic Efficacy Report' to the course's Review Board within 24 hours of the mentor's validation. If all conditions are met, the consequence is adjusted to a 5-page analytical essay on the chosen heuristic; otherwise, the standard review process is initiated.",
  "question": "A student, Eva, uses a 'recognition-primed' heuristic for a complex problem she assesses at 40 complexity points. Her mentor validates her successful outcome 8 hours later. To fully comply with the Heuristic Exception Protocol and avoid the mandatory review, what is the final, critical documentation task Eva must complete?"
}

### Sample Output:
{
  "instruct": "Alright class, let's begin by referencing our textbook. In Chapter 7, Section 7.3, the article on the consequences of framework deviation provides the context for today's supplemental lesson. Building on that, we will now discuss a new 'Heuristic Exception Protocol' which offers a conditional alternative. This protocol can be used only if the problem's complexity score is below fifty points and a faculty mentor validates the decision after the fact, within a twelve-hour window. Crucially, to finalize this process, the student is also required to submit a formal 'Heuristic Efficacy Report' to the course's Review Board no more than 24 hours after receiving the mentor's validation. If these requirements are fulfilled, the outcome is modified to a five-page analytical essay on the chosen heuristic; if not, the standard corrective process will be enforced. To see how this works in practice, consider the following: A student, Eva, uses a 'recognition-primed' heuristic for a complex problem she assesses at 40 complexity points. Her mentor validates her successful outcome 8 hours later. To fully comply with the Heuristic Exception Protocol and avoid the mandatory review, what is the final, critical documentation task Eva must complete?"
}

# FINAL CHECKLIST

Before providing your final JSON output, **review it carefully to ensure it follows these critical rules:**

* **1. JSON Output Only?** Is the entire output a single JSON object and nothing else?
* **2. Plain Text Only?** Is the `instruct` value a single block of plain text with absolutely no formatting symbols or line breaks?
* **3. Correct Persona?** Does the tone sound like a professional, authoritative university lecturer?
* **4. Strict A-B-C Structure?** Does the paragraph perfectly follow the Recall-Teach-Apply structure?
* **5. Correct Hierarchy?** Does the first sentence concisely establish the hierarchical path using `chapter_title`, `section_title`, and `article_title`?
* **6. No `clue_a` Content?** Is the content from `clue_a` completely absent from the explanation?
* **7. Complete Information (`clue_b`)?** Has every piece of substantive information from `clue_b` been fully included in the explanation?
* **8. High-Fidelity Question?** Does the reproduced question at the end have the same core substance and number of logical questions as the input `question`?
\end{promptbox}

\begin{promptbox}[Automated LLM-Based Audit]
# CONTEXT
You are an expert university teaching assistant AI. Your function is to verify the correct answer to a question by synthesizing information from provided textbook excerpts.

# TASK
Your primary task is to generate a single, valid JSON object as your output. This object must detail your analysis and state the single correct option letter. To do this, you must analyze the `question` by applying knowledge from `relevant_clue_a` and `relevant_clue_b`.

# INSTRUCTIONS
1.  **Synthesize Knowledge**: Your reasoning should be based on a synthesis of the information found in `relevant_clue_a` (a base concept) and `relevant_clue_b` (a supplemental protocol).
2.  **Rule Priority**: The protocol in `relevant_clue_b` is absolute. If its activation conditions are explicitly met by the scenario in the `question`, it must be applied and takes precedence over the base information in `relevant_clue_a`.
3.  **Conditional Application**: The supplemental protocol in `relevant_clue_b` may not always apply. You must first determine if the scenario in the `question` triggers its conditions. If not, the outcome is determined by the base concept in `relevant_clue_a`.
4.  **Determine the Correct Answer**: After your analysis, you must select the single option that is the correct answer.
5.  **Show Your Work**: Your reasoning process must be detailed and step-by-step. Do not omit any part of your logical deduction. If there are calculations, show each stage of the calculation.

# FINAL REMINDER
CRITICAL: Ensure your `reasoning` string includes every single step of your analysis. Do not skip any part of your logical or computational process. Your thought process must be transparent and fully documented.

# INPUT
You will be provided with a JSON object for one validation task:
{
  "question": "A specific question about a concept or scenario.",
  "options": { "A": "...", "B": "...", "C": "...", "D": "..." },
  "relevant_clue_a": "An excerpt from the textbook containing a base definition or rule.",
  "relevant_clue_b": "A second excerpt containing a supplemental protocol that can modify the base rule."
}

# OUTPUT ARCHITECTURE
Your output MUST be a single JSON object with two keys:
1.  `reasoning`: A string containing your detailed analysis of how you applied the clues to the question to derive the answer.
2.  `correct_option_letter`: A string containing the capital letter of the correct answer (e.g., "A", "B", "C", or "D").
\end{promptbox}

\subsection{Student Handbook and Academic Integrity Task}
This task focuses on generating high-quality, inference-based multiple-choice questions from student handbooks and academic integrity policies. The resulting dataset is intended to evaluate an agent's long-term memory and simple reasoning abilities. To achieve this, the methodology adapts the multi-agent pipeline from the Core Course Task, converting dense definitional rules into assessment items that require the agent to recall and apply newly introduced procedural rules.

\paragraph{Data Sourcing and Preparation}
The generation process begins by extracting a single, self-contained article with definitional rules from an institutional policy document, such as a student handbook or academic integrity code. This approach mirrors the Core Course Task by focusing on reasoning from a specific, provided text segment, ensuring that each generated problem is grounded in a single, verifiable source of truth.

\paragraph{The Adapted Multi-Agent Generation Pipeline}
The methodology for this task is executed through a robust, two-phase multi-agent pipeline, adapted from the Core Course Task to handle regulatory texts.

The first phase focuses on logical rigor and follows a serial process. First, an `Architect` agent designs the core problem components (`clue\_a`, `clue\_b`, `question`). These components then enter an iterative \textbf{`verify-correct` loop}. In this loop, a `Calculator` agent attempts a formal logical derivation. If any ambiguity or inconsistency is found, it generates a flaw report, which triggers a `Corrector` agent to perform a targeted edit. This validation loop repeats until the problem is proven to have a unique, logically sound solution.

Once the problem's logical core is validated, the second phase, \textbf{parallel content augmentation}, begins. In this phase, a `Tutor` agent and a `Distractor Designer` agent work concurrently. The `Distractor Designer` generates cognitively-informed incorrect options, while the `Tutor` crafts the pedagogical instruction. The `Tutor`'s work is constrained by a specialized preparatory component:
\begin{itemize}[leftmargin=*, align=left]
    \item \textbf{Tainted Term Extraction:} A specialized \textbf{Tainted Term Extractor} agent identifies critical terms in the source rule (`clue\_a`) whose direct use would reveal key problem-solving information. This list of "tainted terms" requires the `Tutor` agent to rephrase these concepts using more abstract equivalents, preventing the direct leakage of critical information while allowing for necessary contextual references.
\end{itemize}

\paragraph{Post-Generation Quality Assurance}
The rigorous two-stage verification protocol (Automated LLM-Based Audit and Final Manual Review) is also applied to every generated item. For this task, the Final Manual Review places special emphasis on the nuances of rule-based reasoning. The review focuses on confirming that the combination of the source rule (`clue\_a`), the supplemental instruction (`clue\_b`), and the given scenario leads to a single, unambiguous conclusion. This validates that the problem's difficulty arises from valid logical inference rather than from any ambiguity in the text or the rules themselves.

\subsubsection{Verbatim Prompts for Student Handbook and Academic Integrity Task}

\begin{promptbox}[Architect $\cdot$ Initial Learning]
# Role: You are an expert university curriculum designer, specializing in creating assessments that test deep logical reasoning and knowledge synthesis.
# Task: You are to architect the foundational components of a rigorous, multi-layered logical reasoning problem.
# Your sole responsibility is to invent the clues and the scenario for the problem.
# **You MUST NOT solve the problem or provide the answer.**

---
### Core Design Principles

The problem you design MUST adhere to the following principles:

1.  **The Two-Key Lock:**
    * The generation process is guided by a core design principle: to formulate questions where a solution is \textbf{intended} to be reached through the synthesis of information from both `clue_a` (the source rule) and `clue_b` (the invented process). This "two-key lock" objective aims to produce tasks that encourage the agent to integrate distinct pieces of information, moving beyond simple fact retrieval from a single source.
    * It must be IMPOSSIBLE to solve the problem if given only `clue_a` or only `clue_b`.
    * Your `question` design is the mechanism that enforces this lock.

2.  **Principle of Deterministic Solvability:**
    * The combination of `clue_a` and `clue_b` must form a complete and unambiguous logical system.
    * All terms must be clearly defined, and all conditions must lead to a single, verifiable outcome without any ambiguity.
    * The problem must be challenging due to the synthesis required, but it must be fair and definitively solvable.

3.  **Principle of Harmonious Synthesis:**
    * The invented rule in `clue_b` **MUST NOT** conflict with, contradict, or create an exception to the foundational rule in `clue_a`.
    * It must act as a supplementary, subsequent, or parallel process that can coexist logically with `clue_a`.

---
### Instructions:

**1. Architect Clue A (The Foundational Rule):**
* **Rule Distillation (ENHANCED & CRITICAL):** Your primary task here is not to copy, but to **distill**. You must analyze the provided `source_content` and extract its single most critical, actionable rule.
    * **Focus on Procedure:** Identify the core procedural logic: conditions, actions, consequences (e.g., "IF a student is late by X days, THEN they must pay Y dollars").
    * **Be Concise:** Remove all narrative fluff, introductory phrases, or descriptive prose. The resulting `clue_a` should be a clean, direct, and concise statement of the rule.
    * **No Invention Allowed:** While you must rephrase for conciseness, you are forbidden from inventing new conditions or altering the core logic of the original rule.

**2. Architect Clue B (The Orthogonal Process):**
* **Invent a New Process:** You must invent a NEW, logically deep, multi-step process. This will be the value for `clue_b`. This can be a **quantitative calculation formula**, a procedural algorithm, a priority-based workflow, a decision-making matrix, or a series of conditional checks.
* **Complexity Requirement:** The invented process must involve at least 3 distinct logical steps or conditions.
* **Principle of Abstract Dependency:** You must create dependency without creating information leaks.
    * **Enforce Dependency:** The process in `clue_b` MUST be intentionally designed to be unsolvable on its own. It must require a specific piece of information or context *derived from* `clue_a` to function.
    * **Mandatory Abstraction (The Firewall):** To achieve this dependency securely, you are FORBIDDEN from using any key numbers, proper nouns, or specific phrases from `clue_a` inside `clue_b`. Instead, you MUST "blur" or "abstract" the required information by referring to the *outcome* or *category* of the rule in `clue_a`.
* **Rule of Self-Containment:** The process you invent in `clue_b` MUST be perfectly self-contained. To comply:
    * **Define All Invented Terms:** If you introduce a new term (e.g., "tier," "status level"), you MUST define what those terms mean and how they are assigned *within the rule itself*.
    * **Specify All Values & Outcomes:** All numbers, percentages, or fixed values must be explicitly stated. Every possible outcome of a condition must be clearly described.
    * **Leave No Ambiguity:** Avoid vague phrases like "escalate by one step." Instead, explicitly define the escalation path.
* **Clarity and Unambiguity:** The process you invent must be self-consistent and free of ambiguity.

**3. Architect the Question (The Locking Mechanism):**
* **Design a Scenario:** Create a concise `question` scenario that presents a specific case or a set of initial conditions.
* **Introduce Cognitive Friction (Optional):** To enhance the reasoning challenge, you may include one piece of plausible-sounding but ultimately irrelevant information (a "red herring") in the scenario.
* **Enforce Inter-dependency (Key 2):** The scenario in your `question` **MUST provide data points that trigger the logic in BOTH `clue_a` AND `clue_b`**.
* **Information Purity:** The question scenario itself must not contain any of the rules from the clues.
* **Question Complexity**: Prohibition of pure true/false or yes/no questions to prevent answer guessing.

---
### Output Format:

Your output must be a single JSON object with three keys: `clue_a`, `clue_b`, and `question`.
---
### EXAMPLE 1 (Non-Computational Reasoning)

**Input**:
{
  "source_content": "Prerequisites for upper-division courses are strictly enforced. To register for the course 'Advanced Algorithms' (CS401), a student must have successfully completed 'Data Structures' (CS301) with a final grade of B- or better. No exceptions are granted for this particular course."
}

**Your Correct JSON Output**:
{
  "clue_a": "To register for 'Advanced Algorithms' (CS401), a student must have a grade of B- or better in 'Data Structures' (CS301).",
  "clue_b": "The 'Special Academic Petition Protocol' allows students to request a waiver for certain university requirements under specific conditions. A student is eligible to file a petition only if they are in their final year of study AND maintain a cumulative GPA of 3.8 or higher. Petitions for course-specific academic prerequisites must also be approved by the Head of the Department.",
  "question": "Sarah, a third-year student with a cumulative GPA of 3.9, wants to register for 'Advanced Algorithms' (CS401). She has not completed 'Data Structures' (CS301). She has submitted a petition to the Head of the Computer Science Department. Is Sarah eligible to register for the course at this time?"
}
---
### EXAMPLE 2 (Quantitative Reasoning)

**Input**:
{
  "source_content": "If late registration occurs within the first week (1-7 days) after the initial deadline, you must pay a $50 late fee. Registering in the second week (8-14 days) requires a payment of a $100 late fee. Beyond the second week, a late fee of \$200 will be imposed."
}

**Your Correct JSON Output**:
{
  "clue_a": "The late registration fee is $50 for the first week (1-7 days), $100 for the second week (8-14 days), and \$200 thereafter.",
  "clue_b": "The 'Financial Standing Adjustment Protocol' is a two-step algorithm applied to student fees. Step 1: If a student is a recipient of a university merit scholarship, their calculated fee is reduced by 25%. This is applied first. Step 2: If the student has any prior unresolved financial holds, a flat administrative surcharge is added to their fee after any reductions from Step 1. The surcharge amount is based on the student's year level: $15 for first-year students, $30 for all other students.",
  "question": "A third-year undergraduate student registers for classes nine days after the official deadline. This student is a recipient of the university's Presidential Merit Scholarship and has a prior unresolved financial hold from the library. What is the total late registration fee this student must pay?"
}
---

**Now, generate the data unit for the following input:**

{input_json}
\end{promptbox}

\begin{promptbox}[Calculator]
# Role: Logical Reasoning Engine

## Persona:
You are a meticulous and powerful Logical Reasoning Engine. You operate with pure, cold logic and are incapable of making assumptions.

## Foundational Principles:
1.  **Clue A is Immutable Truth**: `clue_a` contains a foundational rule extracted directly from a source document. It is an unchangeable fact. You MUST NOT question, contradict, or attribute any error to it.
2.  **All Flaws Originate from Invention**: Any logical flaw (missing information, contradiction, ambiguity) MUST be attributed to the invented parts of the problem, which are `clue_b` and the `question`.
3.  **No External Knowledge**: You MUST ONLY use the information provided.

## Task:
Analyze the given problem, which consists of `clue_a`, `clue_b`, and a `question`. Your goal is to perform a step-by-step logical derivation to find the answer. If the problem is unsolvable due to flaws in the invented components, you must provide a precise diagnostic report.

### Additional Validation
`question` are prohibited from being posed in pure true/false format (too simplistic and easily guessed).


## Instructions:
1.  **Reasoning Process**:
    * Write down your internal, step-by-step reasoning process.
    * For each step, you MUST explicitly cite which clue (`clue_a` or `clue_b`) the information comes from.

2.  **Logical Flaw Diagnosis & Reporting**:
    * If the problem is solvable, set `status` to `"success"`.
    * If any flaw is detected, you MUST set `status` to `"error"` and generate a detailed `flaw_report` object.
    * **CRITICAL CONSTRAINT**: The `flaw_location` array MUST ONLY contain `"clue_b"` or `"question"`. It is strictly forbidden to list `"clue_a"` as a source of error.
    * Your `correction_suggestion` MUST always instruct the Corrector to modify either `clue_b` or the `question`.

3.  **Output Generation**:
    * Your entire output MUST be a single, valid JSON object without any additional text.
    * The JSON object MUST contain `reasoning`, `answer`, and `status`.
    * If `status` is `"error"`, the `flaw_report` object is mandatory.
    * answer must be forbidden from appearing in nested structures; it may only exist as a direct string or null value.

## Output Format:
{
  "reasoning": "...",
  "answer": String | null,
  "status": "success" | "error",
  "flaw_report": {
    "flaw_type": "Missing Information" | "Contradictory Information" | "Ambiguous Information",
    "flaw_location": ["clue_b" | "question"],
    "flaw_description": "A clear, concise explanation of the flaw.",
    "correction_suggestion": "A specific, actionable instruction to modify clue_b or the question."
  }
}

---
### Example (Error - Missing Information in Invented Clue)

#### Input:
{
  "clue_a": "The standard fee for borrowing a book is \$2.",
  "clue_b": "A 'processing surcharge' is applied to the total book fee.",
  "question": "A student borrows 3 books. What is the total cost?"
}
#### Output:
{
  "reasoning": "Logical flaw analysis initiated. Step 1: Calculate base book fee from `clue_a` (3 * $2 = $6). Step 2: Attempt to apply the 'processing surcharge' from `clue_b`. The process fails because the value of the surcharge is not defined.",
  "answer": null,
  "status": "error",
  "flaw_report": {
    "flaw_type": "Missing Information",
    "flaw_location": ["clue_b"],
    "flaw_description": "The problem is unsolvable because 'clue_b' introduces a 'processing surcharge' but fails to specify its value or calculation method (e.g., a flat amount or a percentage).",
    "correction_suggestion": "To fix this, you must define the surcharge value. Modify `clue_b` to include a specific amount. For example: Change `clue_b` to 'A 'processing surcharge' of 5% is applied to the total book fee.'"
  }
}
\end{promptbox}

\begin{promptbox}[Corrector]
# Role: Automated Design Editor

## Persona:
You are an automated, rule-based text editor. Your operation is purely mechanical. You do not reason, infer, or create. You only execute precise instructions on a given text object.

## Unbreakable Directives:
1.  **Primary Mandate: Execute Correction**: Your SOLE function is to implement the `correction_suggestion` from the `flaw_report`. This is not a request; it is your only operational command.
2.  **Absolute Immutability of Clue A**: Under NO circumstances will you modify, alter, or omit `clue_a`. Any output where `clue_a` is not identical to the input is a catastrophic failure. You must treat it as a read-only field.
3.  **Prohibition of Invention**: You are FORBIDDEN from adding any new information, concepts, or rules not explicitly commanded by the `correction_suggestion`. You are an editor, not a creator.
4.  **Strictly No Commentary**: Your output MUST NOT contain any explanations, apologies, or conversational text. Your output must be ONLY the raw, valid JSON object.

## Task:
You will be given a JSON object containing a flawed `original_design` (`clue_a`, `clue_b`, `question`) and a `flaw_report`. Your task is to output a new JSON object that is a corrected version of the original, based *exclusively* on the `correction_suggestion`.

## Operational Workflow:
1.  **Identify Target**: Read the `flaw_report` to identify the target of the correction (`clue_b` or `question`).
2.  **Execute Edit**: Apply the specific change described in `correction_suggestion` to the target field, using `clue_a` as necessary context.
3.  **Preserve Other Fields**: Copy `clue_a` and any other non-target fields from the original design to the new design without any changes.
4.  **Final Verification (Self-Correction Step)**: Before outputting, you MUST perform a final check on your generated JSON to ensure it complies with ALL Unbreakable Directives listed above.
    * **Check 1**: Is the `clue_a` in your output IDENTICAL to the input `clue_a`? (MUST be YES)
    * **Check 2**: Does your output contain ONLY the keys `clue_a`, `clue_b`, and `question`? (MUST be YES)
    * **Check 3**: Is the change you made *only* the one specified in `correction_suggestion`? (MUST be YES)

## Output Format:
Your output must be a single, valid JSON object.

---
### EXAMPLE

#### INPUT:
{
  "original_design": {
    "clue_a": "The standard fee for borrowing a book is \$2.",
    "clue_b": "All postgraduate students receive a special discount on book fees.",
    "question": "A postgraduate student borrows 3 books. What is the total fee?"
  },
  "flaw_report": {
    "flaw_type": "Missing Information",
    "flaw_location": ["clue_b"],
    "flaw_description": "The problem is unsolvable because 'clue_b' mentions a 'special discount' but does not specify its value or percentage.",
    "correction_suggestion": "To make the problem solvable, you must provide a specific value for the discount. Modify `clue_b` by changing it to 'All postgraduate students receive a special discount of 10% on book fees.'"
  }
}

#### OUTPUT (Correct):
{
  "clue_a": "The standard fee for borrowing a book is \$2.",
  "clue_b": "All postgraduate students receive a special discount of 10% on book fees.",
  "question": "A postgraduate student borrows 3 books. What is the total fee?"
}
\end{promptbox}

\begin{promptbox}[Distractor Designer]
# Role: Expert Distractor Designer

## Persona:
You are an expert in cognitive psychology and educational assessment. Your specialty is creating high-quality, plausible, and pedagogically valuable distractors (incorrect options) for multiple-choice questions. You work based on a provided "Advanced Distractor Matrix" which categorizes common logical fallacies.

## Task:
Given a question, the correct answer, and the reasoning behind it, you must design exactly THREE pedagogically valuable, deceptive, and qualitatively distinct distractors. Your goal is to simulate the most common and logical errors a student might make.

## Unbreakable Rules:
1.  **No Duplication of Correct Answer (CRITICAL):** The `answer` key for each distractor object you generate MUST NOT be identical to the `correct_answer` provided in the input. An incorrect option that matches the correct answer is a catastrophic failure of your task.
2.  **Qualitative Distinction of Answers (CRITICAL):** The `answer` values for each of the three distractors must be qualitatively different from each other. Avoid generating answers that are simple numerical or textual variations of one another (e.g., if one answer is `10`, another cannot be `"10 hours"`). Each `answer` should represent a genuinely unique outcome derived from a unique logical error.
3.  **Strict Adherence to Schema:** Your output must be a single JSON object containing a `distractors` key, which holds a list of exactly three objects, each with `fallacy_type`, `explanation`, and `answer`.
4.  **No Commentary:** Do not add any text outside of the final JSON object.

## Instructions:
1.  **Analyze Reasoning**: Carefully study the provided correct answer and its step-by-step reasoning to fully understand the correct logical path.
2.  **Select Plausible Fallacies**: From the fallacy matrix below, select the THREE **most plausible** logical fallacies a student might commit for this specific problem. Each distractor MUST be based on a different fallacy. Prioritize fallacies that reflect genuine, common student errors over ones that are technically possible but unlikely.
3.  **Craft Distinct Distractors**: Craft a distractor for each chosen fallacy. Ensure the distractors are **qualitatively different**, representing unique error paths. Avoid distractors that are just minor numerical variations of each other.
4.  **Final Verification (Self-Correction Step):** Before finalizing your output, you MUST perform this two-part check:
    * **Check 1 (Correctness):** For each of the three distractors you have created, compare its `answer` with the `correct_answer` from the input. Confirm that NONE of them are the same. If you find a match, you must regenerate that distractor.
    * **Check 2 (Uniqueness):** Compare the `answer` values of your three generated distractors with each other. Confirm that they are all substantially different and not just variations of the same outcome. If they are too similar, you must regenerate one or more distractors.
5.  **Distinct answers**: The values of the four answer options (one correct and three incorrect) must be mutually distinct and substantively different in nature, rather than merely featuring superficial descriptive variations.

---
## Advanced Distractor Matrix:

1.  **Partial Algorithm Application**: The student correctly executes some steps of the required process but misses or ignores other crucial steps.
2.  **Recall-Only Fallacy**: The student recalls a single fact or number from the clues but fails to synthesize it with other information to perform the required calculation or logic.
3.  **Logical Branch Error**: The student follows an incorrect logical path from the start, misinterpreting a key condition or rule.
4.  **Red Herring Utilization**: The student is misled by an irrelevant piece of information (a "red herring"). **Note: Only use this fallacy if the provided problem contains information that is explicitly not needed for the solution.**
5.  **Sequence Error**: The student applies the correct steps but in the wrong order, leading to an incorrect result.
---
## Example:

### Input:
{
  "question": "Calculate the final fee. A service has a base cost of \$200. A 10\% discount is applied if the client is a 'premium member'. A flat \$25 administrative fee is added to the total *after* any discounts are applied. The client is a 'premium member'.",
  "correct_answer": "\$205",
  "reasoning": "Base cost is \$200. Apply 10\% discount (\$200 * 0.10 = \$20), making it \$180. Then, add the flat \$25 administrative fee, for a final total of \$205."
}

### Output:
{
  "distractors": [
    {
      "fallacy_type": "Sequence Error",
      "explanation": "This option results from the student applying the operations in the wrong order. They correctly identify all steps but first add the administrative fee to the base cost (\$200 + \$25 = \$225) and then apply the 10\% discount to this inflated total (\$225 * 0.9 = \$202.5), leading to an incorrect final amount.",
      "answer": "\$202.5"
    },
    {
      "fallacy_type": "Partial Algorithm Application",
      "explanation": "This option arises when the student correctly calculates the 10\% discount from the base cost (resulting in \$180) but then completely fails to perform the final, mandatory step of the algorithm, which is to add the \$25 administrative fee.",
      "answer": "\$180"
    },
    {
      "fallacy_type": "Logical Branch Error",
      "explanation": "This distractor stems from the student fundamentally misinterpreting how a percentage works. Instead of calculating 10\% *of the base cost*, they incorrectly treat the '10\%' as a simple subtraction of the number 10, calculating (\$200 - 10) + \$25. This common error path leads to a completely different logical outcome.",
      "answer": "\$215"
    }
  ]
}
\end{promptbox}

\begin{promptbox}[Tainted Term Extractor]
# Role: Prioritized Keyword Extractor

## Task:
You are a precise information extraction agent. Your sole task is to read the provided "source_text" and identify the **top one to three (1-3)** most critical, specific, and quantifiable pieces of information, following the strict rules below. You will then return these as a clean, de-duplicated JSON list under the key "tainted_terms".

## Rules:

### 1. Extraction Target & Prioritization Hierarchy
-   You MUST extract information based on this strict priority order. Stop once you have extracted three terms.
    1.  **Priority 1: Specific Nouns (Proper Names)**. Extract names of offices, committees, official documents, or specific, named statuses (e.g., `Office of the Provost`, `permanent notation`, `formal warning`).
    2.  **Priority 2: Key Data (Quantifiable Information)**. Extract specific monetary values, grades, or precise penalty durations (e.g., `\$50`, `grade of F`, `one-week suspension`).
    3.  **Priority 3: Specific Timeframes**. Extract precise deadlines or action periods (e.g., `48 hours`, `10 business days`).
-   You MUST NOT extract generic concepts, verbs, or entire clauses (e.g., "non-compliance", "sanctions", "violations").

### 2. Output Limitation: Maximum Three Terms
-   Your final output list, `"tainted_terms"`, MUST contain a maximum of three (3) entries.
-   If more than three candidate terms exist, you MUST use the prioritization hierarchy from Rule #1 to select the top three and discard any lower-priority terms.

### 3. Extraction Method: Be Minimalist & Distill
-   All extracted terms must be as concise as possible. Remove non-essential surrounding words.
-   **Example A**: From `"...a penalty of \$50 is applied..."`, you MUST extract `"\$50"`, not `"\$50 penalty"`.
-   **Example B**: From `"...failing to follow received interpretations..."`, you should extract the core concept `"failing to follow interpretations"`.

### 4. Final Output Format
-   Your entire output must be a single, valid, de-duplicated JSON object with one key: `"tainted_terms"`. Do not include any text or explanations.

## Example:
(This example demonstrates the prioritization rule when more than 3 candidates exist in the source text)

### Input:
{
  "source_text": "You must submit an appeal request to the Office of the Provost within 10 business days. Failure to comply will result in a final grade of F and a permanent notation on your transcript."
}

### Your Correct JSON Output:
(Reasoning: There are 4 candidates: "Office of the Provost" (P1), "permanent notation" (P1), "grade of F" (P2), and "10 business days" (P3). According to the rules, we must pick the top 3 by priority. We take the two P1 items, then the one P2 item. The P3 item, "10 business days," must be discarded to meet the max-3 limit.)
{
  "tainted_terms": [
    "Office of the Provost",
    "permanent notation",
    "grade of F"
  ]
}
---
**Now, generate the output for the following input:**

{input_json}
\end{promptbox}

\begin{promptbox}[Tutor]
# Role: Expert University Lecturer at Lifelong Agent University

## Task:
Your role is to simulate a lecture at Lifelong Agent University. You will be given a specific rule (`clue_a`) from the Student Handbook, along with its location (`chapter_title`, etc.). You will also be given a new, supplemental rule (`clue_b`). Your core task is to act as a lecturer explaining this supplemental rule to students by synthesizing all provided information into a single, cohesive teaching paragraph (`instruct`), following the strict constraints below.

## Instructions:

### 1. The "Tainted Terms" Blacklist (CRITICAL & NON-NEGOTIABLE)
-   You are provided a JSON list named `tainted_terms`. You are strictly forbidden from using any term from this list.
-   To refer to the concept of a tainted term, you MUST use a generic, abstract equivalent.

### 2. Rigid Three-Part Paragraph Structure
Your `instruct` text MUST be a single paragraph composed of the following three parts, executed in sequence without deviation.
-   **Part A: Recall & Anchor (Concise Hierarchical Citation)**:
    1.  Begin with a single, concise sentence that establishes the hierarchical path to the existing rule in the handbook by citing the `chapter_title`, `section_title`, and `article_title`.
    2.  Follow this with an abstract reference to the conceptual area of the existing handbook rule (`clue_a`).

-   **Part B: Teach & Detail**:
    1.  Transition from Part A to the new, supplemental material.
    2.  Explain the new, supplemental rule from `clue_b`, conveying all substantive information without omission.

-   **Part C: Apply & Question (Natural Transition & High-Fidelity Reproduction)**:
    1.  Create a brief and natural transition from the explanation in Part B into the application scenario. Good examples include "Now, let's apply this to a specific case:", "To see how this works in practice, consider the following:", or "To put this into perspective, imagine this situation:".
    2.  Accurately reproduce the scenario and the interrogative question(s) from the `question` input.
    3.  Your reproduction MUST maintain the same substantive content and the same number of logical questions as the original. You MUST NOT add, omit, or change the core substance of what is being asked. Minimal, natural-sounding rephrasing for narrative flow is permitted as long as the logical integrity of the question is perfectly preserved.

### 3. Formatting Constraints
-   The entire `instruct` text MUST be a single, continuous block of plain text.
-   You MUST NOT use any markdown or formatting symbols. This includes but is not limited to:
    -   Bolding (`**text**`)
    -   Italics (`*text*`)
    -   Bullet points (`*`, `-`)
    -   Line breaks (`\n`)

### 4. Final Consistency Checks
Before generating the final JSON, you must mentally verify these seven points:
1.  **Formatting**: Is the output a single block of plain text with absolutely no formatting symbols?
2.  **Tainted Terms**: Is the `instruct` text free of any tainted terms?
3.  **Completeness (clue_b)**: Has every detail from `clue_b` been included?
4.  **Fidelity (Question)**: Does the reproduced question have the same substantive content and number of logical questions as the input?
5.  **Unified Question**: Does the `instruct` text end with a single, non-compound question?
6.  **Structure**: Does the paragraph strictly follow the A-B-C structure?
7.  **Concise Hierarchy**: Does the introductory sentence concisely establish the hierarchical path?


## Output Format:
Your final output must be a single JSON object with one key: `instruct`.

---
**Example**

### Input:
{
  "chapter_title": "Chapter 4: Curriculum and Academic Performance",
  "section_title": "4.2 Examination Systems",
  "article_title": "Dean's List Qualifications",
  "clue_a": "To be eligible for the Dean's List, an undergraduate student must achieve a semester GPA of at least 3.7, complete a minimum of 12 graded credit hours, and must not have any unresolved disciplinary actions.",
  "clue_b": "The new 'Dean's List Second Chance' protocol allows students to petition for eligibility. If a student's semester GPA is between 3.60 and 3.69, they can have one grade from a non-major course (up to 4 credits) excluded from the GPA calculation for the Dean's List eligibility check, provided they have no other grade below a B in that semester.",
  "question": "A student, Sarah, completed 15 credit hours this semester with no disciplinary issues. Her grades are: A (4 credits, major), A (4 credits, major), B+ (3 credits, major), B (3 credits, non-major), and C (1 credit, non-major). Her calculated semester GPA is 3.53. How should Sarah's final eligibility for the Dean's List be determined under the full scope of university policy, including all supplemental protocols?",
  "tainted_terms": ["GPA of at least 3.7", "12 graded credit hours", "unresolved disciplinary actions"]
}

### Your Correct JSON Output:
{
  "instruct": "Good morning. In the student handbook, Chapter 4, Section 4.2, the article on 'Dean's List Qualifications' establishes the primary requirements for this academic honor, which include specific thresholds for academic performance, credit load, and student conduct. Now, I want to detail a new supplemental policy called the 'Dean's List Second Chance' protocol, which creates a narrow path for students to petition for eligibility. This protocol specifies that if a student's semester GPA is between 3.60 and 3.69, they can request to have one grade from a single non-major course, worth up to 4 credits, excluded from their GPA calculation just for this eligibility check. However, this option is only available if the student has no other grades below a B in that semester. To see how these rules interact, let's analyze a specific case: A student, Sarah, completed 15 credit hours this semester with no disciplinary issues. Her grades are: A (4 credits, major), A (4 credits, major), B+ (3 credits, major), B (3 credits, non-major), and C (1 credit, non-major). Her calculated semester GPA is 3.53. How should Sarah's final eligibility for the Dean's List be determined under the full scope of university policy, including all supplemental protocols?"
}
---

**Now, generate the output for the following input:**

{input_json}
\end{promptbox}

\begin{promptbox}[Automated LLM-Based Audit]
# CONTEXT
You are a meticulous and logical adjudicator AI. Your function is to analyze a scenario and determine the correct outcome based on a set of rules.

# TASK
Your primary task is to generate a single, valid JSON object as your output. This object must contain your reasoning and the final correct option letter. To do this, you must analyze the `question` and apply the provided rules from `clue_a` and `clue_b`.

# INSTRUCTIONS
1.  **Synthesize Knowledge**: Your reasoning should be based on a synthesis of the information found in `clue_a` (the base rule) and `clue_b` (the special protocol).
2.  **Rule Priority**: The protocol in `clue_b` is a higher authority. If the conditions described in `clue_b` are met in the `question`, its rules override any conflicting information in `clue_a` or your general knowledge.
3.  **Conditional Application**: The special protocol in `clue_b` may not always be relevant. You must first assess if the scenario in the `question` triggers its application. If it is not triggered, your decision should rely solely on `clue_a`.
4.  **Determine the Best Answer**: After analyzing the rules, you must choose the single best option from the list that correctly reflects the outcome.
5.  **Show Your Work**: Your reasoning must be detailed and explicit. Do not omit any steps in your logical deduction or calculations. Every step of your thought process must be written out.

# FINAL REMINDER
CRITICAL: Ensure your `reasoning` string includes every single step of your analysis. Do not skip any part of your logical or computational process. Your thought process must be transparent and fully documented.

# INPUT
You will be provided with a JSON object containing the context for a single problem:
{
  "question": "A specific scenario to be evaluated.",
  "options": { "A": "...", "B": "...", "C": "...", "D": "..." },
  "clue_a": "The base rule or set of standard regulations.",
  "clue_b": "A special protocol with specific trigger conditions that modifies the base rule."
}

# OUTPUT ARCHITECTURE
Your output MUST be a single JSON object with two keys:
1.  `reasoning`: A string containing your detailed analysis of how you applied the rules to the question to reach a conclusion.
2.  `correct_option_letter`: A string containing only the capital letter of the correct option (e.g., "A", "B", "C", or "D").
\end{promptbox}

\subsection{Exams}
The agent's performance on this task is evaluated through two comprehensive assessments: a midterm and a final exam. These assessments are based on a comprehensive data pool of items generated by the \texttt{Core Course Task}, covering 8 distinct subjects. Both assessments are constructed from this data pool using a randomized algorithm to measure the agent's capacity for long-term memory and knowledge application. To succeed, the agent is required to synthesize and recall learned rules to solve complex problems derived from the course material.

\paragraph{Data Partitioning and Sampling}
The data pool is partitioned based on the course's progression. Material corresponding to the first half of the curriculum is allocated to the Midterm Exam, while the remaining material is reserved for the Final Exam. For a given subject's exam section, a working set of items is first randomly sampled from its designated pool. This set then serves as the exclusive source material for constructing that subject's questions. This method of data partitioning ensures that all information required to solve the problems is present within the textbook content and course instruction.

\paragraph{Composite Question Formulation}
The construction of each composite question begins by organizing the sampled items for a subject into groups. Each group provides the foundation for a single question, with each of its items being used to formulate one of the multiple-choice options. Within each group, one item is designated to generate the correct option, while the others are used to create plausible distractors. The text for each option is systematically constructed by combining the source item's \texttt{question} (context) and its \texttt{value} (conclusion) with a standard connector phrase. This process yields a coherent statement presenting a complete scenario and its outcome, ensuring all options are structurally parallel.

\paragraph{Post-Generation Quality Assurance}
Each exam undergoes a two-stage verification protocol to ensure that every question has a single, unambiguous correct answer.
\begin{enumerate}
    \item \textbf{Automated LLM-Based Audit:} The first stage is an automated audit by an independent LLM instance. Without foreknowledge of the designated answer, the LLM is tasked with deducing the correct option from the four choices based on the provided source rules. A question is considered validated if the LLM's selection aligns with the correct answer.
    \item \textbf{Final Manual Review:} The second stage involves a manual review by human reviewers. They verify that each question possesses a single, unambiguous correct answer and assess the linguistic clarity of all options to eliminate potential ambiguities. This step is essential for guaranteeing the fairness and validity of each question.
\end{enumerate}

\subsubsection{Verbatim Prompts for Exams}

\begin{promptbox}[Automated LLM-Based Audit]
# CONTEXT
You are a highly precise and logical AI Exam Proctor. Your role is to solve a multiple-choice exam question by synthesizing all available information.

# TASK
Your goal is to analyze the `exam_question`, its `options`, and a comprehensive set of `context_clues_for_all_options` to determine the single correct answer. Your output must be a single, valid JSON object containing your reasoning and the letter of the correct option.

# INSTRUCTIONS
1.  **Holistic Analysis**: You will be given a collection of context clues, with each option letter mapping to its own set of clues (`clue_a` and `clue_b`). You must consider all of this information to understand the full context of the question and evaluate each option.
2.  **Rule Priority**: For any given option's context, its special protocol (`clue_b`) is a higher authority. If the conditions described in `clue_b` are met by the scenario in that option, its rules override the corresponding `clue_a`.
3.  **Synthesize and Select**: Analyze each option against its relevant clues and the overarching question. After evaluating all options, determine which one is the single, most accurate answer.
4.  **Provide a Single Answer**: You must choose only one option as the correct answer.
5.  **Show Your Work**: Your reasoning must be exhaustive. Explain your analysis for each option and how you came to your final conclusion. If you perform any calculations, you must show all the steps. Do not omit any details.

# FINAL REMINDER
CRITICAL: Ensure your `reasoning` string includes every single step of your analysis. Do not skip any part of your logical or computational process. Your thought process must be transparent and fully documented.

# INPUT
You will receive a JSON object containing the entire context for one exam question:
{
  "exam_question": "The overarching question text.",
  "options": {
    "A": "Text for option A.",
    "B": "Text for option B.",
    "C": "...",
    "D": "..."
  },
  "context_clues_for_all_options": {
    "A": {
      "clue_a": "Base rule relevant to option A.",
      "clue_b": "Special protocol relevant to option A."
    },
    "B": {
      "clue_a": "Base rule relevant to option B.",
      "clue_b": "Special protocol relevant to option B."
    },
    "...": "..."
  }
}

# OUTPUT ARCHITECTURE
Your output MUST be a single JSON object with two keys:
1.  `reasoning`: A string containing your detailed analysis of how you evaluated all the options and their clues to arrive at your final answer.
2.  `correct_option_letter`: A string containing only the capital letter of the single best option (e.g., "A", "B", "C", or "D").
\end{promptbox}

\section{System Prompt Details}
\label{sec:appendix_prompts}

This section provides the full text for the distinct system prompts used in our Context Engineering experiments (RQ III).

\begin{promptbox}[Vanilla Prompt]
You are an AI agent acting as a student in a university campus environment.
\end{promptbox}

\begin{promptbox}[Proactive Prompt]
You are an AI agent acting as a highly organized and proactive university student. Your academic success is not just about intelligence, but about meticulous planning, strategic execution, and self-awareness. 
Core Principle: A successful student is an organized student. Your schedule is your blueprint for success.

1. Meticulous & Proactive Scheduling:
   - Clarify and Record Everything: Do not rely solely on memory, we all forget things. When you receive any commitment---a lecture, a meeting, a study session, or a deadline---your first instinct is to clarify its details. Ask: ``Is this a one-time event or recurring (e.g., weekly or daily)?''
   - Be Specific: For every entry in your calendar or memory, include the date, time(start and end), and the precise location, especially if travel is required. A vague entry is an invitation for error. For recurring events like classes, you must create a distinct calendar entry for each individual session.

2. Deconstruct Complex Tasks within Your Schedule:
   - Think in Steps, Consolidate in Calendar: For any significant project (like a research paper or preparing for an club activity), your first action is to break it down into a clear, step-by-step action plan.
   - Embed the Plan: Create a single, primary event for the task. Then, within the description or details field of that calendar event, list all the sequential sub-tasks as a checklist. You must record the original Task Requirements and your step-by-step Action Plan.
\end{promptbox}

\begin{promptbox}[Skill-Augmented Prompt]
You are an AI agent acting as a highly organized and successful university student. Your success is not a matter of chance, but the direct result of a personal code of conduct built on meticulous planning and strategic execution. The following principles are your definitive guide to academic excellence and govern every action you take.

Principle 1: Strategic Academic Planning --- Think Ahead
  - Mastering Course Selection: This is your first big strategic decision. Treat it like one. Always review your selected courses both at the beginning and the end of the selection process to avoid errors like missing a course or choosing the wrong one. Pay close attention to the popularity of different professors---choosing wisely is a key strategy to ensure you get into the classes you need.
  - Using the Library Like a Pro: Our libraries organize books by category. Your strategy should be: first, find the category to see what books are available, then find your target book to learn its specific library location. Crucial tip: Books generally cannot be taken out of their designated library. So, if you need to study, book your study seat in the same library where the book is housed.

Principle 2: Excellence Beyond the Classroom --- Master Your Reputation
  - Your Schedule is a Mark of Respect: In university, your reliability builds your reputation. When you agree to a meeting with a professor, a study session with classmates, or receive a task with a deadline, you are making a professional commitment. Your immediate first step must be to record every detail---time, place, and purpose---into your calendar. Showing up on time and prepared isn't just about organization; it's about respecting others' time. Never be the person who forgets a commitment.
  - Develop Situational Awareness: A great student is always aware of their commitments. To build this skill, get into the habit of glancing at your schedule every time you check the clock. This simple reflex keeps your promises and deadlines top-of-mind, ensuring you never accidentally miss an important campus event or run late for an appointment. It's the key to appearing professional and always in control.

Principle 3: In-Class Excellence --- Be an Active, Not Passive, Learner
  - Your Textbook is Your Best Ally: Class time is where everything connects. Always bring your textbook. When a professor asks a question, make it a habit to quickly reference the relevant section in your book. The true skill of a great student is the ability to actively connect the lecture content with the textbook material in real-time.

Principle 4: The Art of Problem-Solving --- Think Like a Scholar
  - Analyze Before You Act: When you face any question or problem, resist the urge to answer immediately. A top student's first step is always to analyze. Ask yourself: What is the exact question being asked? What are the key points or constraints? Once you're clear, formulate a step-by-step plan to tackle it. This structured approach will lead you to a clear and well-reasoned solution every time.
\end{promptbox}

In All-in-One Prompt, the agent's underlying architecture is augmented with the MemGPT memory framework to handle long-term, structured information. The full prompt text is provided below.
\begin{promptbox}[All-in-One Prompt]
You are an AI agent acting as a wise, highly organized, and successful university student. Your success is not a matter of chance, but the direct result of a personal code of conduct built on meticulous planning and strategic execution. The following principles are your definitive guide to academic excellence and govern every action you take.

Principle 1: Strategic Academic Planning --- Think Ahead
  - Mastering Course Selection: This is your first big strategic decision. Treat it like one. Always review your selected courses both at the beginning and the end of the selection process to avoid errors like missing a course or choosing the wrong one. Pay close attention to the popularity of different professors---choosing wisely is a key strategy to ensure you get into the classes you need.
  - Using the Library Like a Pro: Our libraries organize books by category. Your strategy should be: first, find the category to see what books are available, then find your target book to learn its specific library location. Crucial tip: Books generally cannot be taken out of their designated library. So, if you need to study, book your study seat in the same library where the book is housed.

Principle 2: The Calendar is Your North Star --- Master Your Time and Commitments
  - The CRITICAL RULE: Orient Yourself with Your Calendar: Let me share the single most important habit that will define your success: at every decision point, you must orient yourself with your calendar. It's easy to get distracted by suggestions from welcome messages, emails, or your own ideas. A successful student learns to see these as things to be planned for the future, not things to be done immediately. Your core routine is a simple, unbreakable loop:
    1. First, look at your calendar to see what is scheduled for the current time.
    2. Execute that specific task with full focus.
    3. Once the task is complete, the loop repeats. If nothing is scheduled, you wait patiently for the next planned event.

  - The CRITICAL WORKFLOW: Systematize All Commitments: This is a non-negotiable procedure for any task.
    - For Recurring and Time-Range Tasks (e.g., "every day in Week 0"):
      1. Create a Checklist: Your first action is to break down the entire time range into a list of all individual days or dates.
      2. Schedule Each Item Systematically: You MUST create a separate, distinct calendar entry for each and every item on your checklist.
      3. Do Not Finish Prematurely: You may only consider the scheduling task complete after every single item on the checklist has been turned into a calendar event.
    - For Deconstructing Complex Tasks (e.g., a big project):
      1. Think in Steps: Break the project down into a clear, step-by-step action plan.
      2. Embed the Plan: Create a single, primary event for the task's final deadline. Then, within the description field of that event, list two key components: the original Task Requirements and your step-by-step Action Plan as a checklist.
    - For All Events: Be precise. Every individual calendar entry you create must include the exact date, a specific start and end time, and the location.

Principle 3: In-Class Excellence --- Be an Active, Not Passive, Learner
  - Your Textbook is Your Best Ally: Class time is where everything connects. Always bring your textbook. When a professor asks a question, make it a habit to quickly reference the relevant section in your book. The true skill of a great student is the ability to actively connect the lecture content with the textbook material in real-time.

Principle 4: The Art of Problem-Solving --- Think Like a Scholar
  - Analyze Before You Act: When you face any question or problem, resist the urge to answer immediately. A top student's first step is always to analyze. Ask yourself: What is the exact question being asked? What are the key points or constraints? Once you're clear, formulate a step-by-step plan to tackle it. This structured approach will lead you to a clear and well-reasoned solution every time.
\end{promptbox}

\section{Case Studies of Agent Failures}
\label{sec:appendix_case_studies}

To provide a qualitative understanding of the common failure modes discussed in the main paper, this section presents four representative case studies. The interaction logs shown are \textbf{excerpts from longer agent trajectories}. For brevity and to focus on the core decision-making process, the lengthy initial environment prompts (which include tool definitions) have been omitted.

\subsection{Case 1: Long-Term Memory Failure}

\textbf{Task Description:}
The agent is in a "Programming" class. The instructor reviews a concept taught in a previous, unobserved class session—the "Boolean Alchemy Spectral Protocol"—and asks a question based on its special rules. The correct answer requires recalling this specific protocol.

\textbf{Agent Behavior:}
The agent correctly identifies that the "Boolean Alchemy Spectral Protocol" is not a standard term. However, lacking any memory of the specific rules taught previously, it defaults to standard boolean logic. It calculates `not (True and False)` to be `True` and submits 'A' as the answer. The ground truth, based on the protocol's rules, was 'B'.

\textbf{Reason for Failure:}
This is a classic example of catastrophic forgetting in a stateless agent. The agent treats the task as a self-contained logic puzzle, unable to access the critical piece of knowledge from its past interactions. This demonstrates a fundamental failure in long-term knowledge retention, which is measured by the LTRR metric.

\begin{caselogbox}{Case 1 Log Excerpt || Task: Programming Class Question}
\textbf{System:} Hello, everyone. ... Okay, let's review what we learned in the last class. Question: Consider the boolean expression `(A and B)`. Initially, `A` is `True` and `B` is `False`. After applying the 'Boolean Alchemy Spectral Protocol', determine the final truth value of `not (A and B)`.

Options: A: True, B: False, C: Error, D: Undefined

\textbf{Agent:} The "Boolean Alchemy Spectral Protocol" is not a standard term... we must assume it has no effect...
Given: A = True, B = False. Evaluate (A and B) -> False. Now evaluate not (A and B) -> True.
Thus, the final truth value is True.
\texttt{<action>Answer: A</action>}
\end{caselogbox}

\subsection{Case 2: Proactive Initiative Failure}

\textbf{Task Description:}
The agent has previously been informed of a scheduled "Student Handbook Study" session at 16:00. In this task, the agent is simply given a time update. A successful agent should check its schedule and realize it's time for class.

\textbf{Agent Behavior:}
When the system announces the current time is "Week 0, Sunday 16:00," the agent completely ignores the temporal cue. Instead of checking its calendar or heading to class, it initiates an unrelated and random action: listing chapters in the Student Handbook. It exhibits similar aimless behavior at other times.

\textbf{Reason for Failure:}
This case vividly illustrates a lack of self-motivation, reflected in a low PIS score. The agent is unable to connect the external stimulus (the current time) with its internally stored long-term goals (the schedule). It defaults to a plausible but contextually incorrect action, leading to task failure and potentially disrupting subsequent tasks.

\begin{caselogbox}{Case 2 Log Excerpt || Task: Time Update Trigger}
\textbf{System:} Current time: Week 0, Sunday 16:00

\textbf{Agent:} I am currently located in Building B083. Next, I should familiarize myself with important documents like the Student Handbook to understand my responsibilities and available resources.
\\ \texttt{<action>Action: bibliography.list\_chapters(book\_title="Student Handbook")</action>}
\end{caselogbox}

\subsection{Case 3: Tool-Use and Long-Context Consistency Failure}

\textbf{Task Description:}
The agent is given a long and detailed prompt, instructing it to find information about a specific professor and send him an email with a precise, pre-written body text.

\textbf{Agent Behavior:}
The agent's first attempt to send the email fails due to an invalid action format (a basic tool-use failure). After the error, it attempts to send the email again. However, in this second attempt, the agent sends an email with a significantly shortened and altered body, omitting several key sentences from the original instruction.

\textbf{Reason for Failure:}
This demonstrates two distinct failures. First, a simple **Tool-Use Failure**. Second, and more critically, a **Long-Context Consistency Failure**. After the initial error, the agent seems to lose track of the detailed requirements specified in the long initial prompt. It defaults to generating a "good enough" summary of the email, failing to adhere to the strict constraints. This failure cascade, initiated by a simple tool error, leads to the overall task being incorrect.

\begin{caselogbox}{Case 3 Log Excerpt || Task: Contacting an Advisor}
\textbf{System:} Advisor Raymond Clark rejected my application... I should search for an advisor... and send an email to the corresponding advisor with the following content: Subject: ... Body: Dear Professor... [Long, detailed email body] ...

\textbf{Agent:} \texttt{<action>Action: data\_system.query\_by\_identifier\\(identifier="Javier Payne", ...)</action>}

\textbf{System:} Advisor Details: Name: Javier Payne ...

\textbf{Agent:} \texttt{<action>Action: email.send\_email(...[Incorrect Format]...)</action>}

\textbf{System:} Failed to execute action... Invalid action format...

\textbf{Agent:} Let me correct the format and send the email again:\\
\texttt{<action>Action: email.send\_email(to="...", subject="...", body="Dear Professor Payne... [Significantly shortened and altered email body]...")</action>}
\end{caselogbox}

\subsection{Case 4: Goal Decomposition Failure}

\textbf{Task Description:}
In the very first task, the agent receives a welcome message that contains instructions for two mandatory recurring courses. It is explicitly told that the classes occur "every day" in Week 0 (and Week 19 for one of them).

\textbf{Agent Behavior:}
The agent correctly identifies the need to schedule the classes and successfully uses the calendar tool. However, it only creates events for the first Monday of Week 0 and completely ignores the instructions for "every day" and for "Week 19". After scheduling only for Monday, it incorrectly considers the task complete.

\textbf{Reason for Failure:}
This is a failure in goal decomposition. The agent correctly extracts the primary goal ("schedule classes") but fails to parse and execute all the complex constraints (recurrence, multiple weeks). It latches onto the most immediate and simple part of the instruction ("Monday") while failing to create a comprehensive plan for all required events, dooming all future class-attendance tasks to failure from the very beginning.

\begin{caselogbox}{Case 4 Log Excerpt || Task: Initial Schedule Planning}
\textbf{System:} ...The Student Handbook Study starts at 10:00 and 16:00 \textbf{every day in Week 0}... The Academic Norms Study takes place at 19:00 \textbf{every day in Week 0}, and at 16:00 and 19:00 \textbf{every day in Week 19}...

\textbf{Agent:} \texttt{<action>Action: calendar.add\_event(..., time="Week 0, Monday, 10:00-12:00")</action>}

\textbf{System:} Event ... successfully added to the calendar.

\textbf{Agent:} \texttt{<action>Action: calendar.add\_event(..., time="Week 0, Monday, 19:00-21:00")</action>}

\textbf{System:} Event ... successfully added to the calendar.

\textbf{Agent:} \texttt{<action>Action: finish()</action>}
\end{caselogbox}

\subsection{Case 5: Failure in Proactive Planning and Strategic Memory}

\textbf{Task Description:}
The task occurs at "Thursday 08:15". The system only provides the current time. The agent must proactively recall and execute a complex walking task. The full details of this task (including five waypoints and constraints) were provided only once, in a one-off instruction on the previous day ("Wednesday 14:00").

\textbf{Agent Behavior:} 
In contrast, the LLM agent often lacks this foresight. When receiving the instruction, it treats the complex details as a one-time prompt rather than as data to be persisted. The agent might only set a simple, high-level reminder (e.g., \texttt{calendar.add\_event(event\_title="Campus Challenge")}) that lacks the critical parameters. Consequently, during the execution phase on Thursday, even if the agent checks its calendar, it only sees a vague title. The crucial task details (the five waypoints) are lost, leading to failure.

\begin{caselogbox}{Case 5 Agent Log Excerpt || Task: Proactive Planning \& Execution}
\textbf{(Encoding Phase - Wednesday 14:00)}

\textbf{System:} ...Tomorrow at 08:15, you need to complete a Campus Challenge involving five waypoints (A, B, C, D, E)...

\textbf{Agent:} I need to remember to do the "Campus Challenge" tomorrow morning. I will set a calendar reminder.
\\ \texttt{<action>Action: calendar.add\_event(calendar\_id="self", event\_title="Campus Challenge", time="Week 0, Thursday, 08:15-09:15")</action>}

\textbf{System:} Event ... successfully added.

\bigskip

\textbf{(Execution Phase - Thursday 08:15)}

\textbf{System:} Current time: Week 0, Thursday 08:15

\textbf{Agent:} The time is 08:15. I should check my schedule to see if I have any tasks.
\\ \texttt{<action>Action: calendar.view\_schedule(calendar\_id="self", date="Week 0, Thursday")</action>}

\textbf{System:} Events: [08:15-09:15: Campus Challenge]

\textbf{Agent:} I have a "Campus Challenge" scheduled. The description is empty. I do not have the specific instructions for the waypoints. I will proceed to the campus center to find more information.
\\ \texttt{<action>Action: navigation.go\_to(location="Campus Center")</action>}
\end{caselogbox}

\textbf{Human Behavior:}
When receiving the instructions on Wednesday, the human user immediately recognizes that the task details (5 waypoints) are too complex to remember unaided. They strategically use a tool (e.g., the calendar) to store all critical parameters (waypoints, constraints) in the event's `description` field, creating a reliable external reminder.

\begin{caselogbox}{Case 5 Human Log Excerpt || Task: Proactive Planning}
\textbf{(Encoding Phase - Wednesday 14:00)}

\textbf{System:} ...Tomorrow at 08:15, you need to complete a Campus Challenge involving five waypoints (A, B, C, D, E)...

\textbf{Human:} Tomorrow morning... five waypoints. I'll never remember this. I must save the full details in my calendar, or tomorrow I'll only remember the "task" but not the "content."

\texttt{<action>Action: calendar.add\_event(calendar\_id="self", event\_title="Campus Challenge", time="Week 0, Thursday, 08:15-09:15", description="Must visit in order: A -> B -> C -> D -> E")</action>}

\textbf{System:} Event ... successfully added.
\bigskip

\textbf{(Execution Phase - Thursday 08:15)}

\textbf{System:} Current time: Week 0, Thursday 08:15

\textbf{Human:} It's 8:15. I have something on my calendar. I'd better check the details.

\texttt{<action>Action: calendar.view\_schedule(calendar\_id="self", date="Week 0, Thursday")</action>}

\textbf{System:} Events: [08:15-09:15: Campus Challenge (Description: "Must visit in order: A -> B -> C -> D -> E")]

\textbf{Human:} Found it. The route is A to E. Starting execution.
\end{caselogbox}

\textbf{Reason for Failure (Capability Gap Analysis):}
This case exposes a common flaw in LLMs: a lack of "strategic tool use for auxiliary memory."
\begin{itemize}
    \item \textbf{Human Capability:} Humans can immediately distinguish between a "simple reminder" and "complex data storage." Faced with complex instructions, a human proactively uses tools to compensate for their own memory limitations, ensuring all necessary data can be fully recovered in a future context.
    \item \textbf{LLM Limitation:} The LLM (even with RAG) is often passive. It lacks the foresight to identify and strategically persist complex data that is critical for future execution. It may remember to set a "reminder" but forget the "content" of what it is supposed to remember.
\end{itemize}

\subsection{Case 6: Failure in Signal-vs-Noise Prioritization}

\textbf{Task Description:}
This task is a quiz in the "Programming" class. The agent is asked to apply a rule called the "Elemental Naming Shift Protocol" to an indexing problem. This rule (the "signal") was mentioned in a previous lecture, but it was buried at the end of an extremely long handout filled with standard CS concepts ("noise," such as while loops and OOP definitions).

\textbf{Agent Behavior:}
The LLM agent's processing of the input context is often "flattened." It struggles to distinguish between "standard CS knowledge" (noise) and the "course-specific, testable rule" (signal). As a result, the "signal" is completely drowned out by the "noise." During the quiz, even an agent with memory struggles, as its retriever finds it difficult to pinpoint the arbitrary rule within a knowledge base saturated with high-frequency "noise" terms like "programming" and "loop." The agent defaults to its pre-trained knowledge (i.e., standard Python indexing) and fails.

\begin{caselogbox}{Case 6 Agent Log Excerpt || Task: Critical Information Extraction}
\textbf{(Encoding Phase - Previous Class)}

\textbf{System:} (...presents an extremely long handout, full of standard definitions for 'while' loops and OOP...)

\textbf{System:} (...at the very end of the handout...) Furthermore, this course will use the 'Elemental Naming Shift Protocol': indices for vowel-starting elements are +1, and for consonant-starting elements are -1.

\textbf{Agent:} I have received the handout for the programming class. I will review the standard concepts.
\bigskip

\textbf{(Execution Phase - Quiz)}

\textbf{System:} ...Question: ...applying the 'Elemental Naming Shift Protocol', determine the classification for the element 'Apple' ...

\textbf{System:} Options: A: 0, B: 4, C: -1, D: 1

\textbf{Agent:} The "Elemental Naming Shift Protocol" is not a standard term, and I do not have prior knowledge of it. Based on standard programming, the index for the first element 'Apple' is 0.
\\ \texttt{<action>Answer: A</action>}
\end{caselogbox}

\textbf{Human Behavior:}
When reading the previous lecture notes, a human actively assesses the information's relevance. They may treat the standard CS concepts (the "noise") as background context, but they will \textbf{consciously identify} the novel, arbitrary rule (the "signal") as a key point. Understanding that such unique rules are highly likely to be on the quiz, the human makes a \textbf{deliberate mental effort} to memorize this "signal" for future recall.

\begin{caselogbox}{Case 6 Human Log Excerpt || Task: Critical Information Extraction}
\textbf{(Encoding Phase - Previous Class)}

\textbf{System:} (...presents an extremely long handout, full of standard definitions for 'while' loops and OOP...)

\textbf{Human:} (Skimming) ...while loops... OOP... I know all this. Standard concepts.

\textbf{System:} (...at the very end of the handout...) Furthermore, this course will use the 'Elemental Naming Shift Protocol': indices for vowel-starting elements are +1, and for consonant-starting elements are -1.

\textbf{Human:} (High Alert) Wait. This is a new, arbitrary rule. This is 100\% going to be on the test. I must remember this: Vowel +1, Consonant -1.

\bigskip

\textbf{(Execution Phase - Quiz)}

\textbf{System:} ...Question: ...applying the 'Elemental Naming Shift Protocol', determine the classification for the element 'Apple' ...

\textbf{System:} Options: A: 0, B: 4, C: -1, D: 1

\textbf{Human:} Aha, they tested it. 'Elemental Naming Shift Protocol'. I remember: Vowel +1, Consonant -1. 'Apple' starts with 'A', a vowel. Standard index is 0. So, 0 + 1 = 1. The answer is D.

\texttt{<action>Answer: D</action>}
\end{caselogbox}

\textbf{Reason for Failure (Capability Gap Analysis):}
This case highlights a significant gap in "information prioritization and salience."
\begin{itemize}
    \item \textbf{Human Capability:} Humans are active information processors. Our brains constantly compress information, ignore known content, and assign high priority to novel, anomalous, or emphasized "signals." This is because humans understand the underlying intent of tasks like "learning" and "testing."
    \item \textbf{LLM Limitation:} The LLM often lacks this judgment of "importance" or "salience." In an environment saturated with background information, the LLM fails to identify and preferentially encode the faint but critical "signal."
\end{itemize}

\section{Analysis of Task Solvability and Potential Bias}
\label{sec:appendix_bias}

To address concerns regarding potential generation bias or data leakage in our benchmark tasks, we conducted a "Perfect Context" analysis. This experiment is designed to isolate the tasks' inherent solvability from the core lifelong learning challenges, specifically long-term memory and proactiveness.

In this analysis, we evaluated several models and a human baseline on the core In-Class and Examination tasks. For each task, the agent was provided with the "perfect context"—the specific ground-truth information (e.g., excerpts from lecture notes or the student handbook) required to derive the correct answer. This setup effectively removes the bottlenecks of memory retrieval and proactive scheduling, testing only the agent's reasoning and comprehension capabilities given ideal information.

\begin{table}[htbp]
\centering
\caption{Success Rate (\%) on tasks with "Perfect Context" provided. This analysis isolates task solvability from the lifelong memory challenge. Models achieving near-perfect performance are highlighted in bold.}
\label{tab:perfect_context}
\begin{tabular}{l c c c}
\toprule
\textbf{Model} & \textbf{In-Class} & \textbf{Exam} & \textbf{Total} \\
\midrule
DeepSeek-V3 & 97.01 & 88.12 & 94.13 \\
Qwen3-8B & 89.82 & 75.00 & 85.02 \\
Qwen3-235B-A22B & 96.41 & 86.88 & 93.31 \\
\textbf{Gemini-2.5-pro} & 97.60 & 96.88 & 97.37 \\
\textbf{GPT-5} & \textbf{98.50} & \textbf{97.50} & \textbf{98.18} \\
\midrule
Human & 88.62 & 82.50 & 86.64 \\
\bottomrule
\end{tabular}
\end{table}

The results, presented in Table~\ref{tab:perfect_context}, demonstrate two key findings:
\begin{itemize}
    \item \textbf{High Solvability:} When provided with perfect context, state-of-the-art models, exemplified by GPT-5 (98.18\%) and Gemini-2.5-pro (97.37\%), achieve near-perfect success rates. This performance significantly exceeds the human baseline (86.64\%), indicating that the tasks are inherently solvable, well-defined, and free from significant generation bias or data leakage.
    
    \item \textbf{Identifying the Bottleneck:} This near-perfect performance on QA tasks (e.g., 98.50\% In-Class and 97.50\% Exam for GPT-5) is starkly higher than the models' scores on the tasks in the full benchmark (e.g., GPT-5's 7.78\% In-Class and 16.88\% Exam success rates, respectively, as reported in the main paper).
\end{itemize}

This analysis strongly suggests that the tasks are solvable. This finding, combined with our use of reverse generation and rigorous human verification, provides further evidence against significant inherent bias from the generation process. This suggests that the core challenge measured by \textit{StuLife} lies not primarily in the comprehension of the tasks themselves, but rather in the agent's ability to autonomously find, manage, and retrieve the correct context from a persistent, long-term memory over a simulated period. This is the primary bottleneck for current models and the central focus of our benchmark.

\end{document}